\documentclass{article} 
\usepackage{iclr2027_conference, times}
\iclrfinalcopy

\usepackage{amsmath,amsfonts,bm}

\def\eqref#1{equation~\ref{#1}}

\def\1{\bm{1}}

\DeclareMathAlphabet{\mathsfit}{\encodingdefault}{\sfdefault}{m}{sl}
\SetMathAlphabet{\mathsfit}{bold}{\encodingdefault}{\sfdefault}{bx}{n}

\usepackage{hyperref}
\usepackage{url}
\usepackage{booktabs}
\usepackage[table]{xcolor}
\usepackage{xcolor}
\usepackage{xspace}
\usepackage{multirow}
\usepackage{array}
\usepackage{graphicx}
\usepackage{fontawesome5}
\usepackage{subcaption}
\usepackage{amssymb}
\usepackage{wrapfig}

\newcommand{\lflvqa}{\textsc{LF-VideoQA}\xspace} 

\title{\lflvqa: Evaluating Long-Video Understanding with Long-Form Theory-of-Mind Questions}

\author{Shaden Shaar$^*$ \& Bradon Thymes$^*$ \& Majd Aldaye \& Claire Cardie \&  Bharath Hariharan\\
Department of Computer Science \\
Cornell University \\
Ithaca, NY 14850, USA \\
\texttt{\small \{ss2753, bmt63\}@cornell.edu}
}

\begin{document}
\maketitle

\def\thefootnote{*}
\footnotetext{These authors contributed equally to this work}
\renewcommand{\thefootnote}{\arabic{footnote}}

\begin{abstract}
The hardest part of understanding a long video is not pinpointing any one moment but synthesizing many, e.g., answering a question that requires tracking what develops across hours of content and articulating how the pieces fit together.  
To date, most long-video QA datasets require only a short answer or a multiple-choice selection, 
oversimplifying the long-video understanding task to retrieval. 
We introduce \lflvqa, a dataset for long-form, open-ended question answering over long narrative videos built on theory-of-mind questions: a character's belief depends on everything they have seen or been told, so each question demands assembling and reasoning from evidence distributed across the video. 
The dataset contains more than 6,300 questions on 94 recent films; each requires reasoning over scenes spanning an average of 23 minutes and comes with a question-specific rubric for recall and precision scoring.
The dataset is created entirely automatically by annotating screenplays with scene events and each character's first-order beliefs and meta-beliefs, then generating (question, answer, rubric) triples through LLM-based question generation and verification; human studies validate their quality.
We evaluate four families of state-of-the-art open-source VLMs across three settings and find the task highly challenging: 
(i) even an oracle with access to the screenplay falls well short of complete answers; 
(ii) the best non-oracle model attains barely half of the recall level of the oracle; 
(iii) unlike on prior long-video benchmarks, captions and subtitles underperform the video itself, indicating that the questions need visual evidence;
and (iv) recall is the predominant bottleneck, with answers omitting many required rubric claims.
\end{abstract}

\begin{figure}[t]
  \centering
  \small
  \includegraphics[width=.92\textwidth]{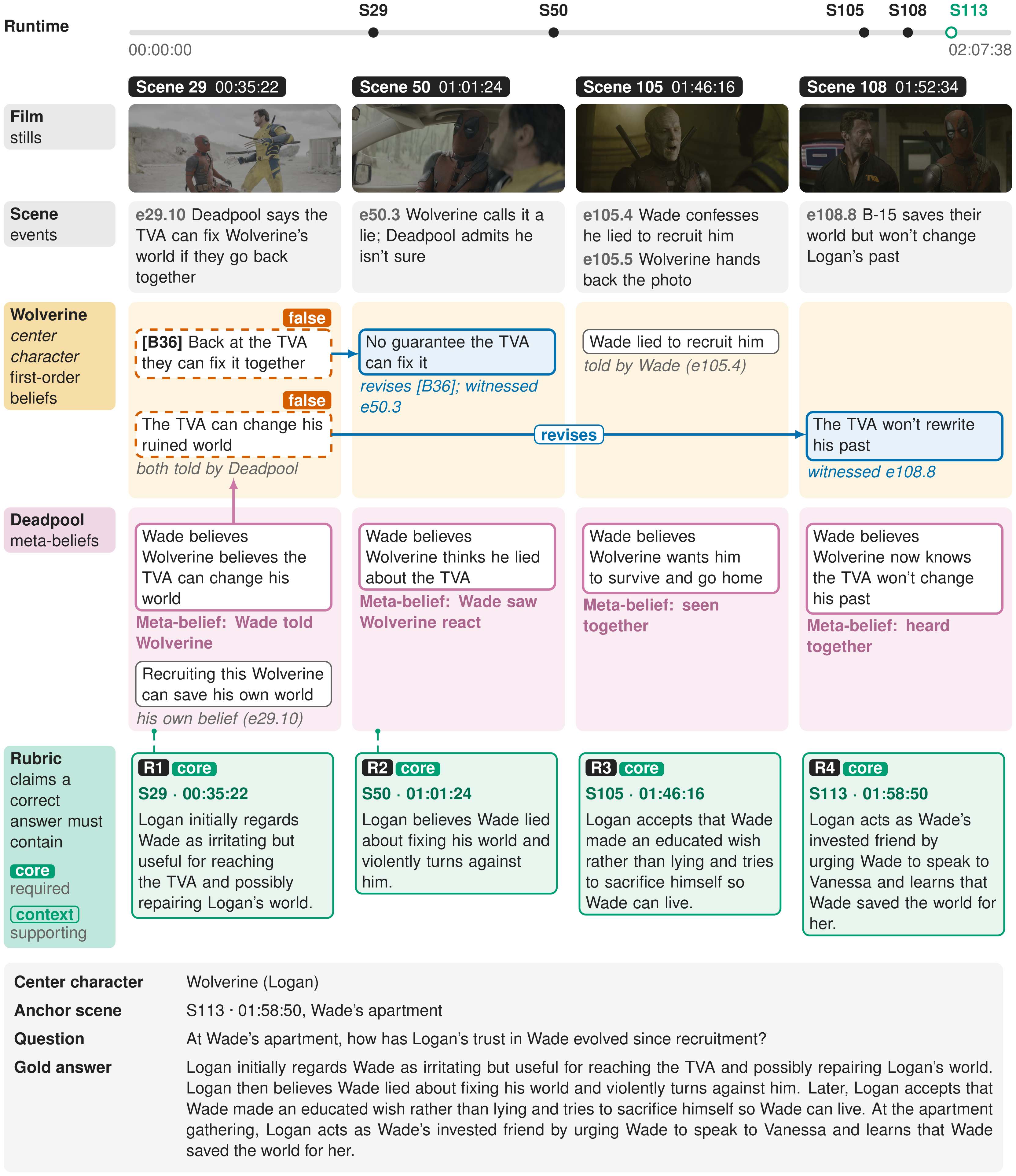}
  \caption{
  \textbf{Dataset Construction on an Example.}
  This figure shows (question, answer, rubric) from \emph{Deadpool \& Wolverine} (2024).
  \emph{Preprocessing:} screenplay scenes are time-stamped against the subtitles and decomposed into scene events.
  \emph{Character belief annotation:} scene by scene, we extract and verify each character's first-order beliefs and meta-beliefs.
  \emph{Question generation and verification:} for a center character (Wolverine) and an anchor (scene~113), we generate the triple and verify every rubric claim against the screenplay.
  }
  \label{fig:belief-example}
\end{figure}

\section{Introduction}


Understanding a long video means following how it develops as a whole, relating what happens now to what happened before~\citep{Wu2021TowardsLV}. In a narrative video (i.e. Movies), events unfold over hours, people acquire new information and change their beliefs, and earlier interactions shape later decisions~\citep{vicol_moviegraphs_2018, papalampidi-etal-2019-movie}. The evidence needed to explain these developments is distributed across many scenes, none of which provides a complete answer on its own~\citep{fung-etal-2023-deepmaven}. 
Long-video understanding therefore requires synthesizing evidence from many moments into a coherent explanation.


Consider the question in Figure~\ref{fig:belief-example}, asked near the end of \emph{Deadpool \& Wolverine} (2024): ``At Wade's apartment, how has Logan's trust in Wade evolved since recruitment?''
Logan's trust rests on a false belief.
At minute 35, Wade (Deadpool) promises Logan (Wolverine) that the TVA (the agency that polices the film's timelines) can repair his ruined world.
At minute 61, Wade admits he is not sure the TVA can fix anything, and then at minute 112, Logan sees the TVA decline to rewrite his past.
A complete answer must trace this arc across 83 minutes of the film, and no single scene contains both the belief and its correction.
Long-form answers make each of these steps assessable by exposing which connections a model draws, which claims it supports, and which parts of the explanation it omits, since a long answer can be decomposed into claims that are checked one by one~\citep{min-etal-2023-factscore, song-etal-2024-veriscore}.
These, theory-of-mind (ToM) question, indeed can capture the true nature of long-output and long-input.


Existing long-video question-answering benchmarks predominantly use multiple-choice or short-answer formats~\citep{mangalam_egoschema_2023, fu2024videomme, wu_longvideobench_2024, wang2024lvbench}.
Although they provide long inputs, many of their questions can be answered by retrieving a few relevant frames or short segments~\citep{mangalam_egoschema_2023}, and the multiple-choice format further allows answering by elimination.
Moreover, captions, subtitles, or even the question alone can support strong performance on many long-video benchmarks~\citep{zhang-etal-2024-simple}, raising questions about the degree to which they depend on visual information.

We introduce \lflvqa, a dataset for long-form, open-ended ToM question answering over long narrative videos, with more than 6,300 questions on 94 recent films. Both the input and the output are long: answering each question requires reasoning over an average of 4.2 scenes spanning 23 minutes, and reference answers average 61 words. The questions follow a film's cast rather than its protagonist, asking about a median of 15.5 characters per film. Our construction pipeline is fully automatic (Figure~\ref{fig:belief-example}): it annotates each film's screenplay, long used as a structured text view of films~\citep{papalampidi-etal-2019-movie, papalampidi-etal-2020-screenplay, stage_2026}, with scene events and each character's first-order beliefs and meta-beliefs (beliefs about other characters' beliefs), then generates (question, answer, rubric) triples through multiple rounds of LLM-based question generation and verification. Each rubric lists the claims a correct answer should make, verified against the screenplay, and supports automatic scoring of recall and precision beyond surface similarity to a reference answer. Human studies validate the quality and interpretability of the resulting questions, answers, and rubrics.

We evaluate four families of state-of-the-art open-source vision-language models and find substantial room for improvement. Even an oracle with access to the screenplay falls well short of producing complete answers, while the best non-oracle model achieves barely half the oracle’s recall. Video input also outperforms text-based representations such as captions and subtitles, suggesting that these representations omit visual evidence needed to answer the questions. Across models and settings, recall is the predominant bottleneck: model answers leave out many of the rubric claims required for a complete response. Together, these findings reveal a persistent gap between identifying relevant information and incorporating that information into a sufficiently complete explanation.

Our contributions are:

\begin{itemize}
  \item \textbf{\lflvqa}, a dataset for long-form, open-ended QA over long narrative videos, with questions requiring evidence from distant scenes and question-specific evaluation rubrics.
  \item \textbf{A fully automatic construction pipeline} that uses screenplay-derived events, beliefs, and meta-beliefs to generate and verify question–answer–rubric triples, with quality validated through human studies.
  \item \textbf{An evaluation of four open-source VLM families} showing substantial difficulty producing complete answers, an advantage for video over text-based representations, and missing required claims as the predominant performance bottleneck.
\end{itemize}
\section{Related Work}
\label{sec:related}

\paragraph{Long inputs, short outputs.}
Video QA benchmarks have grown from clips of seconds to videos of an hour or more, but the answers they require have mostly stayed short (Table~\ref{tab:benchmark-comparison} in Appendix~\ref{app:benchmark-comparison}).
Most long-video benchmarks are multiple-choice~\citep{mangalam_egoschema_2023, fu2024videomme, wu_longvideobench_2024, wang2024lvbench, chandrasegaran_hourvideo_2024}, as are most film and TV benchmarks~\citep{tapaswi_movieqa_2016, lei-etal-2018-tvqa, Zhang2023MoVQAAB, rawal2024cinepile}, and many open-ended benchmarks accept a phrase: answers average 1.9 words in ActivityNet-QA~\citep{yu_activitynetqa_2019}, 2.6 in NExT-QA~\citep{Xiao2021NExTQANP}, and 2.3 in MovieChat-1K~\citep{song2024moviechatdensetokensparse}.
Longer outputs appear in two other settings, neither of which matches ours.
Video summarization produces multi-sentence outputs~\citep{zhou2025mlvubenchmarkingmultitasklong, papalampidi-lapata-2023-hierarchical3d, mahon-lapata-2024-modular}, but they describe the video rather than answer a question about it.
Recent benchmarks pose open-ended questions about long narrative videos~\citep{fung-etal-2023-deepmaven, ataallah-etal-2025-infinibench, jain2026narrativealignedlongform}, but their reference answers are short (19--27 tokens in DeepMovieQA). 
\lflvqa asks for question-specific explanations averaging 61 words, decomposed into verified claims, so we measure which required claims an answer states, omits, or contradicts rather than assigning it a single score.

\paragraph{Theory-of-mind questions.}
Theory of mind is classically tested with false-belief questions answered in a word, such as ``Where will Sally look for her marble?''. However, we expand that in out work to include other variaents of belief states. 
We FANToM adds free-form questions about what participants who missed part of a conversation believe~\citep{kim-etal-2023-fantom}, and CharToM-QA grades answers about novel characters against several key points~\citep{zhou-etal-2025-essence}.
Video ToM benchmarks ask about short clips~\citep{sap-etal-2019-social}, household activities~\citep{jin-etal-2024-mmtom}, or short films~\citep{villa-cueva-etal-2025-moments} with multiple-choice answers; MMToM-QA, for instance, asks which of two beliefs a person more likely holds.
MovieGraph-ToM asks about feature films but gives models selected keyframes and script segments rather than the film~\citep{moviegraphtom_2026}.
A question such as ``how has Logan's trust in Wade evolved since recruitment?'' needs evidence spread over 83 minutes of the film and an answer with several claims, the combination \lflvqa targets.

\paragraph{Video models and agents.}
LLoVi~\citep{zhang-etal-2024-simple} showed that a simple pipeline is a strong long-video baseline: a captioner describes short clips, and an LLM answers from the concatenated captions, improving the previous best result on EgoSchema by 18 points.
Agents instead search the video with tools, retrieving and inspecting a few relevant moments before answering~\citep{wang_videoagent_2024}.
Both designs reduce a video to captions or to a few selected moments.
Our text-based baselines follow LLoVi's caption-then-reason design, and we also evaluate an agentic baseline, to test whether these reductions keep the evidence a long-form answer about a character's beliefs needs.

\section{Dataset and Benchmark Design}
\label{sec:dataset}

We want questions whose input and output are both long: answering them needs evidence distributed across the video, and the answer takes more than a short phrase.
Theory-of-mind (ToM) questions have both properties.
Following \citet{jin-etal-2024-mmtom}, we treat a character's belief at a given moment as the product of what they have observed up to that point, which may differ from what is actually true (a false belief).
Answering such a question also takes a set of claims rather than a single phrase.
Long-form answers are hard to score against a single reference answer~\citep{xu-etal-2023-critical}. We therefore follow claim-level evaluation of long-form text, which decomposes an answer into claims that are checked one by one~\citep{min-etal-2023-factscore, song-etal-2024-veriscore}, but change what the claims are checked against.
We pair each question with a \emph{rubric}: a list of claims a correct answer should make, each describing a single event or state and grounded in the screenplay scenes that support it.
Given a model's answer, we check which of the required claims it states (recall) and, of the rubric claims it commits to, the share it states correctly rather than contradicts (precision; Section~\ref{sec:dataset-eval}).
The rest of this section describes how we build these \emph{(question, answer, rubric)} triples. Figure~\ref{fig:belief-example} follows one of them through the construction pipeline, and Table~\ref{tab:question-types} summarizes the eleven question types we generate (Section~\ref{sec:dataset-stats}).

\definecolor{tabblue}{HTML}{4B4DBE}
\providecommand{\anchorbar}[1]{\textcolor{tabblue}{\rule{0.19cm}{#1cm}}}
\providecommand{\anchorbars}[5]{\vrule width 0pt height 0.28cm depth 0.04cm%
  \makebox[0pt][l]{\textcolor{black!30}{\rule[-0.4pt]{1.05cm}{0.4pt}}}%
  \makebox[0pt][l]{\textcolor{black!30}{\rule[0.09cm]{1.05cm}{0.3pt}}}%
  \anchorbar{#1}\hspace{0.02cm}\anchorbar{#2}\hspace{0.02cm}\anchorbar{#3}\hspace{0.02cm}\anchorbar{#4}\hspace{0.02cm}\anchorbar{#5}}
\providecommand{\anchoraxis}{\makebox[1.05cm][s]{\tiny start\hfill end}}
\providecommand{\stackhead}[2][c]{\begin{tabular}[b]{@{}#1@{}}#2\end{tabular}}
\definecolor{figindigo}{HTML}{4B4DBE}\definecolor{figcrimson}{HTML}{C0394F}
\providecommand{\typesq}[1]{\textcolor{#1}{\rule{1.6mm}{1.6mm}}~}
\begin{table}[htb!]
\centering
\small
\setlength{\tabcolsep}{4pt}
\resizebox{.92\linewidth}{!}{%
\begin{tabular*}{\linewidth}{@{\extracolsep{\fill}}lrrrrrrrrc@{}}
\toprule
Question type & $n$ & \stackhead{Core\\claims} & \stackhead{Context\\claims} & \stackhead{Question\\tokens} & \stackhead{Answer\\tokens} & \stackhead{Claim\\tokens} & \stackhead{Cited\\scenes} & \stackhead{Span\\(min)} & \stackhead{Anchor\\\anchoraxis} \\
\midrule
\typesq{figindigo}Q1 Knowledge split & 206 & 3.0 & 1.3 & 23 & 77 & 14 & 3.1 & 7 & \anchorbars{0.07}{0.07}{0.10}{0.12}{0.07} \\
\typesq{figindigo}Q2 Realization & 857 & 1.5 & 1.8 & 20 & 75 & 17 & 4.1 & 24 & \anchorbars{0.06}{0.09}{0.11}{0.11}{0.07} \\
\typesq{figindigo}Q3 False belief & 325 & 1.4 & 2.1 & 20 & 73 & 14 & 3.4 & 12 & \anchorbars{0.08}{0.08}{0.10}{0.11}{0.07} \\
\typesq{figindigo}Q4 Knowledge source & 1,463 & 1.6 & 1.2 & 20 & 69 & 16 & 3.0 & 5 & \anchorbars{0.09}{0.11}{0.11}{0.07}{0.05} \\
\typesq{figindigo}Q5 First-order belief & 128 & 1.3 & 1.3 & 21 & 65 & 16 & 3.2 & 9 & \anchorbars{0.09}{0.08}{0.11}{0.11}{0.06} \\
\typesq{figindigo}Q6 Second-order belief & 112 & 1.2 & 1.2 & 23 & 61 & 17 & 3.0 & 5 & \anchorbars{0.09}{0.09}{0.10}{0.11}{0.05} \\
\typesq{figindigo}Q7 Accepting a claim & 1,151 & 2.0 & 1.3 & 22 & 73 & 16 & 3.0 & 4 & \anchorbars{0.10}{0.11}{0.09}{0.09}{0.04} \\
\typesq{figindigo}Q8 Order of discovery & 535 & 4.1 & 1.1 & 23 & 85 & 16 & 5.7 & 52 & \anchorbars{0.02}{0.04}{0.04}{0.10}{0.24} \\
\midrule
\typesq{figcrimson}N1 Epistemic arc & 589 & 4.7 & 1.9 & 23 & 94 & 16 & 7.9 & 68 & \anchorbars{0.02}{0.02}{0.05}{0.10}{0.24} \\
\typesq{figcrimson}N2 Trust evolution & 475 & 2.5 & 1.6 & 21 & 68 & 15 & 6.4 & 60 & \anchorbars{0.01}{0.03}{0.05}{0.11}{0.24} \\
\typesq{figcrimson}N3 Causal chain & 486 & 4.0 & 1.2 & 22 & 80 & 15 & 3.7 & 15 & \anchorbars{0.10}{0.09}{0.10}{0.08}{0.07} \\
\midrule
All questions & 6,327 & 2.4 & 1.4 & 21 & 75 & 16 & 4.2 & 23 & \anchorbars{0.07}{0.08}{0.09}{0.09}{0.10} \\
\bottomrule
\end{tabular*}
}
\caption{Statistics per question type. \typesq{figindigo}Belief-state types ask about one moment; \typesq{figcrimson}arc types ask about the whole film. All columns except $n$ are averages per question; span is the runtime between the earliest and latest cited scene; the last column shows where questions are anchored.}
\label{tab:question-types}
\end{table}

\subsection{Data Collection and Preprocessing}
\label{sec:dataset-collection}

For 94 films released in 2023 or later, we collect \emph{screenplays}, \emph{subtitles}, and character rosters from public sources (ScriptSlug, OpenSubtitles, and TMDB, respectively).\footnote{\url{https://www.scriptslug.com}, \url{https://www.opensubtitles.com}, \url{https://www.themoviedb.org}}
We restrict to recent releases to reduce the chance that the films, or discussions of them, appear in the training data of the vision--language models we evaluate.

To generate our \emph{(question, answer, rubric)} triples, we use the screenplay as the ground-truth text representation of each film.
We use a language model (Qwen3.5-27B) to split each screenplay into \emph{scenes} with speaker-attributed dialogue and stage directions, summarize every scene, and resolve coreferent mentions of characters.
Finally, we time-stamp each scene by aligning its dialogue with the subtitles, which places every scene, and every claim later built from it, on the film's timeline.
Appendix~\ref{app:dataset-preprocessing-extra} details each preprocessing step with the prompts used.

\paragraph{Character Belief Annotation.}
From each parsed screenplay, we build a per-character \emph{belief record} with the same model, working through the film scene by scene.
We first decompose every scene into atomic \emph{scene events}, each listing its participants and witnesses.
For \emph{first-order beliefs}, we go through the scenes in order for each character, giving the model the scene, its events, and the character's beliefs so far.
It determines what the character perceived firsthand, what they were told and whether they believed it, what they inferred, which of their beliefs are false, and which earlier beliefs they revised; the result is carried into the next scene.
For example, in Figure~\ref{fig:belief-example}, Wolverine accepts two false promises from Deadpool in scene~29 and revises them in scenes~50 and~108, when what he witnesses contradicts them.
For \emph{meta-beliefs}, we again go scene by scene and record what each character $A$ comes to believe about the beliefs of another character $B$, along with how that belief formed.
Details and prompts are in Appendix~\ref{app:dataset-belief-extraction-extra}.

\subsection{Question Generation and Verification}
\label{sec:dataset-generation}

\paragraph{Writing questions from context-sets.}
Every question is built around two choices.
The \emph{center character} is the character whose mind the question asks about.
The \emph{anchor} is the scene that fixes the moment the question refers to.
The question points to it in a few words by a place, an occasion, or an object, never by what happens there (e.g., ``At Wade's apartment''), so a model must first locate that moment in the video and then reconstruct what the center character knew by then.
Each question also has one of eleven \emph{types}, which fixes the kind of epistemic state it asks about, such as how the center character comes to know a fact or how their view of another character changes, and the parts of the belief record it draws on (Section~\ref{sec:dataset-stats}).

We never give the question writer the whole film. Instead, each question is written from a \emph{context-set} for one center character, anchor, and type.
We discard context-sets 
based on rules,
then we sample the remaining context-sets. 
From each context-set, the question writer (gpt-5.6-sol) produces a candidate question and an outline of claims
or declines to write one.
Appendix~\ref{app:dataset-generation-extra} provides the prompts and discard rules.

\paragraph{Rubrics and question verification.}
Because the belief record is extracted automatically, it can be wrong, and so can the claims drafted from it.
We therefore verify every claim against the screenplay itself.
The verifier (gpt-5.6-sol) labels each claim as supported, unsupported, or contradicted by the relevant parts of the screenplay.
We remove the claims the verifier does not support and rewrite those that are only partly supported.
If the surviving claims no longer answer the question, the verifier rewrites the question so that they answer it completely; about one in ten released questions was revised this way, and questions with too few surviving core claims are discarded.
The surviving claims form the question's rubric.
Each claim is either \emph{core}, a claim the question asks for and a complete answer must make, or \emph{context}, a premise an answer may omit but must not contradict.
We then write the reference answer from the rubric as a short paragraph.

\subsection{Evaluation metrics}
\label{sec:dataset-eval}

For a question $q$, let $\mathcal{K}_q$ be its rubric and $\mathcal{C}_q \subseteq \mathcal{K}_q$ its core claims.
Given a model answer $a_q$, an LLM judge (gpt-5.6-luna) labels every rubric claim as \emph{asserted}, \emph{contradicted}, or \emph{omitted} by the answer, judging only against the answer's text (Appendix~\ref{app:prompts}).
Let $\mathcal{A}_q$ and $\mathcal{X}_q$ be the rubric claims the answer asserts and contradicts.
We compute:
\begin{equation}
\mathrm{R}_q = \frac{|\mathcal{A}_q \cap \mathcal{C}_q|}{|\mathcal{C}_q|},
\qquad
\mathrm{P}_q = \frac{|\mathcal{A}_q|}{|\mathcal{A}_q| + |\mathcal{X}_q|},
\qquad
\mathrm{F1}_q = \frac{2\,\mathrm{P}_q\,\mathrm{R}_q}{\mathrm{P}_q + \mathrm{R}_q},
\label{eq:metrics}
\end{equation}
We report each measure averaged over questions.
We also report the contradiction rate $|\mathcal{X}_q| / |\mathcal{K}_q|$.

\subsection{Dataset Statistics}
\label{sec:dataset-stats}

\def\corpustablebody{%
\centering\small\setlength{\tabcolsep}{4pt}%
\begin{tabular}{@{}lrr@{}}
\toprule
 & Overall & Per film \\
\midrule
Films & 94 & -- \\
Runtime & 182\,h & 112\,min \\
Scenes & 11,712 & 118 \\
Questions & 6,327 & 60 \\
Verified claims & 24,496 & 250 \\
Unique Characters & 1,625 & 15.5 \\
Tokens & & \\
-- subtitles & 1.35M & 14K \\
-- screenplay & 2.96M & 31K \\
\bottomrule
\end{tabular}%
\caption{\textbf{Dataset Size}}%

\label{tab:dataset-corpus}}
\ifdefined\wraptable
\begin{wraptable}{r}{0.42\textwidth}\vspace{-10pt}\corpustablebody\vspace{-8pt}\end{wraptable}
\else
\begin{table}[t]\corpustablebody\end{table}
\fi

We define eleven question types by the parts of the belief record they draw on (Table~\ref{tab:question-types}).
Eight ask about the center character at one moment: how several characters' knowledge of the same situation differs (Q1), when their understanding changes (Q2), what they take to be the case right before acting on a false belief (Q3), how they come to know a fact (Q4), what they take another character to believe (Q5) or to believe about a third (Q6), what they make of a claim they are told (Q7), and the order in which they learn a set of facts (Q8).
The other three ask about the center character across the film: what they come to know and how (N1), how their view of another character changes (N2), and the chain of events that leads them to a belief (N3).

\noindent\textbf{Character coverage.} Importantly, the questions follow a film's entire cast and not just its protagonist (Table~\ref{tab:dataset-corpus}
).
A film has a median of 15.5 center characters, and its most-questioned one accounts for only about a fifth of its questions.
Meaning a model that tracks only the main character therefore answers a minority of the dataset.

\noindent\textbf{Scene coverage.} Every question cites at least two scenes and on average 4.2, spanning a mean of 23 minutes of runtime (Table~\ref{tab:question-types}); for arc questions the median span is 44 minutes.
Together, the questions of a film cite a median of 71\% of its scenes, which cover 72\% of its runtime, and their anchors are spread almost uniformly over the film, with each fifth of the runtime holding between 16\% and 23\% of the questions.
No part of a film can therefore be skipped.
Reference answers average 61 words and contain 2.4 core and 1.4 context claims.

\subsection{Human Study -- Dataset Validation}
\label{sec:dataset-human-study}

Two annotators independently assessed each of 150 questions, their reference answers, and 494 associated claims. Both annotators recommended keeping 83.3\% of questions and 85.3\% of reference answers; both components received unanimous keep decisions for 76.7\% of questions. At the claim level, both annotators judged 80.4\% factually supported and 75.5\% supported by sufficient linked evidence. Exact agreement ranged from 80.6\% to 96.6\% across criteria. These results document predominantly favorable assessments of the inspected subset, but limited chance-adjusted agreement constrains conclusions about the reproducibility of those assessments.

\section{Models and Experimental Settings}
\label{sec:experiments}

We construct a diverse set of baselines to isolate the challenges of perception, long-context understanding and evidence acquisition. 
Unless otherwise stated, all models are evaluated zero-shot.


\subsection{Text-based Baselines}

Our first set of baselines removes the complexity of perception by converting each movie into a time-ordered textual representation.
\vspace*{-.7\baselineskip}
\paragraph{Screenplay (oracle).} This oracle receives the movie screenplay -- the ground truth that was used to generate the question in the first place. 
While impractical, it allows us to see if models can extract relevant information from the screenplay without the carefully crafted pipeline used during question generation.

\vspace*{-.7\baselineskip}
\paragraph{Subtitles.} 
This baseline receives the movie's time-aligned subtitle track (and the question). It measures how well the task can be solved from dialogue alone without any visual information.

\vspace*{-.7\baselineskip}
\paragraph{Video captions.} We divide the movie into 10-second chunks and caption each using a video-language model (VLM). We annotate each video with character names to provide character identity to the VLM. The resulting captions are concatenated and provided to the model. This baseline removes dialogue and therefore measures the extent to which questions can be answered from visual information alone. More information can be found in Appendix~\ref{app:identity}.

\vspace*{-.7\baselineskip}
\paragraph{Video captions + subtitles.} This baseline receives the 10-second video captions interleaved with the subtitles corresponding to the same temporal interval.
This condition provides both visual information and dialogue while preserving their temporal organization. 


\subsection{Video-Based Baselines}
The previous baselines may discard visual information before the answering model observes it. To see if question-dependent visual processing is necessary, we consider two video understanding baselines.
\textbf{(1) Video without audio:}
The model receives uniformly sampled  visual input (at 0.2fps) from the movie together with the question, but no audio. 
\textbf{(2) Video without audio + subtitles:}
This baseline receives the same visual input together with the complete time-aligned subtitle track, which provides dialogue.  
This condition evaluates whether the model can answer narrative questions from the visual stream alone.
We run this setting with are Qwen3.6 and Gemma4 as they are some of the few models that can handle video of that length.

\subsection{Agentic approaches}
The preceding baselines receive a fixed representation of the movie in a single model invocation. 
We also evaluate an agentic baseline that gets tools to iteratively search the movie, inspect selected intervals, and accumulate evidence before producing its final answer.
We use Qwen3.6 35B/A3B as the agentic orchestrator. At each step, the model may call one of the following tools:

\begin{itemize}
    \item \texttt{subtitle\_search}: retrieves subtitle segments relevant to a query.
    \item \texttt{find\_clips}: returns the temporal spans of clips that best match a query.
    \item \texttt{vqa}: answers a free-form question about a selected video interval.
    \item \texttt{browse\_clip}: provides a coarse-grained summary of what happens in a video clip.
    \item \texttt{character\_presence}: returns all characters appearing in a specified clip.
\end{itemize}

In addition to the \textbf{regular} agent that uses the above tools, we also consider a \textbf{fast} agent that uses precomputed 10-second video captions as a representation to do approximate search.
At the other extreme, we also evaluate an \textbf{Oracle} agent that gets access to the screenplay.

\definecolor{tabblue}{HTML}{4C8BF5}
\definecolor{tabred}{HTML}{E8453C}
\makeatletter\@ifundefined{NC@find@Z}{\newcolumntype{Z}{>{\centering\arraybackslash}p{1.05cm}}}{}\makeatother
\providecommand{\logogoogle}{\raisebox{-0.2\height}{\includegraphics[height=1em]{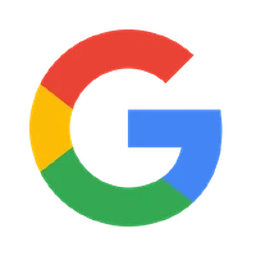}}}
\providecommand{\logokwai}{\raisebox{-0.2\height}{\includegraphics[height=1em]{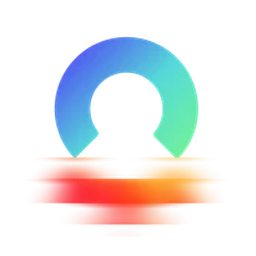}}}
\providecommand{\logoopengvlab}{\raisebox{-0.2\height}{\includegraphics[height=1em]{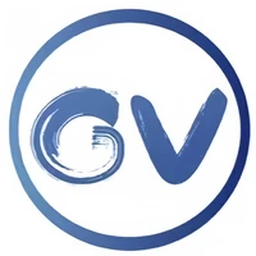}}}
\providecommand{\logoqwen}{\raisebox{-0.2\height}{\includegraphics[height=1em]{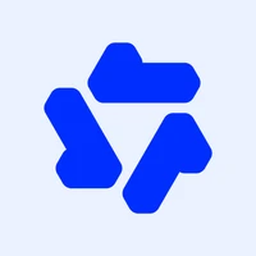}}}
\begin{table*}[t]
\centering
\small
\setlength{\tabcolsep}{5pt}
\renewcommand{\arraystretch}{1.08}
\resizebox{0.94\linewidth}{!}{%
\begin{tabular}{@{}llZZZZZZZZ@{}}
\toprule
 & & \multicolumn{4}{c}{Lexical} & \multicolumn{4}{c}{LLM judge} \\
\cmidrule(lr){3-6}\cmidrule(l){7-10}
Source & Model & RL & ClmRL & Cov@.5 & Hedge$\downarrow$ & Rec & Prec & F1 & Contra$\downarrow$ \\
\midrule
\multicolumn{2}{@{}l}{\textit{Gold answer}} & 1.000 & .686 & .975 & .007 & -- & -- & -- & -- \\
\midrule
\rowcolor{gray!12}\multicolumn{10}{@{}l}{\textsc{Text-based}} \\
\addlinespace[1pt]
\multirow{5}{*}{Screenplay} & \logoqwen\,Qwen3.6\,{\footnotesize\color{gray}35B/A3B} & .160 & \cellcolor{tabblue!21!white}\bf .248 & \cellcolor{tabblue!70!white}\bf .348 & .343 & \cellcolor{tabblue!50!white}.472 & \cellcolor{tabblue!50!white}.667 & \cellcolor{tabblue!50!white}.505 & \bf .128 \\
 & \logoqwen\,Qwen3.8\,{\footnotesize\color{gray}27B}$^{\star}$ & \cellcolor{tabblue!21!white}\bf .170 & \cellcolor{tabblue!10!white}.243 & \cellcolor{tabblue!50!white}.325 & .216 & \cellcolor{tabblue!70!white}\bf .509 & \cellcolor{tabblue!70!white}\bf .703 & \cellcolor{tabblue!70!white}\bf .544 & .103 \\
 & \logogoogle\,Gemma4\,{\footnotesize\color{gray}26B/A4B} & .164 & .220 & \cellcolor{tabblue!34!white}.287 & \bf .519 & \cellcolor{tabblue!10!white}.260 & \cellcolor{tabblue!21!white}.464 & \cellcolor{tabblue!10!white}.290 & .100 \\
 & \logokwai\,Keye-VL-1.5\,{\footnotesize\color{gray}8B} & .140 & .189 & .193 & .008 & .143 & .256 & .162 & .060 \\
 & \logoopengvlab\,InternVL3.5\,{\footnotesize\color{gray}8B} & .142 & .198 & .165 & .211 & .072 & .120 & .081 & .016 \\
\cmidrule(lr){2-10}
\multirow{5}{*}{Subtitles} & \logoqwen\,Qwen3.6\,{\footnotesize\color{gray}35B/A3B} & .141 & \bf .209 & .252 & \cellcolor{tabred!21!white}.770 & .149 & .261 & .165 & \cellcolor{tabred!10!white}.145 \\
 & \logoqwen\,Qwen3.8\,{\footnotesize\color{gray}27B}$^{\star}$ & \bf .150 & .206 & \cellcolor{tabblue!10!white}\bf .257 & .573 & \bf .157 & \bf .275 & \bf .175 & \cellcolor{tabred!21!white}\bf .163 \\
 & \logogoogle\,Gemma4\,{\footnotesize\color{gray}26B/A4B} & .139 & .182 & .206 & \cellcolor{tabred!70!white}\bf .937 & .055 & .128 & .064 & .061 \\
 & \logokwai\,Keye-VL-1.5\,{\footnotesize\color{gray}8B} & .138 & .183 & .160 & .009 & .087 & .166 & .100 & .080 \\
 & \logoopengvlab\,InternVL3.5\,{\footnotesize\color{gray}8B} & .138 & .188 & .146 & .271 & .005 & .011 & .006 & .005 \\
\cmidrule(lr){2-10}
\multirow{5}{*}{Video Captioning} & \logoqwen\,Qwen3.6\,{\footnotesize\color{gray}35B/A3B} & .134 & .188 & \bf .211 & \cellcolor{tabred!34!white}.891 & .013 & .041 & .016 & .078 \\
 & \logoqwen\,Qwen3.8\,{\footnotesize\color{gray}27B} & .126 & .180 & .194 & .544 & .003 & .016 & .004 & .024 \\
 & \logogoogle\,Gemma4\,{\footnotesize\color{gray}26B/A4B} & .123 & .167 & .183 & \cellcolor{tabred!50!white}\bf .930 & .001 & .007 & .002 & .017 \\
 & \logokwai\,Keye-VL-1.5\,{\footnotesize\color{gray}8B}$^{\star}$ & \bf .144 & \bf .212 & .196 & .120 & \bf .062 & \bf .130 & \bf .071 & \bf .097 \\
 & \logoopengvlab\,InternVL3.5\,{\footnotesize\color{gray}8B} & .137 & .186 & .147 & .235 & .001 & .002 & .001 & .001 \\
\cmidrule(lr){2-10}
\multirow{5}{*}{Video Captioning + Subtitles} & \logoqwen\,Qwen3.6\,{\footnotesize\color{gray}35B/A3B}$^{\star}$ & \bf .155 & \bf .229 & \cellcolor{tabblue!21!white}\bf .283 & .577 & \bf .222 & \bf .368 & \bf .244 & \cellcolor{tabred!50!white}\bf .169 \\
 & \logoqwen\,Qwen3.8\,{\footnotesize\color{gray}27B} & .133 & .197 & .224 & .307 & .099 & .182 & .112 & .056 \\
 & \logogoogle\,Gemma4\,{\footnotesize\color{gray}26B/A4B} & .132 & .179 & .177 & \bf .641 & .031 & .065 & .036 & .021 \\
 & \logokwai\,Keye-VL-1.5\,{\footnotesize\color{gray}8B} & .148 & .217 & .205 & .102 & .076 & .151 & .087 & .114 \\
 & \logoopengvlab\,InternVL3.5\,{\footnotesize\color{gray}8B} & .137 & .187 & .146 & .234 & .004 & .008 & .004 & .002 \\
\midrule
\rowcolor{gray!12}\multicolumn{10}{@{}l}{\textsc{Video-based}} \\
\addlinespace[1pt]
\multirow{2}{*}{Video w/o Audio} & \logoqwen\,Qwen3.6\,{\footnotesize\color{gray}35B/A3B}$^{\star}$ & \bf .145 & \bf .191 & .101 & .012 & \bf .086 & \bf .172 & \bf .099 & \bf .099 \\
 & \logogoogle\,Gemma4\,{\footnotesize\color{gray}26B/A4B} & .112 & .144 & \bf .102 & \cellcolor{tabred!10!white}\bf .756 & .002 & .009 & .003 & .062 \\
\cmidrule(lr){2-10}
\multirow{2}{*}{Video w/o Audio + Subtitles} & \logoqwen\,Qwen3.6\,{\footnotesize\color{gray}35B/A3B}$^{\star}$ & \cellcolor{tabblue!10!white}\bf .167 & \bf .242 & \bf .257 & .118 & \cellcolor{tabblue!21!white}\bf .279 & \cellcolor{tabblue!10!white}\bf .452 & \cellcolor{tabblue!21!white}\bf .304 & \cellcolor{tabred!34!white}.167 \\
 & \logogoogle\,Gemma4\,{\footnotesize\color{gray}26B/A4B} & .146 & .193 & .159 & \bf .692 & .062 & .135 & .071 & \cellcolor{tabred!70!white}\bf .226 \\
\midrule
\rowcolor{gray!12}\multicolumn{10}{@{}l}{\textsc{Agentic}} \\
\addlinespace[1pt]
\multirow{3}{*}{\logoqwen\,Qwen3.6\,{\footnotesize\color{gray}35B/A3B}} & Oracle$^{\star}$ & \cellcolor{tabblue!70!white}\bf .202 & \cellcolor{tabblue!70!white}\bf .290 & \bf .194 & .124 & \cellcolor{tabblue!34!white}\bf .302 & \cellcolor{tabblue!34!white}\bf .484 & \cellcolor{tabblue!34!white}\bf .332 & .093 \\
 & Fast & \cellcolor{tabblue!50!white}.190 & \cellcolor{tabblue!50!white}.276 & .163 & .171 & .208 & .352 & .229 & .113 \\
 & Regular & \cellcolor{tabblue!34!white}.186 & \cellcolor{tabblue!34!white}.270 & .150 & \bf .197 & .181 & .312 & .201 & \bf .115 \\
\bottomrule
\end{tabular}}

\caption{\textbf{Results on \lflvqa.} Rows are answerer models; in the video-captioning answerer also write the captions. Rec, Prec, F1, and Contra follow Section~\ref{sec:dataset-eval}. Blue: five best values across all runs; red: five worst values of Hedge and Contra. \textbf{Bold}: highest value within a source.}
\label{tab:results-allp}
\end{table*}

\section{Results and Discussion}

Table \ref{tab:results-allp} reports performance across text-based, video-based, and agentic settings. We report four lexical measures—RL, ClmRL, Cov@.5, and Hedge.  Additionally, we also report recall, precision, F1, and contradiction rate from the LLM-based evaluator. Higher values are better for all metrics except Hedge and contradiction.

\paragraph{Screenplay-based Oracles Are Still Not Perfect}
As one might expect, oracles using the screenplay setting produces the strongest results by a substantial margin:  Qwen3.8 27B obtains the highest LLM-judge F1 score of 0.544, with a recall of 0.509 and precision of 0.703. 
This is not surprising, since screenplays include not just dialogue but also actions, settings and other narratively important information. Nevertheless, even the best screenplay-based model attains only 0.509 recall, leaving nearly half of the evaluated answer content unrecovered.
This suggests that recovering and synthesizing evidence from long context is still a challenge.

\paragraph{Subtitles Alone Are Insufficient for Narrative Understanding}

Replacing the screenplay with subtitles causes a large performance decline: for example, for Qwen3.8, LLM-judge F1 falls from 0.544 with the screenplay to 0.175 with subtitles. 
Claim coverage also decreases from 0.348 to 0.252 for Qwen3.6 and from 0.325 to 0.257 for Qwen3.8.

This gap shows that access to the complete dialogue does not provide a complete representation of the narrative. Many answers depend on actions, reactions, settings, and causal connections that are not explicitly verbalized. 


\paragraph{Visual Information Is Most Useful When Combined With Language}

Conversely, automatically generated video captions perform poorly when used as the only movie representation. The strongest caption-only system, Keye-VL-1.5, obtains an LLM-judge F1 of 0.071. These results suggest that local visual descriptions alone do not preserve enough information to reconstruct the broader narrative.

Combining video captions with subtitles substantially improves Qwen3.6. Its F1 increases from 0.016 with captions alone and 0.165 with subtitles alone to 0.244 when the two are combined. Recall rises to 0.222 and precision to 0.368. This improvement shows that the two representations provide complementary information: subtitles capture spoken content, while captions introduce visible actions and settings.

The gain is not universal, however. Qwen3.8 obtains an F1 of 0.112 with video captions and subtitles, below its subtitle-only score of 0.175. 
Thus, providing additional information does not guarantee that a model will use it effectively. The model must still identify relevant observations and integrate them with the dialogue without being distracted by noisy or redundant descriptions.

Direct video input follows a similar pattern, where combining audio with video yields much better performance than video alone (0.304 F1 with subtitles vs 0.099 without for Qwen3.6).
The direct-video result is stronger than the corresponding caption-based result for Qwen3.6: video with subtitles obtains 0.304 F1, compared with 0.244 for video captions with subtitles. This suggests that converting the video into textual descriptions discards information that the multimodal model can recover from the original visual input. At the same time, all these accuracies lag significantly behind the screenplay-based oracles, suggesting that models struggle to extract all relevant information.

\paragraph{Oracle Evidence Improves Agentic Performance}

The agentic experiments evaluate whether iterative evidence search can overcome the limitations of fixed-context inputs. Using Qwen3.6 as the orchestrator, the Regular agent obtains 0.181 recall, 0.312 precision, and 0.201 F1. The Fast configuration improves these scores to 0.208 recall, 0.352 precision, and 0.229 F1. This is still substantially worse than simply providing the video and subtitles as input to Qwen3.6 (0.304 F1).

The Oracle agent performs substantially better, reaching 0.302 recall, 0.484 precision, and 0.332 F1. 
This result confirms that evidence localization is an important bottleneck: the agent answers substantially more of the question when it can retrieve higher-quality evidence.

However, Oracle remains well below the screenplay-based Qwen3.6 result, obtaining 0.332 F1 compared with 0.505. Oracle retrieval therefore does not eliminate the difficulty of the task. Even when the agent receives improved evidence, it must still determine which observations matter, connect events across the movie, and synthesize them into a complete answer.

\definecolor{tabblue}{HTML}{4C8BF5}
\providecommand{\logogoogle}{\raisebox{-0.2\height}{\includegraphics[height=1em]{figures/logos/google.png}}}
\providecommand{\logokwai}{\raisebox{-0.2\height}{\includegraphics[height=1em]{figures/logos/kwai.png}}}
\providecommand{\logoopengvlab}{\raisebox{-0.2\height}{\includegraphics[height=1em]{figures/logos/opengvlab.png}}}
\providecommand{\logoqwen}{\raisebox{-0.2\height}{\includegraphics[height=1em]{figures/logos/qwen.png}}}
\newcolumntype{Y}{>{\centering\arraybackslash}p{1.02cm}}
\begin{table*}[t]
\centering
\small
\setlength{\tabcolsep}{3pt}
\renewcommand{\arraystretch}{1.08}
\resizebox{\linewidth}{!}{%
\begin{tabular}{@{}llYYYYYYYY|YYY|Y@{}}
\toprule
 & & \multicolumn{8}{c}{Belief \& knowledge} & \multicolumn{3}{c}{Narrative} & \\
\cmidrule(lr){3-10}\cmidrule(lr){11-13}
Context & Captioner / tier & Q1 & Q2 & Q3 & Q4 & Q5 & Q6 & Q7 & Q8 & N1 & N2 & N3 & All \\
\midrule
Screenplay & \textit{--} & \cellcolor{tabblue!70!white}\textbf{.64}\,{\tiny\color{gray}$\pm$\,.05} & \cellcolor{tabblue!70!white}\textbf{.39}\,{\tiny\color{gray}$\pm$\,.03} & \cellcolor{tabblue!70!white}\textbf{.61}\,{\tiny\color{gray}$\pm$\,.04} & \cellcolor{tabblue!70!white}\textbf{.60}\,{\tiny\color{gray}$\pm$\,.03} & \cellcolor{tabblue!50!white}.31\,{\tiny\color{gray}$\pm$\,.10} & \cellcolor{tabblue!34!white}.20\,{\tiny\color{gray}$\pm$\,.07} & \cellcolor{tabblue!70!white}\textbf{.61}\,{\tiny\color{gray}$\pm$\,.02} & \cellcolor{tabblue!70!white}\textbf{.33}\,{\tiny\color{gray}$\pm$\,.03} & \cellcolor{tabblue!70!white}\textbf{.35}\,{\tiny\color{gray}$\pm$\,.03} & \cellcolor{tabblue!70!white}\textbf{.49}\,{\tiny\color{gray}$\pm$\,.04} & \cellcolor{tabblue!70!white}\textbf{.55}\,{\tiny\color{gray}$\pm$\,.03} & \cellcolor{tabblue!70!white}\textbf{.50}\,{\tiny\color{gray}$\pm$\,.01} \\
\cmidrule(l){2-14}
Subtitles & \textit{--} & .22\,{\tiny\color{gray}$\pm$\,.04} & .10\,{\tiny\color{gray}$\pm$\,.02} & .27\,{\tiny\color{gray}$\pm$\,.04} & .18\,{\tiny\color{gray}$\pm$\,.02} & .09\,{\tiny\color{gray}$\pm$\,.05} & .09\,{\tiny\color{gray}$\pm$\,.05} & .20\,{\tiny\color{gray}$\pm$\,.03} & .13\,{\tiny\color{gray}$\pm$\,.02} & .16\,{\tiny\color{gray}$\pm$\,.02} & .16\,{\tiny\color{gray}$\pm$\,.03} & .14\,{\tiny\color{gray}$\pm$\,.03} & .16\,{\tiny\color{gray}$\pm$\,.01} \\
\cmidrule(l){2-14}
\multirow{5}{*}{Video Captioning} & \logoqwen\,Qwen3.6\,{\footnotesize\color{gray}35B/A3B} & .02\,{\tiny\color{gray}$\pm$\,.01} & .01\,{\tiny\color{gray}$\pm$\,.01} & .00\,{\tiny\color{gray}$\pm$\,.01} & .01\,{\tiny\color{gray}$\pm$\,.01} & .00\,{\tiny\color{gray}$\pm$\,.00} & .00\,{\tiny\color{gray}$\pm$\,.00} & .01\,{\tiny\color{gray}$\pm$\,.00} & .02\,{\tiny\color{gray}$\pm$\,.01} & .04\,{\tiny\color{gray}$\pm$\,.01} & .03\,{\tiny\color{gray}$\pm$\,.01} & .03\,{\tiny\color{gray}$\pm$\,.01} & .02\,{\tiny\color{gray}$\pm$\,.00} \\
 & \logoqwen\,Qwen3.8\,{\footnotesize\color{gray}27B} & .01\,{\tiny\color{gray}$\pm$\,.01} & .01\,{\tiny\color{gray}$\pm$\,.01} & .01\,{\tiny\color{gray}$\pm$\,.01} & .01\,{\tiny\color{gray}$\pm$\,.01} & .00\,{\tiny\color{gray}$\pm$\,.00} & .00\,{\tiny\color{gray}$\pm$\,.00} & .00\,{\tiny\color{gray}$\pm$\,.00} & .01\,{\tiny\color{gray}$\pm$\,.01} & .02\,{\tiny\color{gray}$\pm$\,.01} & .02\,{\tiny\color{gray}$\pm$\,.01} & .02\,{\tiny\color{gray}$\pm$\,.01} & .01\,{\tiny\color{gray}$\pm$\,.00} \\
 & \logogoogle\,Gemma4\,{\footnotesize\color{gray}26B/A4B} & .01\,{\tiny\color{gray}$\pm$\,.01} & .00\,{\tiny\color{gray}$\pm$\,.01} & .00\,{\tiny\color{gray}$\pm$\,.00} & .01\,{\tiny\color{gray}$\pm$\,.01} & .00\,{\tiny\color{gray}$\pm$\,.00} & .00\,{\tiny\color{gray}$\pm$\,.00} & .00\,{\tiny\color{gray}$\pm$\,.00} & .01\,{\tiny\color{gray}$\pm$\,.01} & .01\,{\tiny\color{gray}$\pm$\,.01} & .02\,{\tiny\color{gray}$\pm$\,.01} & .01\,{\tiny\color{gray}$\pm$\,.01} & .01\,{\tiny\color{gray}$\pm$\,.00} \\
 & \logokwai\,Keye-VL-1.5\,{\footnotesize\color{gray}8B} & .00\,{\tiny\color{gray}$\pm$\,.00} & .00\,{\tiny\color{gray}$\pm$\,.00} & .00\,{\tiny\color{gray}$\pm$\,.00} & .01\,{\tiny\color{gray}$\pm$\,.00} & .00\,{\tiny\color{gray}$\pm$\,.00} & .00\,{\tiny\color{gray}$\pm$\,.00} & .00\,{\tiny\color{gray}$\pm$\,.00} & .01\,{\tiny\color{gray}$\pm$\,.00} & .01\,{\tiny\color{gray}$\pm$\,.00} & .01\,{\tiny\color{gray}$\pm$\,.01} & .01\,{\tiny\color{gray}$\pm$\,.01} & .00\,{\tiny\color{gray}$\pm$\,.00} \\
 & \logoopengvlab\,InternVL3.5\,{\footnotesize\color{gray}8B} & .00\,{\tiny\color{gray}$\pm$\,.00} & .00\,{\tiny\color{gray}$\pm$\,.00} & .00\,{\tiny\color{gray}$\pm$\,.00} & .01\,{\tiny\color{gray}$\pm$\,.01} & .00\,{\tiny\color{gray}$\pm$\,.00} & .00\,{\tiny\color{gray}$\pm$\,.00} & .00\,{\tiny\color{gray}$\pm$\,.00} & .01\,{\tiny\color{gray}$\pm$\,.01} & .01\,{\tiny\color{gray}$\pm$\,.01} & .01\,{\tiny\color{gray}$\pm$\,.01} & .01\,{\tiny\color{gray}$\pm$\,.00} & .00\,{\tiny\color{gray}$\pm$\,.00} \\
\cmidrule(l){2-14}
\multirow{5}{*}{\shortstack[l]{Video Captioning\\+ Subtitles}} & \logoqwen\,Qwen3.6\,{\footnotesize\color{gray}35B/A3B} & \cellcolor{tabblue!21!white}.33\,{\tiny\color{gray}$\pm$\,.05} & \cellcolor{tabblue!10!white}.16\,{\tiny\color{gray}$\pm$\,.02} & .30\,{\tiny\color{gray}$\pm$\,.05} & .26\,{\tiny\color{gray}$\pm$\,.03} & \cellcolor{tabblue!10!white}.21\,{\tiny\color{gray}$\pm$\,.07} & .09\,{\tiny\color{gray}$\pm$\,.05} & \cellcolor{tabblue!21!white}.32\,{\tiny\color{gray}$\pm$\,.02} & \cellcolor{tabblue!34!white}.19\,{\tiny\color{gray}$\pm$\,.03} & \cellcolor{tabblue!34!white}.22\,{\tiny\color{gray}$\pm$\,.03} & \cellcolor{tabblue!21!white}.26\,{\tiny\color{gray}$\pm$\,.04} & \cellcolor{tabblue!34!white}.21\,{\tiny\color{gray}$\pm$\,.03} & \cellcolor{tabblue!21!white}.24\,{\tiny\color{gray}$\pm$\,.02} \\
 & \logoqwen\,Qwen3.8\,{\footnotesize\color{gray}27B} & \cellcolor{tabblue!10!white}.28\,{\tiny\color{gray}$\pm$\,.05} & \cellcolor{tabblue!21!white}.17\,{\tiny\color{gray}$\pm$\,.03} & \cellcolor{tabblue!10!white}.31\,{\tiny\color{gray}$\pm$\,.05} & .24\,{\tiny\color{gray}$\pm$\,.03} & \cellcolor{tabblue!21!white}.21\,{\tiny\color{gray}$\pm$\,.06} & .14\,{\tiny\color{gray}$\pm$\,.06} & .30\,{\tiny\color{gray}$\pm$\,.03} & \cellcolor{tabblue!10!white}.17\,{\tiny\color{gray}$\pm$\,.02} & \cellcolor{tabblue!21!white}.22\,{\tiny\color{gray}$\pm$\,.03} & .21\,{\tiny\color{gray}$\pm$\,.04} & \cellcolor{tabblue!10!white}.20\,{\tiny\color{gray}$\pm$\,.03} & \cellcolor{tabblue!10!white}.23\,{\tiny\color{gray}$\pm$\,.02} \\
 & \logogoogle\,Gemma4\,{\footnotesize\color{gray}26B/A4B} & .28\,{\tiny\color{gray}$\pm$\,.05} & .14\,{\tiny\color{gray}$\pm$\,.02} & .30\,{\tiny\color{gray}$\pm$\,.04} & .23\,{\tiny\color{gray}$\pm$\,.03} & .15\,{\tiny\color{gray}$\pm$\,.06} & .13\,{\tiny\color{gray}$\pm$\,.06} & .28\,{\tiny\color{gray}$\pm$\,.03} & .16\,{\tiny\color{gray}$\pm$\,.03} & \cellcolor{tabblue!10!white}.20\,{\tiny\color{gray}$\pm$\,.03} & \cellcolor{tabblue!10!white}.22\,{\tiny\color{gray}$\pm$\,.03} & .20\,{\tiny\color{gray}$\pm$\,.03} & .22\,{\tiny\color{gray}$\pm$\,.01} \\
 & \logokwai\,Keye-VL-1.5\,{\footnotesize\color{gray}8B} & .18\,{\tiny\color{gray}$\pm$\,.04} & .09\,{\tiny\color{gray}$\pm$\,.02} & .21\,{\tiny\color{gray}$\pm$\,.05} & .12\,{\tiny\color{gray}$\pm$\,.02} & .11\,{\tiny\color{gray}$\pm$\,.06} & .13\,{\tiny\color{gray}$\pm$\,.07} & .18\,{\tiny\color{gray}$\pm$\,.03} & .12\,{\tiny\color{gray}$\pm$\,.02} & .13\,{\tiny\color{gray}$\pm$\,.02} & .11\,{\tiny\color{gray}$\pm$\,.03} & .09\,{\tiny\color{gray}$\pm$\,.03} & .13\,{\tiny\color{gray}$\pm$\,.01} \\
 & \logoopengvlab\,InternVL3.5\,{\footnotesize\color{gray}8B} & .28\,{\tiny\color{gray}$\pm$\,.05} & .14\,{\tiny\color{gray}$\pm$\,.02} & .29\,{\tiny\color{gray}$\pm$\,.05} & .20\,{\tiny\color{gray}$\pm$\,.03} & .17\,{\tiny\color{gray}$\pm$\,.07} & \cellcolor{tabblue!10!white}.15\,{\tiny\color{gray}$\pm$\,.06} & .27\,{\tiny\color{gray}$\pm$\,.03} & .16\,{\tiny\color{gray}$\pm$\,.03} & .19\,{\tiny\color{gray}$\pm$\,.03} & .17\,{\tiny\color{gray}$\pm$\,.03} & .15\,{\tiny\color{gray}$\pm$\,.02} & .20\,{\tiny\color{gray}$\pm$\,.02} \\
\cmidrule(l){2-14}
Video w/o Audio & \textit{--} & .10\,{\tiny\color{gray}$\pm$\,.03} & .06\,{\tiny\color{gray}$\pm$\,.02} & .16\,{\tiny\color{gray}$\pm$\,.04} & .08\,{\tiny\color{gray}$\pm$\,.02} & .12\,{\tiny\color{gray}$\pm$\,.06} & .10\,{\tiny\color{gray}$\pm$\,.07} & .17\,{\tiny\color{gray}$\pm$\,.02} & .05\,{\tiny\color{gray}$\pm$\,.02} & .08\,{\tiny\color{gray}$\pm$\,.02} & .14\,{\tiny\color{gray}$\pm$\,.04} & .05\,{\tiny\color{gray}$\pm$\,.02} & .10\,{\tiny\color{gray}$\pm$\,.01} \\
\cmidrule(l){2-14}
Video w/o Audio + Subtitles & \textit{--} & \cellcolor{tabblue!50!white}.37\,{\tiny\color{gray}$\pm$\,.06} & \cellcolor{tabblue!34!white}.20\,{\tiny\color{gray}$\pm$\,.04} & \cellcolor{tabblue!50!white}.44\,{\tiny\color{gray}$\pm$\,.06} & \cellcolor{tabblue!34!white}.33\,{\tiny\color{gray}$\pm$\,.03} & \cellcolor{tabblue!70!white}\textbf{.35}\,{\tiny\color{gray}$\pm$\,.09} & \cellcolor{tabblue!70!white}\textbf{.28}\,{\tiny\color{gray}$\pm$\,.10} & \cellcolor{tabblue!50!white}.39\,{\tiny\color{gray}$\pm$\,.03} & \cellcolor{tabblue!50!white}.19\,{\tiny\color{gray}$\pm$\,.03} & \cellcolor{tabblue!50!white}.25\,{\tiny\color{gray}$\pm$\,.03} & \cellcolor{tabblue!50!white}.33\,{\tiny\color{gray}$\pm$\,.05} & \cellcolor{tabblue!21!white}.20\,{\tiny\color{gray}$\pm$\,.03} & \cellcolor{tabblue!34!white}.30\,{\tiny\color{gray}$\pm$\,.02} \\
\cmidrule(l){2-14}
\multirow{3}{*}{\logoqwen\,Qwen3.6\,{\footnotesize\color{gray}35B/A3B}} & Oracle & \cellcolor{tabblue!34!white}.34\,{\tiny\color{gray}$\pm$\,.05} & \cellcolor{tabblue!50!white}.28\,{\tiny\color{gray}$\pm$\,.03} & \cellcolor{tabblue!34!white}.42\,{\tiny\color{gray}$\pm$\,.05} & \cellcolor{tabblue!50!white}.46\,{\tiny\color{gray}$\pm$\,.03} & \cellcolor{tabblue!34!white}.29\,{\tiny\color{gray}$\pm$\,.09} & \cellcolor{tabblue!50!white}.22\,{\tiny\color{gray}$\pm$\,.08} & \cellcolor{tabblue!34!white}.39\,{\tiny\color{gray}$\pm$\,.03} & \cellcolor{tabblue!21!white}.18\,{\tiny\color{gray}$\pm$\,.02} & .17\,{\tiny\color{gray}$\pm$\,.02} & \cellcolor{tabblue!34!white}.29\,{\tiny\color{gray}$\pm$\,.04} & \cellcolor{tabblue!50!white}.26\,{\tiny\color{gray}$\pm$\,.03} & \cellcolor{tabblue!50!white}.33\,{\tiny\color{gray}$\pm$\,.02} \\
 & Fast & .22\,{\tiny\color{gray}$\pm$\,.04} & .16\,{\tiny\color{gray}$\pm$\,.03} & \cellcolor{tabblue!21!white}.35\,{\tiny\color{gray}$\pm$\,.05} & \cellcolor{tabblue!21!white}.29\,{\tiny\color{gray}$\pm$\,.03} & .20\,{\tiny\color{gray}$\pm$\,.07} & \cellcolor{tabblue!21!white}.16\,{\tiny\color{gray}$\pm$\,.07} & \cellcolor{tabblue!10!white}.31\,{\tiny\color{gray}$\pm$\,.02} & .13\,{\tiny\color{gray}$\pm$\,.02} & .14\,{\tiny\color{gray}$\pm$\,.02} & .19\,{\tiny\color{gray}$\pm$\,.03} & .16\,{\tiny\color{gray}$\pm$\,.03} & .23\,{\tiny\color{gray}$\pm$\,.02} \\
 & Regular & .21\,{\tiny\color{gray}$\pm$\,.04} & .13\,{\tiny\color{gray}$\pm$\,.02} & .29\,{\tiny\color{gray}$\pm$\,.05} & \cellcolor{tabblue!10!white}.26\,{\tiny\color{gray}$\pm$\,.03} & .19\,{\tiny\color{gray}$\pm$\,.07} & .13\,{\tiny\color{gray}$\pm$\,.06} & .28\,{\tiny\color{gray}$\pm$\,.03} & .11\,{\tiny\color{gray}$\pm$\,.02} & .12\,{\tiny\color{gray}$\pm$\,.02} & .16\,{\tiny\color{gray}$\pm$\,.03} & .13\,{\tiny\color{gray}$\pm$\,.03} & .20\,{\tiny\color{gray}$\pm$\,.01} \\
\bottomrule
\end{tabular}}
\caption{\textbf{Qwen3.6 across question types.} The answerer is always Qwen3.6; rows vary the source and the captioner. Cells are F1 (w/ 95\% bootstrap interval). Blue: five highest per question type, darker for higher; \textbf{bold}: highest.}

\label{tab:qtype-f1}
\end{table*}

\section{Conclusion}
We introduce a benchmark for evaluating whether models can understand feature-length narrative videos and produce detailed, open-ended answers. Unlike settings centered on short clips, localized evidence, or brief responses, our task requires models to track characters and events across an entire movie, reason about implicit beliefs, motivations, and relationships, and synthesize multiple pieces of evidence into a coherent explanation. To evaluate these responses beyond surface-level similarity, the dataset provides question-specific rubrics that measure both the completeness and faithfulness of generated answers.

Our experiments reveal limitations at multiple stages of current long-video systems. Models perform best when movies are represented by narratively structured screenplays, while performance declines substantially with subtitles, generated video captions, and direct video input. Combining visual information with subtitles recovers some missing information, but does not close this representation gap. Agentic evidence search also helps: oracle evidence substantially improves performance over regular retrieval, demonstrating that evidence localization is an important bottleneck. However, the remaining gap under oracle access shows that retrieval alone is insufficient. Models continue to struggle to integrate temporally distributed evidence and express all components of a long-form answer.
These findings suggest that feature-length video understanding is not solely a context-length problem. Progress will require better preservation of narratively important visual information, more reliable evidence localization, stronger character and event tracking, and models capable of synthesizing retrieved observations into complete, grounded explanations. We hope this benchmark provides a foundation for studying these challenges and developing systems that can understand extended multimodal narratives rather than merely recognize isolated moments.

\subsection*{AI use statement}
\label{sec:ai-use}

In this work, we used generative AI tools to generate synthetic data sets. 
The questions, reference answers, and rubrics of \lflvqa{} are generated with OpenAI models: gpt-5.6-sol writes them and verifies each claim against the screenplay, and gpt-5.6-luna checks the reference answers and serves as the LLM judge (Sections~\ref{sec:dataset-generation} and~\ref{sec:dataset-eval}).
Qwen3.5-27B parses the screenplays into scenes and builds the belief records (Section~\ref{sec:dataset-collection}).
All prompts are given in Appendices~\ref{app:prompts}.
We used Claude Code (Anthropic) as a coding assistant to implement the construction pipeline and the experiments.

We have \textbf{not} used generative AI tools to propose or design pipelines, experiments and analysis: the research questions, the dataset design, the experiments, and the conclusions are our own.
Additionally, we used generative AI tools (Claude) to create and edit software code, to build the prompt and example tables in the appendix, and to edit the text of the paper for readability.
We have reviewed all AI-assisted work.
Two annotators independently reviewed 150 generated questions together with their reference answers and claims (Section~\ref{sec:dataset-human-study}).
We reviewed and tested all AI-assisted code, checked the generated tables against their source files, and checked every edit to the text.

\subsection*{Ethics statement}
\lflvqa{} is a question-answering dataset on films.
We do not release or redistribute the videos, screenplays, subtitles, or cast photos.
Instead, we provide scripts that let users download the screenplays, subtitles, and cast photos from the public sources we collected them from (Section~\ref{sec:dataset-collection}); the scripts do not cover the videos, and we ask users to obtain them legally.
All of this work is for non-commercial research use only.

\bibliography{iclr2027_conference}
\bibliographystyle{iclr2027_conference}

\newpage

\appendix
\section{Comparison with Existing Benchmarks}
\label{app:benchmark-comparison}

Table~\ref{tab:benchmark-comparison} compares the lengths of questions and answers in \lflvqa with those of existing video QA benchmarks.
Among video benchmarks that report answer length, \lflvqa answers are the longest: 61 words, against at most 29.5 in Neptune and 17.5 among film benchmarks (MoVQA).
The benchmarks that ask open-ended questions about full-length films either report short answers (19--27 tokens in DeepMovieQA) or do not report answer length, and all of them grade an answer as a whole.
\lflvqa is the only benchmark here whose answers are question-specific explanations about full-length films, graded claim by claim for both recall and precision.

\providecommand{\yes}{\textcolor{green!55!black}{$\checkmark$}}
\providecommand{\no}{\textcolor{black!35}{$\times$}}
\providecommand{\partly}{\textcolor{orange!80!black}{$\sim$}}
\begin{table}[h]
\centering
\small
\setlength{\tabcolsep}{5pt}
\renewcommand{\arraystretch}{1.1}
\resizebox{\linewidth}{!}{%
\begin{tabular}{@{}llcccrr@{}}
\toprule
Benchmark & Video length & Full films & Open-ended & ToM & Q len & A len \\
\midrule
\multicolumn{7}{@{}l}{\textsc{General long video}} \\
ActivityNet-QA~\citep{yu_activitynetqa_2019} & 3 min & \no & \yes & \no & 8.7 & 1.9 \\
NExT-QA~\citep{Xiao2021NExTQANP} & 44 s & \no & \yes & \no & 11.6 & 2.6 \\
EgoSchema~\citep{mangalam_egoschema_2023} & 3 min & \no & \no & \no & -- & -- \\
Video-MME~\citep{fu2024videomme} & 17 min & \no & \no & \no & 12.7 & 5.5 \\
LongVideoBench~\citep{wu_longvideobench_2024} & 7.9 min & \no & \no & \no & 43.5 & 8.3 \\
MLVU~\citep{zhou2025mlvubenchmarkingmultitasklong} & 15.5 min & \no & \partly & \no & -- & -- \\
LVBench~\citep{wang2024lvbench} & 68 min & \no & \no & \no & -- & -- \\
HourVideo~\citep{chandrasegaran_hourvideo_2024} & 45.7 min & \no & \no & \no & -- & -- \\
Neptune~\citep{nagrani2025neptunelongorbitbenchmarking} & 2.5 min & \no & \yes & \no & 16.3 & 29.5 \\
\midrule
\multicolumn{7}{@{}l}{\textsc{Films and TV}} \\
MovieQA~\citep{tapaswi_movieqa_2016} & 3.4 min clips & \partly & \no & \no & 9.3 & 5.6 \\
TVQA~\citep{lei-etal-2018-tvqa} & 60--90 s & \no & \no & \no & 13.5 & 5.2 \\
CinePile~\citep{rawal2024cinepile} & 2.7 min & \no & \no & \no & -- & -- \\
MovieChat-1K~\citep{song2024moviechatdensetokensparse} & 9.4 min & \no & \yes & \no & 7.8 & 2.3 \\
MoMentS~\citep{villa-cueva-etal-2025-moments} & 14.6 min & \no & \no & \yes & 12.6 & 14.6 \\
MoVQA~\citep{Zhang2023MoVQAAB} & 126 min & \yes & \no & \no & 11.5 & 17.5 \\
DeepMovieQA~\citep{fung-etal-2023-deepmaven} & 52 min & \yes & \yes & \no & -- & 19--27$^\dagger$ \\
InfiniBench~\citep{ataallah-etal-2025-infinibench} & 53 min & \partly & \partly & \partly & -- & -- \\
MovieGraph-ToM~\citep{moviegraphtom_2026} & film snippets & \partly & \partly & \yes & -- & -- \\
NA-VQA~\citep{jain2026narrativealignedlongform} & ${\sim}$2 h & \yes & \yes & \partly & -- & -- \\
\midrule
\rowcolor{red!7}\textbf{\lflvqa (ours)} & \textbf{116 min} & \yes & \yes & \yes & \textbf{16.3} & \textbf{61.0} \\
\bottomrule
\end{tabular}}
\caption{\textbf{\lflvqa beside existing video QA benchmarks}, as reported by each paper. \emph{Full films}: questions are asked over entire feature films (\partly: clips or selected snippets of films, or a mix of films and shorter videos). \emph{Open-ended}: answers are free text (\partly: some tasks). \emph{ToM}: questions target characters' mental states (\partly: some question categories). Q len and A len are mean words; where answers are not open-ended, A len is the length of the correct option; --: not reported. $^\dagger$Tokens, as reported by DeepMovieQA. Each \lflvqa question draws on 4.2 scenes spanning 23 minutes on average, and in o200k tokens its questions and answers average 21 and 75 (Table~\ref{tab:question-types}).}
\label{tab:benchmark-comparison}
\end{table}

\section{Compute Resources}
\label{app:compute}

We used two sources of compute: GPUs for preprocessing and experiments, and the OpenAI API for question generation and evaluation.

\paragraph{GPU Hours.}The GPU hours split between preprocessing and experiments. Preprocessing (Section~\ref{sec:dataset-collection}), i.e., parsing screenplays into scenes and extracting the belief record of every character, was done with Qwen3.5-27B and took around 800 GPU hours. Most of this went to belief annotation, since it requires a model call for every character in every scene. The experiments in Section~\ref{sec:experiments} took around 600 GPU hours in total, covering video captioning, the long-context and agentic baselines, and character identification for all four model families. We did not train any models.

\paragraph{OpenAI Credits.} The API credits were spent on gpt-5.6-sol and gpt-5.6-luna. Generating the question--answer pairs (Section~\ref{sec:dataset-generation}), which includes writing questions and reference answers, and verifying every claim against the screenplay, cost about USD 1,200 with gpt-5.6-sol. Checking the reference answers and running the LLM-as-a-judge evaluation for all baselines cost about USD 300 with gpt-5.6-luna. 
The judge comparison in Appendix~\ref{app:llm-judge} cost a further USD 8 with gpt-5.6-terra and USD 5 with gemini-3.8-flash.

\section{Data Collection and Preprocessing}
\label{app:dataset-preprocessing-extra}

This appendix details the preprocessing steps of Section~\ref{sec:dataset-collection}, which turn each screenplay into a time-stamped sequence of scenes.
Every step uses Qwen3.5-27B, and Tables~\ref{tab:prompt-page}--\ref{tab:prompt-coref} in Appendix~\ref{app:prompt-tables} give the prompts.

\paragraph{Parsing.}
We render each page of the screenplay as an image, and the model reads it visually, using indentation to separate scene headings, action lines, character cues, parentheticals, and dialogue (Table~\ref{tab:prompt-page}).
It returns the page's scenes with speaker-attributed dialogue and stage directions.
We merge a scene that continues across a page break with its continuation, so each scene appears once.

\paragraph{Scene summaries.}
The model summarizes every scene in a few sentences (Table~\ref{tab:prompt-scene-summary}).
The summaries later give the verifier an overview of the whole film (Section~\ref{sec:dataset-generation}).

\paragraph{Coreference.}
Screenplays refer to one character by several names: a full name, a first name, a role such as \textsc{the woman}, or a spelling variant.
Given every name with the scenes it appears in, the model groups the names that refer to the same character (Table~\ref{tab:prompt-coref}), and we map each group to one name used throughout the film.

\paragraph{Alignment to the subtitles.}
We time-stamp each scene by aligning the screenplay's dialogue lines to the subtitle cues with a monotonic sequence alignment.
A line is matched when at least half of its tokens match a cue, and a scene takes the time span of its matched lines.
Scenes without a matched line get an interpolated span and are never used as anchors.
The alignment also records when each character's name is first spoken aloud, which keeps unspoken names out of the questions (Appendix~\ref{app:dataset-generation-extra}).

\section{Character Belief Annotation}
\label{app:dataset-belief-extraction-extra}

This appendix details how we build the belief record (Section~\ref{sec:dataset-collection}).
Every call uses Qwen3.5-27B and works on one scene at a time, in film order, so a character's record at a scene depends only on what came before it.
Tables~\ref{tab:prompt-events}--\ref{tab:prompt-meta} in Appendix~\ref{app:prompt-tables} give the prompts, which call scene events \emph{canonical events}.

\paragraph{Scene events.}
Each scene is decomposed into atomic scene events, one action or one consequential line of dialogue each, with the characters who take part in the event and the characters who witness it (Table~\ref{tab:prompt-events}).
Every event receives an identifier that later steps cite.

\paragraph{First-order beliefs.}
For every character and scene, two calls update the character's record, each given the scene, its events, and the record so far.
The first decides whether the character is present and which events they witness (Table~\ref{tab:prompt-fo-presence}).
The second records what the character is told and whether they believe it, what they infer, which of their beliefs are false given the scene events, and which earlier beliefs they revise; every inference must cite the events it rests on (Table~\ref{tab:prompt-fo-beliefs}).
The result is carried into the next scene.

\paragraph{Meta-beliefs.}
For every character $A$ and scene, the model records $A$'s new or changed beliefs about each other character $B$'s beliefs, limited to the second order ($A$ believes that $B$ believes $X$), together with how each formed, for example: $A$ told $B$, $A$ saw $B$'s behavior, they witnessed an event together, or $B$ was absent when something happened (Table~\ref{tab:prompt-meta}).
Each meta-belief must cite an event of the scene and is discarded otherwise.

\section{Belief Records Behind the Questions}
\label{app:belief-records}

Figures~\ref{fig:belief-example-sinners} to~\ref{fig:belief-example-barbie} show the belief records behind three more questions, each from a different film.
They use the same layout as Figure~\ref{fig:belief-example}: a dashed box marks a false belief, an arrow a revision, and a grey box a claim the character rejects.

\begin{figure}[p]
  \centering
  \includegraphics[width=.94\textwidth]{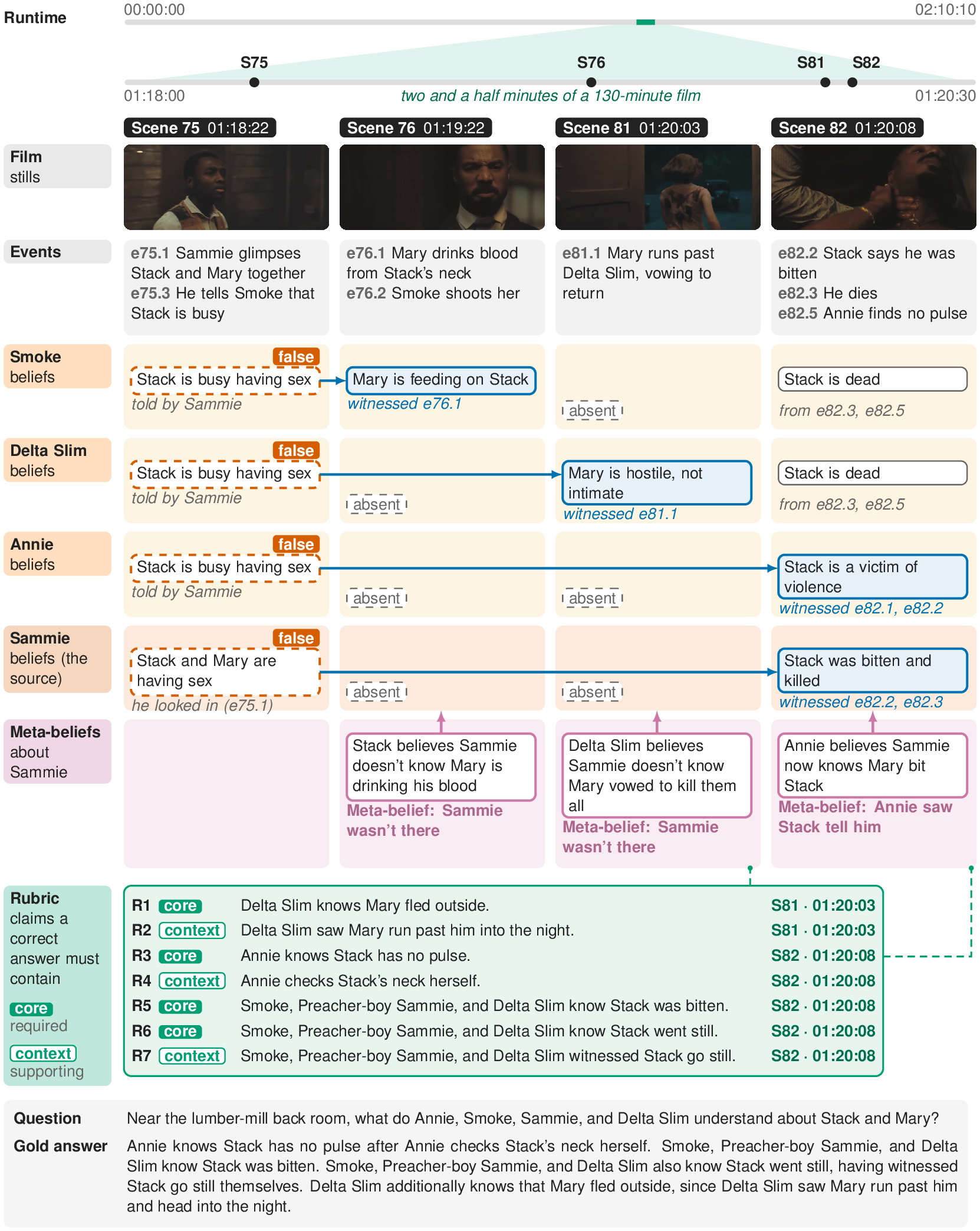}
  \input{figures/belief_example_sinners_caption}
\end{figure}

\begin{figure}[p]
  \centering
  \includegraphics[width=.94\textwidth]{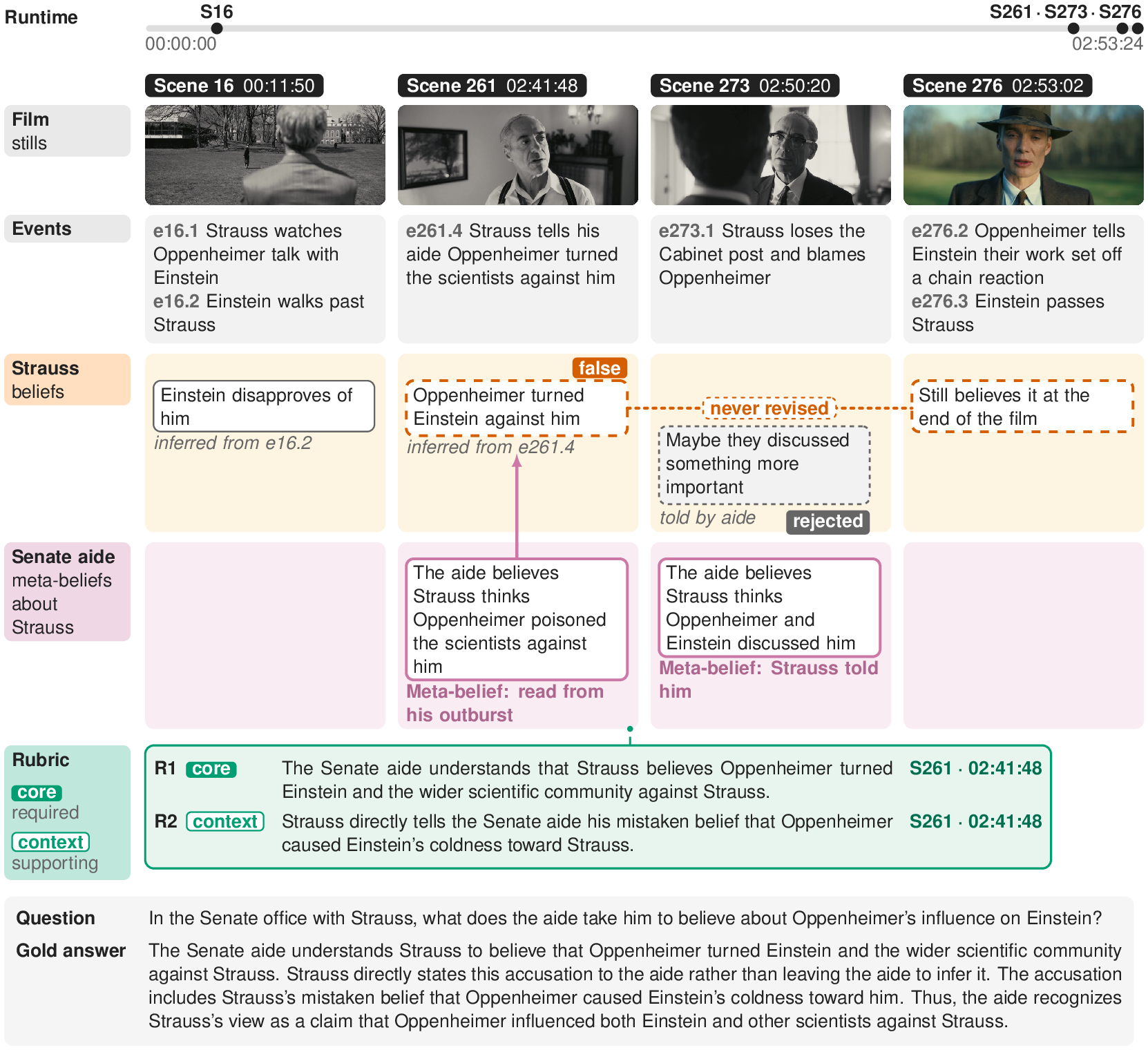}
  \input{figures/belief_example_oppenheimer_caption}
\end{figure}

\begin{figure}[p]
  \centering
  \includegraphics[width=.94\textwidth]{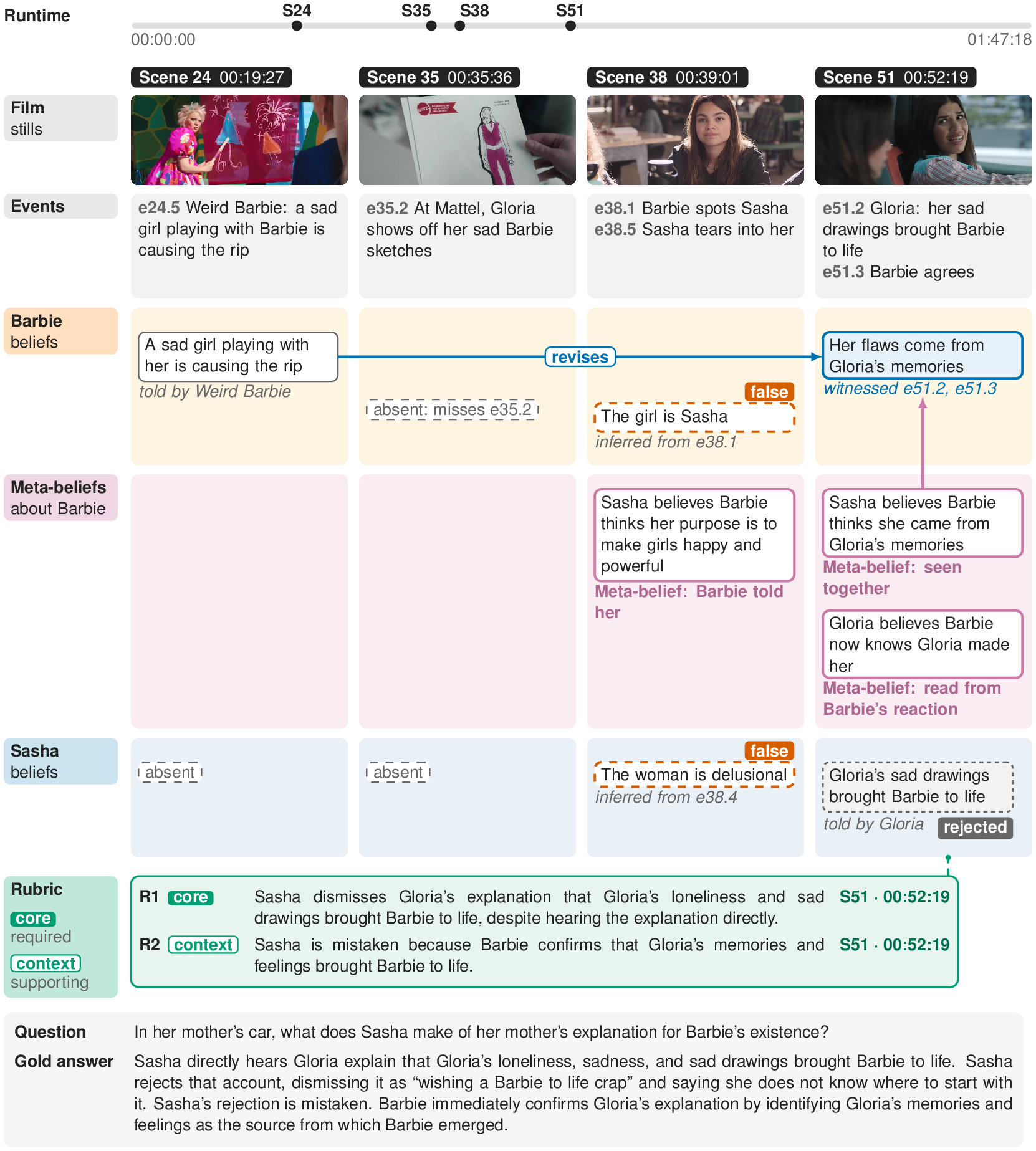}
  \input{figures/belief_example_barbie_caption}
\end{figure}

\section{Question Generation and Verification}
\label{app:dataset-generation-extra}
\label{app:templates}

This appendix details question generation and verification (Section~\ref{sec:dataset-generation}): what a context-set contains, the rules for discarding context-sets, and the prompts that turn a context-set into a (question, answer, rubric) triple.

\subsection{Context-Sets}
\label{app:packs}

A context-set is built for one question type, one center character, and one anchor scene.
It is the only input the question writer sees: it never contains the screenplay text, and the question writer cannot look anything up.
Table~\ref{tab:pack-fields} lists its fields, and Table~\ref{tab:packs} the belief entries of each question type.
Three fields keep questions answerable from the film alone.
The screenplay names characters the film never names aloud, such as a role label like \textsc{line judge}, so the context-set lists the names the film has spoken by the anchor, the names it speaks only later, and the names it never speaks; the question writer may use the first two and never the third.

\begin{table}[htbp]
\centering\footnotesize
\setlength{\tabcolsep}{4pt}
\renewcommand{\arraystretch}{1.15}
\begin{tabular}{@{}l p{0.72\textwidth}@{}}
\toprule
Field & Contents \\
\midrule
\texttt{holder} & The center character: the character the question is about; for Q1 a cohort of characters. \\
\texttt{anchor} & The scene the question points at, with its time interval and whether the interval is aligned to the subtitles or interpolated. \\
\texttt{basis} & The belief entries: the belief-record entries the type is defined over (Table~\ref{tab:packs}); each cites the scene events it rests on. Meta-belief entries also carry the target character's own beliefs at that scene, so that the question writer can see where the center character's model of another character diverges from that character's actual state. \\
\texttt{scenes} & The anchor scene and one scene either side, each with its summary, location and scene events with their participants and witnesses. \\
\texttt{roster} & The roster entry of every character the belief entries or the scenes mention. \\
\texttt{name\_budget} & Names the film has spoken aloud by the anchor. \\
\texttt{named\_later} & Names the film speaks only after the anchor, each with the time it is first spoken. \\
\texttt{forbidden\_names} & Names the film never speaks aloud. \\
\texttt{anchor\_location} & The anchor scene's location, and, as \texttt{anchor\_location\_rivals}, the number of other scenes at the same location with the center character present, which tells the question writer whether naming the place is enough to pin the moment down. \\
\bottomrule
\end{tabular}
\caption{Fields of a context-set, given to the question writer as JSON.}
\label{tab:pack-fields}
\end{table}

\begin{table}[htbp]
\centering\footnotesize
\setlength{\tabcolsep}{4pt}
\renewcommand{\arraystretch}{1.15}
\begin{tabular}{@{}l >{\raggedright\arraybackslash}p{3.3cm} >{\raggedright\arraybackslash}p{5.0cm} >{\raggedright\arraybackslash}p{1.9cm}@{}}
\toprule
Type & Asks & Belief entries in the context-set & Anchor \\
\midrule
\multicolumn{4}{@{}l}{\textsc{Belief state} \textit{--- one context-set per (center character, scene)}} \\
\addlinespace[2pt]
Q1 Knowledge split & What each of several characters takes to be the case at one moment & For every character with a record at the scene: the events they witnessed and the false beliefs they hold; kept only if $\geq 3$ characters are in $\geq 2$ knowledge states & the scene \\
Q2 Realization & When and through what a character's understanding changed & The character's \emph{revisions} at the scene, each with the events that triggered it & the scene \\
Q3 False belief & What a character takes to be the case just before acting on it & The character's \emph{false beliefs held} at the scene & the scene \\
Q4 Knowledge source & How a character comes to know a fact & The character's \emph{meta-beliefs} formed at the scene that carry a source (told, observed behavior, shared witnessing, absence), joined to the target's own state & the scene \\
Q5 First-order belief & What $X$ takes $Y$ to believe & $X$'s \emph{first-order meta-beliefs} about $Y$ formed at the scene, joined to $Y$'s own beliefs there & the scene \\
Q6 Second-order belief & What $X$ takes $Y$ to believe about $Z$ & $X$'s \emph{third-party meta-beliefs} formed at the scene, joined to $Y$'s own beliefs there & the scene \\
Q7 Accepting a claim & What a character makes of something they are told & The propositions the character was \emph{told} at the scene, with whether each was believed; context-sets are selected so that acceptance and rejection both occur & the scene \\
N3 Causal chain & What chain of events leads a character to a belief & The character's \emph{inferences} with their basis events and \emph{revisions} with their trigger events at the scene & the scene \\
\midrule
\multicolumn{4}{@{}l}{\textsc{Arc} \textit{--- one context-set per center character (or per pair), carrying the whole timeline}} \\
\addlinespace[2pt]
Q8 Order of discovery & In what order a character learned a set of facts & Every scene at which the character witnessed, was told, inferred, or revised something, in order; $\geq 3$ scenes, thinned to 24 keeping the first and last & last aligned scene of the arc \\
N1 Epistemic arc & What a character comes to know across the film & Same as Q8 & last aligned scene of the arc \\
N2 Trust evolution & How $X$'s model of $Y$ changes across the film & Every \emph{meta-belief} of $X$ about $Y$, in order, each joined to $Y$'s state at that scene; $\geq 3$ scenes, thinned to 24 & last aligned scene of the arc \\
\bottomrule
\end{tabular}
\caption{Belief entries of each question type. For Q1, the center character is a group of characters.}
\label{tab:packs}
\end{table}

\paragraph{Discard rules.}
Before any model is called, we discard a context-set when a viewer who watched the whole film could not answer a question built from it: the center character is never named aloud, the events the belief entries cite fall outside the scenes the context-set carries, or, for the belief-state types, the center character takes no part in the anchor scene.
We also discard context-sets anchored on a scene without a subtitle match, since its time is only interpolated.
These rules discard 23\% of context-sets.
We then sample the remaining context-sets so that questions spread across scenes, characters, and runtime.

\subsection{Generation Stages and the Judge}
\label{app:prompts}

We build each (question, answer, rubric) triple in three stages with gpt-5.6-sol, and grade model answers with a judge, gpt-5.6-luna.
Tables~\ref{tab:type-specs}--\ref{tab:prompt-judge} in Appendix~\ref{app:prompt-tables} give the prompts.

\paragraph{Stage 1: Writing the question.}
The question writer receives its prompt (Table~\ref{tab:prompt-question}), the specification of the question type (Table~\ref{tab:type-specs}), two example triples from a film that never itself receives questions, and the context-set as JSON.
It returns a question and an outline of two to four claims, each naming the belief-record entry it rests on, together with a \emph{distractor}: the specific wrong answer a careless viewer would give.
It may instead decline a context-set with no plausible wrong answer.
We then discard questions that fail automatic checks: a question must be a single open interrogative that cannot be answered by yes or no, offers no alternatives to choose between, names no character the film never names, and never reveals how or when the center character learned something or whether they are right.

\paragraph{Stage 2: Writing the reference answer and rubric.}
From the question and its outline, a second call writes the reference answer as 60 to 140 words of prose, together with the claims it will be graded by (Table~\ref{tab:prompt-answer}).
Every claim must be plainly asserted by the answer, and every assertion in the answer must be backed by a claim.
Each claim is a single proposition that a grader can judge alone, and is either \emph{core}, which a complete answer must make, or \emph{context}, which an answer may omit but must not contradict.

\paragraph{Stage 3: Verifying against the screenplay.}
The verifier checks every claim against the screenplay (Table~\ref{tab:prompt-verify}).
It sees a summary of every scene and the full text of the anchor scene, of every scene the claims cite, and of the scene where a false or revised belief first formed.
When unsure, it rejects the claim: a wrong claim corrupts every answer graded against it, while a dropped claim costs one point of recall.
We remove the claims the verifier does not support, rewrite those that are only partly supported (for instance, a claim whose only unsupported part is that a belief is false), and rewrite the reference answer to assert only the surviving claims.
If the surviving claims no longer answer the question, the verifier rewrites the question under the same constraints as Stage~1; about one in ten released questions was revised this way.
We retire a question when too few core claims survive, when the screenplay contradicts a core claim, or when its rewrite fails the automatic checks.
Finally, the judge grades each reference answer against its own rubric, and we retire any question whose answer omits a core claim.

\paragraph{Evaluation: the judge.}
At evaluation time, the judge labels every rubric claim as \emph{asserted}, \emph{contradicted}, or \emph{omitted} by a model answer, judging only against the answer's text (Table~\ref{tab:prompt-judge}).
Content the answer adds beyond the rubric is not graded, and an answer that declines the question omits every claim.
Appendix~\ref{app:examples} shows the judge's verdicts on example answers.
Appendix~\ref{app:llm-judge} tests how reliable the judge is.

\section{Reliability of the LLM Judge}
\label{app:llm-judge}


We test whether the claim verdicts depend on the choice of judge.
Besides gpt-5.6-luna, two judges from other model families, gpt-5.6-terra and gemini-3.8-flash, graded the same model answers with the same prompt (Table~\ref{tab:prompt-judge}).\footnote{gpt-5.6-luna runs at medium reasoning effort, gpt-5.6-terra at low reasoning effort, and gemini-3.8-flash at a low thinking level with temperature 0.}
We sampled 200 released questions from 55 films and took the answers of 33 baseline settings to each question, so that each judge labels 27,875 rubric claims as \emph{asserted}, \emph{contradicted}, or \emph{omitted}.
Table~\ref{tab:judge-agreement} reports agreement on these claim verdicts.
$\kappa$ and Krippendorff's $\alpha$ are between 0.71 and 0.78 for every pair of judges, and 0.74 for all three.

\begin{table}[htb]
\centering
\small
\begin{tabular}{@{}lcc@{}}
\toprule
Judges & $\kappa$ & Krippendorff's $\alpha$ \\
\midrule
gpt-5.6-luna, gpt-5.6-terra & 0.778 & 0.778 \\
gpt-5.6-luna, gemini-3.8-flash & 0.732 & 0.732 \\
gpt-5.6-terra, gemini-3.8-flash & 0.714 & 0.712 \\
All three & 0.742 & 0.742 \\
\bottomrule
\end{tabular}
\caption{\textbf{Claim-level agreement between judges} on 27,875 claim verdicts (asserted, contradicted, or omitted). $\kappa$ is Cohen's for a pair of judges and Fleiss' for all three.}
\label{tab:judge-agreement}
\end{table}

\section{Examples of Questions, Answers and Rubrics}
\label{app:examples}

This appendix shows examples from the dataset and model answers to them.

\subsection{One Question per Type}
\label{app:example-triples}

Tables~\ref{tab:example-triples-a} to~\ref{tab:example-triples-c} give one question per type, each with its reference answer and rubric.
Core claims are marked $\blacksquare$ and context claims $\square$.
Under each rubric, a map shows where its evidence lies in the film, with the anchor scene marked by a triangle.
\definecolor{tabblue}{HTML}{4C8BF5}
\definecolor{tabred}{HTML}{E8453C}
\colorlet{packband}{gray!12}
\providecommand{\pq}[1]{\par\noindent{\color{black!60}\scshape question:} \itshape #1\par}
\providecommand{\pa}[1]{\par\noindent{\color{black!60}\scshape gold answer:} #1\par\smallskip}
\providecommand{\pcore}[1]{\par\hangindent 1.1em\hangafter 1\noindent$\blacksquare$~#1}
\providecommand{\pctx}[1]{\par\hangindent 1.1em\hangafter 1\noindent{\color{black!45}$\square$~#1}}

\begin{table}[p]
\centering\scriptsize
\setlength{\tabcolsep}{4pt}\renewcommand{\arraystretch}{1.0}
\begin{tabular}{@{}p{\textwidth}@{}}
\toprule
\rowcolor{packband}\multicolumn{1}{@{}p{\dimexpr\textwidth\relax}@{}}{\rule{0pt}{2.4ex}\textbf{Q1\ Knowledge split}\quad \emph{The Holdovers} (2023)\quad center character: Angus Tully\hfill{\color{black!55}3 core / 5 claims \quad evidence: 2 scenes over 10 min}}\\
\begin{minipage}[t]{\linewidth}\vspace{1pt}\pq{At the Winning Ticket, what do Paul, Angus, and Miss Crane each understand about Christmas Eve?}\pa{At the Winning Ticket, Miss Crane knows that she has invited both Paul Hunham and Angus Tully to her Christmas Eve party. Paul hears the invitation. He knows it includes both Angus and himself. Angus is away from the table at the moment when Miss Crane extends that invitation, and he does not yet know about the invitation that she has made.}\pcore{{\color{black!55}\scriptsize 1}~Miss Crane knows she invited Paul Hunham and Angus Tully to her Christmas Eve party.}\pctx{{\color{black!55}\scriptsize 2}~Paul Hunham hears Miss Crane's invitation.}\pcore{{\color{black!55}\scriptsize 3}~Paul Hunham knows Miss Crane's invitation includes Paul Hunham and Angus Tully.}\pctx{{\color{black!55}\scriptsize 4}~Angus Tully is away from the table during Miss Crane's invitation.}\pcore{{\color{black!55}\scriptsize 5}~Angus Tully does not yet know about Miss Crane's invitation.}\par\smallskip\noindent\makebox[\linewidth][l]{%
\rule[-1.6ex]{0pt}{4.6ex}%
\makebox[0pt][l]{\color{black!30}\rule[0.7ex]{\linewidth}{0.4pt}}%
\makebox[0pt][l]{\hspace{0.4169\linewidth}\color{tabblue}\rule[0.2ex]{0.0212\linewidth}{1.7ex}}%
\makebox[0pt][l]{\hspace{0.4824\linewidth}\color{tabblue}\rule[0.2ex]{0.0102\linewidth}{1.7ex}}%
\makebox[0pt][l]{\hspace{0.4169\linewidth}\raisebox{2.2ex}{\makebox[0pt][c]{\tiny\color{tabblue}54:46}}}%
\makebox[0pt][l]{\hspace{0.4824\linewidth}\raisebox{2.2ex}{\makebox[0pt][c]{\tiny\color{tabblue}1:03:23}}}%
\makebox[0pt][l]{\hspace{0.4169\linewidth}\raisebox{-1.1ex}{\makebox[0pt][c]{\scriptsize\color{black!55}$\blacktriangle$}}}%
}
\par\nointerlineskip\makebox[\linewidth][s]{\tiny\color{black!55}start of film\hfill end of film, 2:11:23}
\par\smallskip\noindent{\scriptsize{\color{black!60}\scshape evidence in the film:} {\color{tabblue}\rule[-0.1ex]{0.55em}{0.55em}}\,\textbf{54:46} The Winning Ticket {\color{black!55}(claims 1--5)}\quad {\color{tabblue}\rule[-0.1ex]{0.55em}{0.55em}}\,\textbf{1:03:23} Kitchen {\color{black!55}(claim 5)}}\vspace{2pt}\end{minipage}\\
\addlinespace[2pt]
\rowcolor{packband}\multicolumn{1}{@{}p{\dimexpr\textwidth\relax}@{}}{\rule{0pt}{2.4ex}\textbf{Q2\ Realization}\quad \emph{Conclave} (2024)\quad center character: Joseph Tremblay\hfill{\color{black!55}2 core / 5 claims \quad evidence: 2 scenes over 71 min}}\\
\begin{minipage}[t]{\linewidth}\vspace{1pt}\pq{In Casa Santa Marta's courtyard corridor, when does Tremblay's view of Lawrence's methods change?}\pa{Tremblay's view changes when Lawrence personally confronts him in the courtyard corridor. Until then, Tremblay had regarded Lawrence as guided by diplomatic protocol and courtesy. Lawrence directly demands that Tremblay withdraw and accuses him of sabotaging Adeyemi. The confrontation makes Tremblay realize that Lawrence will bypass protocol and directly attack a political rival.}\pctx{{\color{black!55}\scriptsize 1}~Joseph Tremblay had regarded Cardinal Lawrence as guided by diplomatic protocol and courtesy.}\pcore{{\color{black!55}\scriptsize 2}~Cardinal Lawrence directly demands that Joseph Tremblay withdraw.}\pcore{{\color{black!55}\scriptsize 3}~Cardinal Lawrence accuses Joseph Tremblay of sabotaging Adeyemi.}\pctx{{\color{black!55}\scriptsize 4}~Joseph Tremblay realizes Cardinal Lawrence will bypass diplomatic protocol.}\pctx{{\color{black!55}\scriptsize 5}~Joseph Tremblay realizes Cardinal Lawrence will directly attack a political rival.}\par\smallskip\noindent\makebox[\linewidth][l]{%
\rule[-1.6ex]{0pt}{4.6ex}%
\makebox[0pt][l]{\color{black!30}\rule[0.7ex]{\linewidth}{0.4pt}}%
\makebox[0pt][l]{\hspace{0.0532\linewidth}\color{black!45}\rule[0.2ex]{0.0136\linewidth}{1.7ex}}%
\makebox[0pt][l]{\hspace{0.6668\linewidth}\color{tabblue}\rule[0.2ex]{0.0165\linewidth}{1.7ex}}%
\makebox[0pt][l]{\hspace{0.0532\linewidth}\raisebox{2.2ex}{\makebox[0pt][c]{\tiny\color{black!45}6:00}}}%
\makebox[0pt][l]{\hspace{0.6668\linewidth}\raisebox{2.2ex}{\makebox[0pt][c]{\tiny\color{tabblue}1:15:19}}}%
\makebox[0pt][l]{\hspace{0.6668\linewidth}\raisebox{-1.1ex}{\makebox[0pt][c]{\scriptsize\color{black!55}$\blacktriangle$}}}%
}
\par\nointerlineskip\makebox[\linewidth][s]{\tiny\color{black!55}start of film\hfill end of film, 1:52:57}
\par\smallskip\noindent{\scriptsize{\color{black!60}\scshape evidence in the film:} {\color{black!45}\rule[-0.1ex]{0.55em}{0.55em}}\,\textbf{6:00} Papal Suite / Sitting Room {\color{black!55}(claim 1)}\quad {\color{tabblue}\rule[-0.1ex]{0.55em}{0.55em}}\,\textbf{1:15:19} Casa Santa Marta / Courtyard Corridor {\color{black!55}(claims 2--5)}}\vspace{2pt}\end{minipage}\\
\addlinespace[2pt]
\rowcolor{packband}\multicolumn{1}{@{}p{\dimexpr\textwidth\relax}@{}}{\rule{0pt}{2.4ex}\textbf{Q3\ False belief}\quad \emph{Nosferatu} (2024)\quad center character: Louise\hfill{\color{black!55}3 core / 4 claims \quad evidence: 3 scenes over 72 min}}\\
\begin{minipage}[t]{\linewidth}\vspace{1pt}\pq{In the Harding nursery, what danger does Louise believe she and Clara face?}\pa{Louise believes that a monster is in the Harding nursery. She also believes that this monster threatens to kill her. Louise further believes that Clara faces the same lethal danger from the monster. The shadowy figure seen nearby, however, proves to be Ellen. Thus, in Louise's understanding, both Louise and Clara are in danger of being killed by the monster.}\pcore{{\color{black!55}\scriptsize 1}~Louise believes a monster is in the Harding nursery.}\pcore{{\color{black!55}\scriptsize 2}~Louise believes the monster will kill Louise.}\pcore{{\color{black!55}\scriptsize 3}~Louise believes the monster will kill Clara.}\pctx{{\color{black!55}\scriptsize 4}~The nearby shadowy figure proves to be Ellen.}\par\smallskip\noindent\makebox[\linewidth][l]{%
\rule[-1.6ex]{0pt}{4.6ex}%
\makebox[0pt][l]{\color{black!30}\rule[0.7ex]{\linewidth}{0.4pt}}%
\makebox[0pt][l]{\hspace{0.6502\linewidth}\color{black!45}\rule[0.2ex]{0.0161\linewidth}{1.7ex}}%
\makebox[0pt][l]{\hspace{0.0932\linewidth}\color{tabblue}\rule[0.2ex]{0.0161\linewidth}{1.7ex}}%
\makebox[0pt][l]{\hspace{0.1113\linewidth}\color{tabblue}\rule[0.2ex]{1.6pt}{1.7ex}}%
\makebox[0pt][l]{\hspace{0.0932\linewidth}\raisebox{2.2ex}{\makebox[0pt][c]{\tiny\color{tabblue}11:40}}}%
\makebox[0pt][l]{\hspace{0.6502\linewidth}\raisebox{2.2ex}{\makebox[0pt][c]{\tiny\color{black!45}1:21:24}}}%
\makebox[0pt][l]{\hspace{0.6502\linewidth}\raisebox{-1.1ex}{\makebox[0pt][c]{\scriptsize\color{black!55}$\blacktriangle$}}}%
}
\par\nointerlineskip\makebox[\linewidth][s]{\tiny\color{black!55}start of film\hfill end of film, 2:05:11}
\par\smallskip\noindent{\scriptsize{\color{black!60}\scshape evidence in the film:} {\color{tabblue}\rule[-0.1ex]{0.55em}{0.55em}}\,\textbf{11:40} Harding House. Rear Drawing Room {\color{black!55}(claim 1)}\quad {\color{tabblue}\rule[-0.1ex]{0.55em}{0.55em}}\,\textbf{13:56} Front Drawing Room {\color{black!55}(claims 1--3)}\quad {\color{black!45}\rule[-0.1ex]{0.55em}{0.55em}}\,\textbf{1:21:24} Harding House. Upstairs Hallway / Nursery {\color{black!55}(claim 4)}}\vspace{2pt}\end{minipage}\\
\addlinespace[2pt]
\rowcolor{packband}\multicolumn{1}{@{}p{\dimexpr\textwidth\relax}@{}}{\rule{0pt}{2.4ex}\textbf{Q4\ Knowledge source}\quad \emph{Nickel Boys} (2024)\quad center character: Elwood\hfill{\color{black!55}2 core / 2 claims \quad evidence: 2 scenes over 4 min}}\\
\begin{minipage}[t]{\linewidth}\vspace{1pt}\pq{Inside the White House, how does Elwood work out what the loud machinery is for?}\pa{Inside the White House, Elwood witnesses Black Mike being taken into the beating room and hears the machinery roar during Black Mike's punishment. Elwood thus observes the sequence of Black Mike entering, the punishment occurring, and the machinery roaring. From that sequence, Elwood infers that the machinery masks the victims' screams. He works out its purpose from what he witnesses and hears.}\pcore{{\color{black!55}\scriptsize 1}~Elwood witnesses Black Mike being taken into the beating room and hears the machinery roar during Black Mike's punishment.}\pcore{{\color{black!55}\scriptsize 2}~Elwood infers from that sequence that the machinery masks the victims' screams.}\par\smallskip\noindent\makebox[\linewidth][l]{%
\rule[-1.6ex]{0pt}{4.6ex}%
\makebox[0pt][l]{\color{black!30}\rule[0.7ex]{\linewidth}{0.4pt}}%
\makebox[0pt][l]{\hspace{0.3330\linewidth}\color{tabblue}\rule[0.2ex]{0.0061\linewidth}{1.7ex}}%
\makebox[0pt][l]{\hspace{0.3507\linewidth}\color{tabblue}\rule[0.2ex]{0.0119\linewidth}{1.7ex}}%
\makebox[0pt][l]{\hspace{0.3330\linewidth}\raisebox{2.2ex}{\makebox[0pt][c]{\tiny\color{tabblue}44:59}}}%
\makebox[0pt][l]{\hspace{0.3330\linewidth}\raisebox{-1.1ex}{\makebox[0pt][c]{\scriptsize\color{black!55}$\blacktriangle$}}}%
}
\par\nointerlineskip\makebox[\linewidth][s]{\tiny\color{black!55}start of film\hfill end of film, 2:15:05}
\par\smallskip\noindent{\scriptsize{\color{black!60}\scshape evidence in the film:} {\color{tabblue}\rule[-0.1ex]{0.55em}{0.55em}}\,\textbf{44:59} White House {\color{black!55}(claims 1, 2)}\quad {\color{tabblue}\rule[-0.1ex]{0.55em}{0.55em}}\,\textbf{47:22} White House / Beating Room {\color{black!55}(claim 2)}}\vspace{2pt}\end{minipage}\\
\bottomrule
\end{tabular}
\caption{\textbf{Released questions, one per type: Belief-state types, Q1--Q4.} Core claims are marked $\blacksquare$, context claims $\square$; $\blacktriangle$ marks the anchor scene.}
\label{tab:example-triples-a}
\end{table}

\begin{table}[p]
\centering\scriptsize
\setlength{\tabcolsep}{4pt}\renewcommand{\arraystretch}{1.0}
\begin{tabular}{@{}p{\textwidth}@{}}
\toprule
\rowcolor{packband}\multicolumn{1}{@{}p{\dimexpr\textwidth\relax}@{}}{\rule{0pt}{2.4ex}\textbf{Q5\ First-order belief}\quad \emph{Oppenheimer} (2023)\quad center character: Gray\hfill{\color{black!55}2 core / 2 claims \quad evidence: 5 scenes over 22 min}}\\
\begin{minipage}[t]{\linewidth}\vspace{1pt}\pq{During Oppenheimer's security hearing, what does Gray understand about Robb's view of Oppenheimer?}\pa{Gray understands that Robb doubts Oppenheimer told the whole truth. Gray therefore recognizes Robb's distrust of Oppenheimer's truthfulness. Gray also understands that Robb sees Oppenheimer as professionally willing to follow orders but morally inconsistent about the atomic and hydrogen bombs. In Gray's understanding, Robb separates Oppenheimer's willingness to perform assigned work from Oppenheimer's differing moral positions on the two weapons.}\pcore{{\color{black!55}\scriptsize 1}~Gray understands that Robb doubts Oppenheimer told the whole truth.}\pcore{{\color{black!55}\scriptsize 2}~Gray understands that Robb sees Oppenheimer as professionally willing to follow orders but morally inconsistent about the atomic and hydrogen bombs.}\par\smallskip\noindent\makebox[\linewidth][l]{%
\rule[-1.6ex]{0pt}{4.6ex}%
\makebox[0pt][l]{\color{black!30}\rule[0.7ex]{\linewidth}{0.4pt}}%
\makebox[0pt][l]{\hspace{0.8374\linewidth}\color{tabblue}\rule[0.2ex]{0.0093\linewidth}{1.7ex}}%
\makebox[0pt][l]{\hspace{0.8562\linewidth}\color{tabblue}\rule[0.2ex]{0.0043\linewidth}{1.7ex}}%
\makebox[0pt][l]{\hspace{0.8630\linewidth}\color{tabblue}\rule[0.2ex]{0.0140\linewidth}{1.7ex}}%
\makebox[0pt][l]{\hspace{0.9139\linewidth}\color{tabblue}\rule[0.2ex]{0.0144\linewidth}{1.7ex}}%
\makebox[0pt][l]{\hspace{0.9551\linewidth}\color{tabblue}\rule[0.2ex]{0.0083\linewidth}{1.7ex}}%
\makebox[0pt][l]{\hspace{0.8374\linewidth}\raisebox{2.2ex}{\makebox[0pt][c]{\tiny\color{tabblue}2:25:13}}}%
\makebox[0pt][l]{\hspace{0.9139\linewidth}\raisebox{2.2ex}{\makebox[0pt][c]{\tiny\color{tabblue}2:38:29}}}%
\makebox[0pt][l]{\hspace{0.9551\linewidth}\raisebox{-1.1ex}{\makebox[0pt][c]{\scriptsize\color{black!55}$\blacktriangle$}}}%
}
\par\nointerlineskip\makebox[\linewidth][s]{\tiny\color{black!55}start of film\hfill end of film, 2:53:24}
\par\smallskip\noindent{\scriptsize{\color{black!60}\scshape evidence in the film:} {\color{tabblue}\rule[-0.1ex]{0.55em}{0.55em}}\,\textbf{2:25:13} Room 2022, Atomic Energy Commission {\color{black!55}(claim 1)}\quad {\color{tabblue}\rule[-0.1ex]{0.55em}{0.55em}}\,\textbf{2:28:28} Room 2022, Atomic Energy Commission {\color{black!55}(claim 1)}\quad {\color{tabblue}\rule[-0.1ex]{0.55em}{0.55em}}\,\textbf{2:29:39} Room 2022, Atomic Energy Commission {\color{black!55}(claim 1)}\quad {\color{tabblue}\rule[-0.1ex]{0.55em}{0.55em}}\,\textbf{2:38:29} Room 2022, Atomic Energy Commission {\color{black!55}(claim 1)}\quad {\color{tabblue}\rule[-0.1ex]{0.55em}{0.55em}}\,\textbf{2:45:37} Room 2022, Atomic Energy Commission {\color{black!55}(claim 2)}}\vspace{2pt}\end{minipage}\\
\addlinespace[2pt]
\rowcolor{packband}\multicolumn{1}{@{}p{\dimexpr\textwidth\relax}@{}}{\rule{0pt}{2.4ex}\textbf{Q6\ Second-order belief}\quad \emph{The Wild Robot} (2024)\quad center character: Fink\hfill{\color{black!55}2 core / 3 claims \quad evidence: 2 scenes over 5 min}}\\
\begin{minipage}[t]{\linewidth}\vspace{1pt}\pq{In the lodge beside the woven nest, what does Fink take Brightbill to believe about Roz's place in his past and family?}\pa{Fink understands that Brightbill holds Roz responsible for the accident that killed Brightbill's biological family. Fink also understands that Brightbill no longer accepts Roz as his mother. More specifically, Fink understands that Brightbill rejects Fink's defense that the accident was not Roz's fault. Thus, in Fink's view, Brightbill blames Roz for losing his original family and denies that Roz still occupies a maternal place in his family.}\pcore{{\color{black!55}\scriptsize 1}~Fink understands that Brightbill holds Roz responsible for the accident that killed Brightbill's biological family.}\pcore{{\color{black!55}\scriptsize 2}~Fink understands that Brightbill no longer accepts Roz as his mother.}\pctx{{\color{black!55}\scriptsize 3}~Fink understands that Brightbill rejects Fink's defense that the accident was not Roz's fault.}\par\smallskip\noindent\makebox[\linewidth][l]{%
\rule[-1.6ex]{0pt}{4.6ex}%
\makebox[0pt][l]{\color{black!30}\rule[0.7ex]{\linewidth}{0.4pt}}%
\makebox[0pt][l]{\hspace{0.3521\linewidth}\color{tabblue}\rule[0.2ex]{0.0352\linewidth}{1.7ex}}%
\makebox[0pt][l]{\hspace{0.3899\linewidth}\color{tabblue}\rule[0.2ex]{0.0120\linewidth}{1.7ex}}%
\makebox[0pt][l]{\hspace{0.3521\linewidth}\raisebox{2.2ex}{\makebox[0pt][c]{\tiny\color{tabblue}35:46}}}%
\makebox[0pt][l]{\hspace{0.3899\linewidth}\raisebox{-1.1ex}{\makebox[0pt][c]{\scriptsize\color{black!55}$\blacktriangle$}}}%
}
\par\nointerlineskip\makebox[\linewidth][s]{\tiny\color{black!55}start of film\hfill end of film, 1:41:36}
\par\smallskip\noindent{\scriptsize{\color{black!60}\scshape evidence in the film:} {\color{tabblue}\rule[-0.1ex]{0.55em}{0.55em}}\,\textbf{35:46} Far Side Of The Lake {\color{black!55}(claim 1)}\quad {\color{tabblue}\rule[-0.1ex]{0.55em}{0.55em}}\,\textbf{39:37} Lodge {\color{black!55}(claims 1--3)}}\vspace{2pt}\end{minipage}\\
\addlinespace[2pt]
\rowcolor{packband}\multicolumn{1}{@{}p{\dimexpr\textwidth\relax}@{}}{\rule{0pt}{2.4ex}\textbf{Q7\ Accepting a claim}\quad \emph{Sinners} (2025)\quad center character: Delta Slim\hfill{\color{black!55}2 core / 4 claims \quad evidence: 2 scenes over 6 min}}\\
\begin{minipage}[t]{\linewidth}\vspace{1pt}\pq{At the railroad station, over the bottle of Chicago beer, what does Delta Slim make of Stack's pitch?}\pa{Delta Slim does not dismiss Stack's promised pay and beer as a bluff; he treats both parts of the offer seriously. His initial refusal reflects his commitment to play at Messengers, rather than disbelief that Stack will deliver. When Stack raises the offer, Delta Slim is won over. His performance with Sammie immediately afterward confirms that he accepted Stack's deal.}\pctx{{\color{black!55}\scriptsize 1}~Delta Slim believes Stack's promised pay and beer are genuine rather than a bluff.}\pcore{{\color{black!55}\scriptsize 2}~Delta Slim initially refuses because of Delta Slim's commitment to play at Messengers.}\pcore{{\color{black!55}\scriptsize 3}~Stack's raised offer wins Delta Slim over.}\pctx{{\color{black!55}\scriptsize 4}~Delta Slim's performance with Sammie confirms that Delta Slim accepted Stack's raised offer.}\par\smallskip\noindent\makebox[\linewidth][l]{%
\rule[-1.6ex]{0pt}{4.6ex}%
\makebox[0pt][l]{\color{black!30}\rule[0.7ex]{\linewidth}{0.4pt}}%
\makebox[0pt][l]{\hspace{0.1638\linewidth}\color{tabblue}\rule[0.2ex]{0.0228\linewidth}{1.7ex}}%
\makebox[0pt][l]{\hspace{0.1889\linewidth}\color{tabblue}\rule[0.2ex]{0.0212\linewidth}{1.7ex}}%
\makebox[0pt][l]{\hspace{0.1638\linewidth}\raisebox{2.2ex}{\makebox[0pt][c]{\tiny\color{tabblue}22:29}}}%
\makebox[0pt][l]{\hspace{0.1638\linewidth}\raisebox{-1.1ex}{\makebox[0pt][c]{\scriptsize\color{black!55}$\blacktriangle$}}}%
}
\par\nointerlineskip\makebox[\linewidth][s]{\tiny\color{black!55}start of film\hfill end of film, 2:17:17}
\par\smallskip\noindent{\scriptsize{\color{black!60}\scshape evidence in the film:} {\color{tabblue}\rule[-0.1ex]{0.55em}{0.55em}}\,\textbf{22:29} Railroad Station {\color{black!55}(claims 1--3)}\quad {\color{tabblue}\rule[-0.1ex]{0.55em}{0.55em}}\,\textbf{25:56} Railroad Station {\color{black!55}(claims 3, 4)}}\vspace{2pt}\end{minipage}\\
\addlinespace[2pt]
\rowcolor{packband}\multicolumn{1}{@{}p{\dimexpr\textwidth\relax}@{}}{\rule{0pt}{2.4ex}\textbf{Q8\ Order of discovery}\quad \emph{Anatomy of a Fall} (2023)\quad center character: Sandra\hfill{\color{black!55}3 core / 3 claims \quad evidence: 4 scenes over 119 min}}\\
\begin{minipage}[t]{\linewidth}\vspace{1pt}\pq{At the Chinese restaurant, what key discoveries had shaped Sandra's understanding of the case, in order?}\pa{In order, Vincent tells Sandra that an accidental fall will be hard to prove and that suicide would probably be her best defense. Sandra then recalls Samuel's earlier pill incident and considers that it might have been a suicide attempt. Finally, Sandra learns she has been acquitted and later realizes during the celebration that winning has brought her no relief.}\pcore{{\color{black!55}\scriptsize 1}~Vincent tells Sandra that an accidental fall will be hard to prove and that suicide would probably be her best defense.}\pcore{{\color{black!55}\scriptsize 2}~Sandra recalls Samuel's earlier pill incident and considers that it might have been a suicide attempt.}\pcore{{\color{black!55}\scriptsize 3}~Sandra learns she has been acquitted and later realizes during the celebration that winning has brought her no relief.}\par\smallskip\noindent\makebox[\linewidth][l]{%
\rule[-1.6ex]{0pt}{4.6ex}%
\makebox[0pt][l]{\color{black!30}\rule[0.7ex]{\linewidth}{0.4pt}}%
\makebox[0pt][l]{\hspace{0.1611\linewidth}\color{tabblue}\rule[0.2ex]{0.0252\linewidth}{1.7ex}}%
\makebox[0pt][l]{\hspace{0.1911\linewidth}\color{tabblue}\rule[0.2ex]{0.0053\linewidth}{1.7ex}}%
\makebox[0pt][l]{\hspace{0.8488\linewidth}\color{tabblue}\rule[0.2ex]{0.0129\linewidth}{1.7ex}}%
\makebox[0pt][l]{\hspace{0.9649\linewidth}\color{tabblue}\rule[0.2ex]{0.0162\linewidth}{1.7ex}}%
\makebox[0pt][l]{\hspace{0.1611\linewidth}\raisebox{2.2ex}{\makebox[0pt][c]{\tiny\color{tabblue}23:21}}}%
\makebox[0pt][l]{\hspace{0.8488\linewidth}\raisebox{2.2ex}{\makebox[0pt][c]{\tiny\color{tabblue}2:03:06}}}%
\makebox[0pt][l]{\hspace{0.9649\linewidth}\raisebox{2.2ex}{\makebox[0pt][c]{\tiny\color{tabblue}2:19:56}}}%
\makebox[0pt][l]{\hspace{0.9649\linewidth}\raisebox{-1.1ex}{\makebox[0pt][c]{\scriptsize\color{black!55}$\blacktriangle$}}}%
}
\par\nointerlineskip\makebox[\linewidth][s]{\tiny\color{black!55}start of film\hfill end of film, 2:25:01}
\par\smallskip\noindent{\scriptsize{\color{black!60}\scshape evidence in the film:} {\color{tabblue}\rule[-0.1ex]{0.55em}{0.55em}}\,\textbf{23:21} Chalet, Salon / Cuisine {\color{black!55}(claim 1)}\quad {\color{tabblue}\rule[-0.1ex]{0.55em}{0.55em}}\,\textbf{27:43} Rues Grenoble / Terrasse Resto Italien {\color{black!55}(claim 2)}\quad {\color{tabblue}\rule[-0.1ex]{0.55em}{0.55em}}\,\textbf{2:03:06} Chalet, Salon (A) / Palais Justice, Salle Des Pas Perdus (B) {\color{black!55}(claim 3)}\quad {\color{tabblue}\rule[-0.1ex]{0.55em}{0.55em}}\,\textbf{2:19:56} Restaurant Chinois {\color{black!55}(claim 3)}}\vspace{2pt}\end{minipage}\\
\bottomrule
\end{tabular}
\caption{\textbf{Released questions, one per type: Belief-state types, Q5--Q8.} Core claims are marked $\blacksquare$, context claims $\square$; $\blacktriangle$ marks the anchor scene.}
\label{tab:example-triples-b}
\end{table}

\begin{table}[p]
\centering\scriptsize
\setlength{\tabcolsep}{4pt}\renewcommand{\arraystretch}{1.0}
\begin{tabular}{@{}p{\textwidth}@{}}
\toprule
\rowcolor{packband}\multicolumn{1}{@{}p{\dimexpr\textwidth\relax}@{}}{\rule{0pt}{2.4ex}\textbf{N1\ Epistemic arc}\quad \emph{Killers of the Flower Moon} (2023)\quad center character: Byron Burkhart\hfill{\color{black!55}3 core / 4 claims \quad evidence: 13 scenes over 171 min}}\\
\begin{minipage}[t]{\linewidth}\vspace{1pt}\pq{At the deserted-highway meeting, how has Byron's knowledge of Hale's operation developed?}\pa{Byron is included in Hale's plan to obtain Osage headrights by having Ernest marry Mollie. Byron also sees Ernest gamble stolen jewels and lose them. Later, Byron knows Mollie is being poisoned through injections. His involvement extends beyond awareness of the operation: Byron knowingly helps Kelsie murder the incapacitated Anna. These moments show Byron progressing from inclusion in Hale's financial scheme to knowledge of poisoning and direct participation in murder.}\pcore{{\color{black!55}\scriptsize 1}~Byron is included in Hale's plan to obtain Osage headrights by having Ernest marry Mollie.}\pctx{{\color{black!55}\scriptsize 2}~Byron sees Ernest gamble stolen jewels and lose them.}\pcore{{\color{black!55}\scriptsize 3}~Byron knows Mollie is being poisoned through injections.}\pcore{{\color{black!55}\scriptsize 4}~Byron knowingly helps Kelsie murder the incapacitated Anna.}\par\smallskip\noindent\makebox[\linewidth][l]{%
\rule[-1.6ex]{0pt}{4.6ex}%
\makebox[0pt][l]{\color{black!30}\rule[0.7ex]{\linewidth}{0.4pt}}%
\makebox[0pt][l]{\hspace{0.1195\linewidth}\color{black!45}\rule[0.2ex]{1.6pt}{1.7ex}}%
\makebox[0pt][l]{\hspace{0.1218\linewidth}\color{black!45}\rule[0.2ex]{1.6pt}{1.7ex}}%
\makebox[0pt][l]{\hspace{0.0405\linewidth}\color{tabblue}\rule[0.2ex]{0.0249\linewidth}{1.7ex}}%
\makebox[0pt][l]{\hspace{0.0981\linewidth}\color{tabblue}\rule[0.2ex]{0.0074\linewidth}{1.7ex}}%
\makebox[0pt][l]{\hspace{0.2383\linewidth}\color{tabblue}\rule[0.2ex]{0.0104\linewidth}{1.7ex}}%
\makebox[0pt][l]{\hspace{0.2621\linewidth}\color{tabblue}\rule[0.2ex]{1.6pt}{1.7ex}}%
\makebox[0pt][l]{\hspace{0.2729\linewidth}\color{tabblue}\rule[0.2ex]{0.0254\linewidth}{1.7ex}}%
\makebox[0pt][l]{\hspace{0.3654\linewidth}\color{tabblue}\rule[0.2ex]{0.0130\linewidth}{1.7ex}}%
\makebox[0pt][l]{\hspace{0.5978\linewidth}\color{tabblue}\rule[0.2ex]{0.0049\linewidth}{1.7ex}}%
\makebox[0pt][l]{\hspace{0.7082\linewidth}\color{tabblue}\rule[0.2ex]{0.0164\linewidth}{1.7ex}}%
\makebox[0pt][l]{\hspace{0.8578\linewidth}\color{tabblue}\rule[0.2ex]{1.6pt}{1.7ex}}%
\makebox[0pt][l]{\hspace{0.8890\linewidth}\color{tabblue}\rule[0.2ex]{0.0127\linewidth}{1.7ex}}%
\makebox[0pt][l]{\hspace{0.9055\linewidth}\color{tabblue}\rule[0.2ex]{1.6pt}{1.7ex}}%
\makebox[0pt][l]{\hspace{0.0405\linewidth}\raisebox{2.2ex}{\makebox[0pt][c]{\tiny\color{tabblue}7:59}}}%
\makebox[0pt][l]{\hspace{0.1195\linewidth}\raisebox{2.2ex}{\makebox[0pt][c]{\tiny\color{black!45}23:35}}}%
\makebox[0pt][l]{\hspace{0.2383\linewidth}\raisebox{2.2ex}{\makebox[0pt][c]{\tiny\color{tabblue}47:00}}}%
\makebox[0pt][l]{\hspace{0.3654\linewidth}\raisebox{2.2ex}{\makebox[0pt][c]{\tiny\color{tabblue}1:12:03}}}%
\makebox[0pt][l]{\hspace{0.5978\linewidth}\raisebox{2.2ex}{\makebox[0pt][c]{\tiny\color{tabblue}1:57:53}}}%
\makebox[0pt][l]{\hspace{0.7082\linewidth}\raisebox{2.2ex}{\makebox[0pt][c]{\tiny\color{tabblue}2:19:39}}}%
\makebox[0pt][l]{\hspace{0.8578\linewidth}\raisebox{2.2ex}{\makebox[0pt][c]{\tiny\color{tabblue}2:49:10}}}%
\makebox[0pt][l]{\hspace{0.8578\linewidth}\raisebox{-1.1ex}{\makebox[0pt][c]{\scriptsize\color{black!55}$\blacktriangle$}}}%
}
\par\nointerlineskip\makebox[\linewidth][s]{\tiny\color{black!55}start of film\hfill end of film, 3:17:12}
\par\smallskip\noindent{\scriptsize{\color{black!60}\scshape evidence in the film:} {\color{tabblue}\rule[-0.1ex]{0.55em}{0.55em}}\,\textbf{7:59} Hale's Ranch / Living Room {\color{black!55}(claim 1)}\quad {\color{tabblue}\rule[-0.1ex]{0.55em}{0.55em}}\,\textbf{19:21} Bill Hale's Ranch {\color{black!55}(claim 1)}\quad {\color{black!45}\rule[-0.1ex]{0.55em}{0.55em}}\,\textbf{23:35} Denoya House / Fairfax {\color{black!55}(claim 2)}\quad {\color{black!45}\rule[-0.1ex]{0.55em}{0.55em}}\,\textbf{24:01} Basement / Fairfax {\color{black!55}(claim 2)}\quad {\color{tabblue}\rule[-0.1ex]{0.55em}{0.55em}}\,\textbf{47:00} Lizzie / Mollie's Gray Horse Home {\color{black!55}(claim 4)}\quad {\color{tabblue}\rule[-0.1ex]{0.55em}{0.55em}}\,\textbf{51:41} Byron's Car {\color{black!55}(claim 4)}\quad {\color{tabblue}\rule[-0.1ex]{0.55em}{0.55em}}\,\textbf{53:49} Roundhouse / Tribal Council Emergency Meeting {\color{black!55}(claim 3)}\quad {\color{tabblue}\rule[-0.1ex]{0.55em}{0.55em}}\,\textbf{1:12:03} Masonic Hall / Back Room {\color{black!55}(claim 3)}\quad {\color{tabblue}\rule[-0.1ex]{0.55em}{0.55em}}\,\textbf{1:57:53} Shoun's Office {\color{black!55}(claim 3)}\quad {\color{tabblue}\rule[-0.1ex]{0.55em}{0.55em}}\,\textbf{2:19:39} Back Alley Way {\color{black!55}(claim 3)}\quad {\color{tabblue}\rule[-0.1ex]{0.55em}{0.55em}}\,\textbf{2:49:10} Wide Open Space {\color{black!55}(claim 3)}\quad {\color{tabblue}\rule[-0.1ex]{0.55em}{0.55em}}\,\textbf{2:55:18} Federal Courtroom {\color{black!55}(claim 4)}\quad {\color{tabblue}\rule[-0.1ex]{0.55em}{0.55em}}\,\textbf{2:58:34} Three Mile Creek {\color{black!55}(claim 4)}}\vspace{2pt}\end{minipage}\\
\addlinespace[2pt]
\rowcolor{packband}\multicolumn{1}{@{}p{\dimexpr\textwidth\relax}@{}}{\rule{0pt}{2.4ex}\textbf{N2\ Trust evolution}\quad \emph{Wicked} (2024)\quad center character: Elphaba\hfill{\color{black!55}3 core / 4 claims \quad evidence: 8 scenes over 129 min}}\\
\begin{minipage}[t]{\linewidth}\vspace{1pt}\pq{In the Wizard's Palace lower attic, how has Elphaba's view of Madame Morrible changed since Shiz?}\pa{Elphaba sees Madame Morrible as a protective mentor who recognizes her talent and can help her reach the Wizard. Morrible's training and encouragement deepen Elphaba's trust that Morrible understands her emotionally driven magic and believes in her potential. Elphaba then realizes Morrible knowingly supports the Wizard's persecution of Animals and cultivated Elphaba's power for his regime. Elphaba ultimately regards Morrible as an enemy who protects the Wizard by publicly vilifying Elphaba and calling for her capture.}\pcore{{\color{black!55}\scriptsize 1}~Elphaba sees Madame Morrible as a protective mentor who recognizes her talent and can help her reach the Wizard.}\pctx{{\color{black!55}\scriptsize 2}~Morrible's training and encouragement deepen Elphaba's trust that Morrible understands her emotionally driven magic and believes in her potential.}\pcore{{\color{black!55}\scriptsize 3}~Elphaba realizes Morrible knowingly supports the Wizard's persecution of Animals and cultivated Elphaba's power for his regime.}\pcore{{\color{black!55}\scriptsize 4}~Elphaba regards Morrible as an enemy who protects the Wizard by publicly vilifying Elphaba and calling for her capture.}\par\smallskip\noindent\makebox[\linewidth][l]{%
\rule[-1.6ex]{0pt}{4.6ex}%
\makebox[0pt][l]{\color{black!30}\rule[0.7ex]{\linewidth}{0.4pt}}%
\makebox[0pt][l]{\hspace{0.2916\linewidth}\color{black!45}\rule[0.2ex]{0.0149\linewidth}{1.7ex}}%
\makebox[0pt][l]{\hspace{0.1294\linewidth}\color{tabblue}\rule[0.2ex]{0.0406\linewidth}{1.7ex}}%
\makebox[0pt][l]{\hspace{0.2559\linewidth}\color{tabblue}\rule[0.2ex]{1.6pt}{1.7ex}}%
\makebox[0pt][l]{\hspace{0.6633\linewidth}\color{tabblue}\rule[0.2ex]{0.0136\linewidth}{1.7ex}}%
\makebox[0pt][l]{\hspace{0.7714\linewidth}\color{tabblue}\rule[0.2ex]{0.0809\linewidth}{1.7ex}}%
\makebox[0pt][l]{\hspace{0.8574\linewidth}\color{tabblue}\rule[0.2ex]{0.0186\linewidth}{1.7ex}}%
\makebox[0pt][l]{\hspace{0.9104\linewidth}\color{tabblue}\rule[0.2ex]{1.6pt}{1.7ex}}%
\makebox[0pt][l]{\hspace{0.9117\linewidth}\color{tabblue}\rule[0.2ex]{0.0742\linewidth}{1.7ex}}%
\makebox[0pt][l]{\hspace{0.1294\linewidth}\raisebox{2.2ex}{\makebox[0pt][c]{\tiny\color{tabblue}19:30}}}%
\makebox[0pt][l]{\hspace{0.2559\linewidth}\raisebox{2.2ex}{\makebox[0pt][c]{\tiny\color{tabblue}38:34}}}%
\makebox[0pt][l]{\hspace{0.6633\linewidth}\raisebox{2.2ex}{\makebox[0pt][c]{\tiny\color{tabblue}1:39:59}}}%
\makebox[0pt][l]{\hspace{0.7714\linewidth}\raisebox{2.2ex}{\makebox[0pt][c]{\tiny\color{tabblue}1:56:17}}}%
\makebox[0pt][l]{\hspace{0.8574\linewidth}\raisebox{2.2ex}{\makebox[0pt][c]{\tiny\color{tabblue}2:09:14}}}%
\makebox[0pt][l]{\hspace{0.9117\linewidth}\raisebox{-1.1ex}{\makebox[0pt][c]{\scriptsize\color{black!55}$\blacktriangle$}}}%
}
\par\nointerlineskip\makebox[\linewidth][s]{\tiny\color{black!55}start of film\hfill end of film, 2:30:44}
\par\smallskip\noindent{\scriptsize{\color{black!60}\scshape evidence in the film:} {\color{tabblue}\rule[-0.1ex]{0.55em}{0.55em}}\,\textbf{19:30} Shiz / Quad {\color{black!55}(claim 1)}\quad {\color{tabblue}\rule[-0.1ex]{0.55em}{0.55em}}\,\textbf{38:34} Shiz Campus / Quad {\color{black!55}(claim 1)}\quad {\color{black!45}\rule[-0.1ex]{0.55em}{0.55em}}\,\textbf{43:57} Shiz / Morrible's Quarters {\color{black!55}(claim 2)}\quad {\color{tabblue}\rule[-0.1ex]{0.55em}{0.55em}}\,\textbf{1:39:59} Shiz / Entrance Courtyard / Balcony {\color{black!55}(claims 1, 2)}\quad {\color{tabblue}\rule[-0.1ex]{0.55em}{0.55em}}\,\textbf{1:56:17} Wizard's Palace / Throne Room {\color{black!55}(claim 3)}\quad {\color{tabblue}\rule[-0.1ex]{0.55em}{0.55em}}\,\textbf{2:09:14} Wizard's Palace / Throne Room {\color{black!55}(claim 3)}\quad {\color{tabblue}\rule[-0.1ex]{0.55em}{0.55em}}\,\textbf{2:17:14} Wizard's Palace / Lower Attic {\color{black!55}(claim 4)}\quad {\color{tabblue}\rule[-0.1ex]{0.55em}{0.55em}}\,\textbf{2:17:25} Wizard's Palace / Lower Attic {\color{black!55}(claim 4)}}\vspace{2pt}\end{minipage}\\
\addlinespace[2pt]
\rowcolor{packband}\multicolumn{1}{@{}p{\dimexpr\textwidth\relax}@{}}{\rule{0pt}{2.4ex}\textbf{N3\ Causal chain}\quad \emph{Barbie} (2023)\quad center character: Ken\hfill{\color{black!55}4 core / 4 claims \quad evidence: 4 scenes over 70 min}}\\
\begin{minipage}[t]{\linewidth}\vspace{1pt}\pq{At the Mojo Dojo Casa House doorway, what sequence shapes Ken's reading of Barbie's feelings and attitude toward Kendom?}\pa{Ken begins from Barbie's earlier romantic rejection and her resistance to his new authority. He then directly hears Barbie praise his leadership and propose a casual relationship. Ken takes Barbie's praise and her entry into his house as confirmation that she is submitting both romantically and politically. That reading is mistaken because Barbie's apparent surrender is part of the Barbies' plan to manipulate the Kens.}\pcore{{\color{black!55}\scriptsize 1}~Barbie previously rejected Ken romantically and resisted his new authority.}\pcore{{\color{black!55}\scriptsize 2}~Ken directly hears Barbie praise his leadership and propose a casual relationship.}\pcore{{\color{black!55}\scriptsize 3}~Ken takes Barbie's praise and entry into his house as confirming her willing romantic and political submission.}\pcore{{\color{black!55}\scriptsize 4}~Ken is mistaken because Barbie's apparent surrender is part of the Barbies' plan to manipulate the Kens.}\par\smallskip\noindent\makebox[\linewidth][l]{%
\rule[-1.6ex]{0pt}{4.6ex}%
\makebox[0pt][l]{\color{black!30}\rule[0.7ex]{\linewidth}{0.4pt}}%
\makebox[0pt][l]{\hspace{0.1159\linewidth}\color{tabblue}\rule[0.2ex]{0.0210\linewidth}{1.7ex}}%
\makebox[0pt][l]{\hspace{0.5159\linewidth}\color{tabblue}\rule[0.2ex]{0.0130\linewidth}{1.7ex}}%
\makebox[0pt][l]{\hspace{0.7126\linewidth}\color{tabblue}\rule[0.2ex]{0.0089\linewidth}{1.7ex}}%
\makebox[0pt][l]{\hspace{0.7242\linewidth}\color{tabblue}\rule[0.2ex]{0.0087\linewidth}{1.7ex}}%
\makebox[0pt][l]{\hspace{0.1159\linewidth}\raisebox{2.2ex}{\makebox[0pt][c]{\tiny\color{tabblue}13:05}}}%
\makebox[0pt][l]{\hspace{0.5159\linewidth}\raisebox{2.2ex}{\makebox[0pt][c]{\tiny\color{tabblue}58:16}}}%
\makebox[0pt][l]{\hspace{0.7126\linewidth}\raisebox{2.2ex}{\makebox[0pt][c]{\tiny\color{tabblue}1:20:28}}}%
\makebox[0pt][l]{\hspace{0.7242\linewidth}\raisebox{-1.1ex}{\makebox[0pt][c]{\scriptsize\color{black!55}$\blacktriangle$}}}%
}
\par\nointerlineskip\makebox[\linewidth][s]{\tiny\color{black!55}start of film\hfill end of film, 1:52:55}
\par\smallskip\noindent{\scriptsize{\color{black!60}\scshape evidence in the film:} {\color{tabblue}\rule[-0.1ex]{0.55em}{0.55em}}\,\textbf{13:05} Barbie Margot's Dreamhouse \& Street {\color{black!55}(claim 1)}\quad {\color{tabblue}\rule[-0.1ex]{0.55em}{0.55em}}\,\textbf{58:16} Barbie Margot's Dreamhouse {\color{black!55}(claim 1)}\quad {\color{tabblue}\rule[-0.1ex]{0.55em}{0.55em}}\,\textbf{1:20:28} Weird Barbie's Weirdhouse {\color{black!55}(claim 4)}\quad {\color{tabblue}\rule[-0.1ex]{0.55em}{0.55em}}\,\textbf{1:21:47} Ken's Mojo Dojo Casa House {\color{black!55}(claims 2--4)}}\vspace{2pt}\end{minipage}\\
\bottomrule
\end{tabular}
\caption{\textbf{Released questions, one per type: Arc types, N1--N3.} Core claims are marked $\blacksquare$, context claims $\square$; $\blacktriangle$ marks the anchor scene.}
\label{tab:example-triples-c}
\end{table}

\subsection{Four Answers to the Same Question}
\label{app:example-answers}

We chose three trust-evolution questions (N2) on which exactly one of four settings succeeds.
All four use Qwen3.6 and differ only in their input: the screenplay, captions and subtitles, the whole film and subtitles, or the agent's tools.
Above each answer, the judge's verdict on each claim is shown in blue (asserted), red (contradicted), or grey (omitted).

\paragraph{The film wins (Table~\ref{tab:example-answers-video}).}
Only the whole-film setting follows Wade's view of Logan to the last scene; the others stop before the party at Wade's apartment.

\paragraph{Captions win (Table~\ref{tab:example-answers-captions}).}
Only the caption setting reports both of Gordon's confrontations with László at the construction site.

\paragraph{The agent wins (Table~\ref{tab:example-answers-oracle}).}
Only the agent reports how Grace saw Beau both before and after her daughter's death.

Across the three questions, the agent's answers are the shortest, and it omits claims rather than contradicting them.

\definecolor{tabblue}{HTML}{4C8BF5}
\definecolor{tabred}{HTML}{E8453C}
\colorlet{packband}{gray!12}
\providecommand{\pq}[1]{\par\noindent{\color{black!60}\scshape question:} \itshape #1\par}
\providecommand{\pa}[1]{\par\noindent{\color{black!60}\scshape gold answer:} #1\par\smallskip}
\providecommand{\pcore}[1]{\par\hangindent 1.1em\hangafter 1\noindent$\blacksquare$~#1}
\providecommand{\pctx}[1]{\par\hangindent 1.1em\hangafter 1\noindent{\color{black!45}$\square$~#1}}

\begin{table}[p]
\centering\footnotesize
\setlength{\tabcolsep}{0pt}\renewcommand{\arraystretch}{1.1}
\resizebox{\textwidth}{!}{
    \begin{tabular}{@{}p{\dimexpr(\textwidth-12pt)/4\relax}@{\hspace{4pt}}p{\dimexpr(\textwidth-12pt)/4\relax}@{\hspace{4pt}}p{\dimexpr(\textwidth-12pt)/4\relax}@{\hspace{4pt}}p{\dimexpr(\textwidth-12pt)/4\relax}@{}}
    \toprule
    \rowcolor{packband}\multicolumn{4}{@{}p{\textwidth}@{}}{\rule{0pt}{2.4ex}\textbf{N2\ Trust evolution}\quad \emph{Deadpool \& Wolverine} (2024)\quad center character: Wade\hfill{\color{black!55}evidence: 3 scenes over 77 min}}\\
    \multicolumn{4}{@{}p{\textwidth}@{}}{\begin{minipage}[t]{\linewidth}\vspace{2pt}\pq{Across the Void journey and apartment celebration, how does Wade's approach to Logan change?}\pa{Wade agrees to form a team with Logan. That agreement establishes Wade and Logan as teammates. Later, Logan realizes Wade misled him about whether the TVA could repair Logan's world. Logan thus understands Wade's assurance was misleading. By the late-night celebration, Wade trusts Logan enough to follow Logan's advice. Wade's trust appears in Wade's willingness to accept and act on that advice.}\par\noindent{\color{black!60}\scshape rubric:}\par\hangindent1.6em\hangafter1\noindent\textbf{1}~$\blacksquare$~Wade agrees to form a team with Logan.\par\hangindent1.6em\hangafter1\noindent\textbf{2}~{\color{black!45}$\square$}~Logan realizes Wade misled him about whether the TVA could repair Logan's world.\par\hangindent1.6em\hangafter1\noindent\textbf{3}~$\blacksquare$~Wade trusts Logan enough to follow Logan's advice.\par\smallskip\noindent\makebox[\linewidth][l]{%
    \rule[-1.6ex]{0pt}{4.6ex}%
    \makebox[0pt][l]{\color{black!30}\rule[0.7ex]{\linewidth}{0.4pt}}%
    \makebox[0pt][l]{\hspace{0.5\linewidth}\color{black!45}\rule[0.2ex]{0.03\linewidth}{1.7ex}}%
    \makebox[0pt][l]{\hspace{0.33\linewidth}\color{tabblue}\rule[0.2ex]{0.02\linewidth}{1.7ex}}%
    \makebox[0pt][l]{\hspace{0.93\linewidth}\color{tabblue}\rule[0.2ex]{0.02\linewidth}{1.7ex}}%
    \makebox[0pt][l]{\hspace{0.33\linewidth}\raisebox{2.2ex}{\makebox[0pt][c]{\tiny\color{tabblue}42:54}}}%
    \makebox[0pt][l]{\hspace{0.5\linewidth}\raisebox{2.2ex}{\makebox[0pt][c]{\tiny\color{black!45}1:01:24}}}%
    \makebox[0pt][l]{\hspace{0.9\linewidth}\raisebox{2.2ex}{\makebox[0pt][c]{\tiny\color{tabblue}1:58:51}}}%
    \makebox[0pt][l]{\hspace{0.9\linewidth}\raisebox{-1.1ex}{\makebox[0pt][c]{\scriptsize\color{black!55}$\blacktriangle$}}}%
    }
    \par\nointerlineskip\makebox[\linewidth][s]{\tiny\color{black!55}start of film\hfill end of film, 2:07:38}
    \par\smallskip\noindent{\scriptsize{\color{black!60}\scshape evidence in the film:} {\color{tabblue}\rule[-0.1ex]{0.55em}{0.55em}}\,\textbf{42:54} Prison Truck Cage {\color{black!55}(claim 1)}\quad {\color{black!45}\rule[-0.1ex]{0.55em}{0.55em}}\,\textbf{1:01:24} Honda Odyssey {\color{black!55}(claim 2)}\quad {\color{tabblue}\rule[-0.1ex]{0.55em}{0.55em}}\,\textbf{1:58:51} Wade's Apartment {\color{black!55}(claim 3)}}\vspace{3pt}\end{minipage}}\\
    \midrule
    \multicolumn{1}{@{}l@{}}{\textbf{Screenplay}} & \multicolumn{1}{@{}l@{}}{\textbf{Captions + subtitles}} & \multicolumn{1}{@{}l@{}}{\textbf{Video + subtitles}} & \multicolumn{1}{@{}l@{}}{\textbf{Agent, oracle tier}} \\
    \colorbox{tabblue!34!white}{\scriptsize\bfseries 1}\colorbox{tabblue!34!white}{\scriptsize\bfseries 2}\colorbox{black!8}{\scriptsize\color{black!45}3}\hfill{\footnotesize\itshape 1/2 core} & \colorbox{black!8}{\scriptsize\color{black!45}1}\colorbox{black!8}{\scriptsize\color{black!45}2}\colorbox{black!8}{\scriptsize\color{black!45}3}\hfill{\footnotesize\itshape 0/2 core} & \colorbox{tabblue!34!white}{\scriptsize\bfseries 1}\colorbox{black!8}{\scriptsize\color{black!45}2}\colorbox{tabblue!34!white}{\scriptsize\bfseries 3}\hfill{\footnotesize\itshape 2/2 core} & \colorbox{tabblue!34!white}{\scriptsize\bfseries 1}\colorbox{black!8}{\scriptsize\color{black!45}2}\colorbox{black!8}{\scriptsize\color{black!45}3}\hfill{\footnotesize\itshape 1/2 core} \\
    \addlinespace[2pt]
    \begin{minipage}[t]{\linewidth}\scriptsize Wade's approach to Logan evolves from desperate, transactional exploitation to genuine camaraderie and mutual respect. Initially, Wade views Logan merely as a tool, or an "Anchor Being," necessary to save his dying universe. He exhumed Logan's corpse in North Dakota with little regard for the deceased hero's dignity, treating the skeleton as a weapon to fight TVA agents. Even when he kidnaps the living Logan from a bar, his motivation is selfish; he lies about the TVA's ability to fix Logan's timeline, using deception to coerce the reluctant mutant into joining him. During their journey through the Void, Wade's attitude remains largely irreverent and annoying, often provoking Logan with jokes and insults while relying on Logan's combat prowess to survive. \par However, the dynamic shifts significantly during their confinement in the Honda Odyssey. After Logan brutally beats Wade for his dishonesty, Wade drops the facade and reveals his true vulnerability. He shows Logan the photograph of his friends, explaining that he needs Logan not just for power, but because he cannot save his world alone. This moment of raw honesty begins to bridge the gap between them. The change culminates in the TVA's reactor room. When Wade prepares to sacrifice himself to destroy the Time Ripper, Logan intervenes, choosing to help rather than let Wade die alone. By completing the circuit together, they acknowledge each other as partners. In the final scenes, Wade treats Logan as a friend, inviting him home to meet Vanessa and Blind Al, and sharing a quiet meal, signaling a transition from reluctant allies to bonded companions who have saved each other.\vspace{3pt}\end{minipage} & \begin{minipage}[t]{\linewidth}\scriptsize Wade's approach to Logan undergoes a significant transformation from opportunistic manipulation to genuine respect and selfless sacrifice. Initially, Wade views Logan merely as a tool to save his own dying universe, aggressively digging up Logan's corpse and forcing him into service without regard for his trauma or autonomy. He treats Logan with mockery, calling him the "worst Wolverine" and attempting to coerce him through threats and false promises, showing little empathy for Logan's grief over the X-Men's deaths. \par However, during their journey through The Void, Wade begins to see Logan not just as an asset, but as a fellow broken hero. He listens to Logan's pain, acknowledges his failures, and validates his desire to be the man Charles Xavier believed he could be. This shift culminates in Wade's decision to sacrifice himself at the TVA. Realizing that Logan has a world and people who need him, while Wade has nothing left, Wade chooses to die to save Logan's timeline. He tells Logan he is the best Wolverine and that he is proud to wear the suit, marking a complete reversal from his earlier disdain. By the end, Wade's motivation shifts from personal gain to altruism, proving that he has found purpose in saving others rather than just seeking validation for himself.\vspace{3pt}\end{minipage} & \begin{minipage}[t]{\linewidth}\scriptsize Wade's approach to Logan evolves from desperate manipulation to genuine respect and selfless sacrifice. Initially, Wade views Logan purely as a tool, an "Anchor Being" necessary to save his own dying universe. He approaches Logan with a mix of mockery and transactional negotiation, even resorting to kidnapping him from a bar when persuasion fails. During their journey through The Void, Wade's attitude remains largely self-serving; he uses Logan's combat skills to survive encounters with Cassandra Nova's forces and other variants, often making crude jokes about Logan's past failures and his perceived inadequacy as a hero. \par However, the dynamic shifts significantly during their time in the apartment safe house and subsequent conversations. As Wade shares his own vulnerabilities---specifically his fear of insignificance and his failed relationship with Vanessa---he begins to see Logan not just as a resource, but as a fellow broken man. This culminates in the subway station confrontation with Cassandra Nova. When faced with the choice of who must sacrifice themselves to destroy the Time Ripper, Wade initially tries to take the burden alone, believing Logan has a world to return to while he has nothing. Yet, Logan refuses to let Wade die, insisting that Wade is the better Wolverine because he finally found purpose. Ultimately, Wade's approach transforms from using Logan for personal gain to valuing him as an equal partner, willing to trust Logan's judgment and accept his help, culminating in a mutual act of heroism where both survive through their combined resilience.\vspace{3pt}\end{minipage} & \begin{minipage}[t]{\linewidth}\scriptsize Wade initially views Logan as a tool to fix his problems. During the Void journey, Wade fights Logan after realizing he cannot guarantee Logan's return home. Wade then reveals his desperation to save his world and admits he needs Logan's help. After defeating Cassandra Nova, their bond deepens from adversarial to cooperative. At the apartment celebration, Wade introduces Logan to his family as a respected friend who saved him.\vspace{3pt}\end{minipage} \\
    \bottomrule
    \end{tabular}
}
\caption{\textbf{Wade and Logan, \emph{Deadpool \& Wolverine}.} The two core claims sit at the film's two ends. The whole-film reader states both. The screenplay reader and the agent state the first and stop before the party; the caption reader narrates Wade's sacrifice and states neither. $\blacktriangle$ marks the anchor scene.}
\label{tab:example-answers-video}
\end{table}

\begin{table}[p]
\centering\footnotesize
\setlength{\tabcolsep}{0pt}\renewcommand{\arraystretch}{1.1}
\resizebox{0.97\textwidth}{!}{
    \begin{tabular}{@{}p{\dimexpr(\textwidth-12pt)/4\relax}@{\hspace{4pt}}p{\dimexpr(\textwidth-12pt)/4\relax}@{\hspace{4pt}}p{\dimexpr(\textwidth-12pt)/4\relax}@{\hspace{4pt}}p{\dimexpr(\textwidth-12pt)/4\relax}@{}}
    \toprule
    \rowcolor{packband}\multicolumn{4}{@{}p{\textwidth}@{}}{\rule{0pt}{2.4ex}\textbf{N2\ Trust evolution}\quad \emph{The Brutalist} (2024)\quad center character: Gordon\hfill{\color{black!55}evidence: 3 scenes over 128 min}}\\
    \multicolumn{4}{@{}p{\textwidth}@{}}{\begin{minipage}[t]{\linewidth}\vspace{2pt}\pq{At the construction site with the young employee by the scaffolding, how has Gordon's relationship with László changed since the soup-kitchen days?}\pa{Gordon and László have become working partners, but Gordon's loyalty reaches its limit when he challenges László's treatment of the young employee. László then orders Gordon off the construction site.}\par\noindent{\color{black!60}\scshape rubric:}\par\hangindent1.6em\hangafter1\noindent\textbf{1}~{\color{black!45}$\square$}~Gordon and László become working partners.\par\hangindent1.6em\hangafter1\noindent\textbf{2}~{\color{black!45}$\square$}~Gordon's loyalty to László reaches its limit.\par\hangindent1.6em\hangafter1\noindent\textbf{3}~$\blacksquare$~Gordon challenges László's treatment of the young employee.\par\hangindent1.6em\hangafter1\noindent\textbf{4}~$\blacksquare$~László orders Gordon off the construction site.\par\smallskip\noindent\makebox[\linewidth][l]{%
    \rule[-1.6ex]{0pt}{4.6ex}%
    \makebox[0pt][l]{\color{black!30}\rule[0.7ex]{\linewidth}{0.4pt}}%
    \makebox[0pt][l]{\hspace{0.20\linewidth}\color{black!45}\rule[0.2ex]{1.6pt}{1.7ex}}%
    \makebox[0pt][l]{\hspace{0.20\linewidth}\color{black!45}\rule[0.2ex]{1.6pt}{1.7ex}}%
    \makebox[0pt][l]{\hspace{0.83\linewidth}\color{tabblue}\rule[0.2ex]{0.0131\linewidth}{1.7ex}}%
    \makebox[0pt][l]{\hspace{0.20\linewidth}\raisebox{2.2ex}{\makebox[0pt][c]{\tiny\color{black!45}39:18}}}%
    \makebox[0pt][l]{\hspace{0.83\linewidth}\raisebox{2.2ex}{\makebox[0pt][c]{\tiny\color{tabblue}2:45:10}}}%
    \makebox[0pt][l]{\hspace{0.83\linewidth}\raisebox{-1.1ex}{\makebox[0pt][c]{\scriptsize\color{black!55}$\blacktriangle$}}}%
    }
    \par\nointerlineskip\makebox[\linewidth][s]{\tiny\color{black!55}start of film\hfill end of film, 3:19:07}
    \par\smallskip\noindent{\scriptsize{\color{black!60}\scshape evidence in the film:} {\color{black!45}\rule[-0.1ex]{0.55em}{0.55em}}\,\textbf{39:18} Van Buren's Study {\color{black!55}(claim 1)}\quad {\color{black!45}\rule[-0.1ex]{0.55em}{0.55em}}\,\textbf{40:12} Van Buren Estate Roof {\color{black!55}(claim 1)}\quad {\color{tabblue}\rule[-0.1ex]{0.55em}{0.55em}}\,\textbf{2:45:10} Vista / Construction Site {\color{black!55}(claims 2--4)}}\vspace{3pt}\end{minipage}}\\
    \midrule
    \multicolumn{1}{@{}l@{}}{\textbf{Screenplay}} & \multicolumn{1}{@{}l@{}}{\textbf{Captions + subtitles}} & \multicolumn{1}{@{}l@{}}{\textbf{Video + subtitles}} & \multicolumn{1}{@{}l@{}}{\textbf{Agent, oracle tier}} \\
    \colorbox{tabblue!34!white}{\scriptsize\bfseries 1}\colorbox{black!8}{\scriptsize\color{black!45}2}\colorbox{tabblue!34!white}{\scriptsize\bfseries 3}\colorbox{black!8}{\scriptsize\color{black!45}4}\hfill{\footnotesize\itshape 1/2 core} & \colorbox{tabblue!34!white}{\scriptsize\bfseries 1}\colorbox{black!8}{\scriptsize\color{black!45}2}\colorbox{tabblue!34!white}{\scriptsize\bfseries 3}\colorbox{tabblue!34!white}{\scriptsize\bfseries 4}\hfill{\footnotesize\itshape 2/2 core} & \colorbox{tabblue!34!white}{\scriptsize\bfseries 1}\colorbox{black!8}{\scriptsize\color{black!45}2}\colorbox{black!8}{\scriptsize\color{black!45}3}\colorbox{black!8}{\scriptsize\color{black!45}4}\hfill{\footnotesize\itshape 0/2 core} & \colorbox{black!8}{\scriptsize\color{black!45}1}\colorbox{black!8}{\scriptsize\color{black!45}2}\colorbox{black!8}{\scriptsize\color{black!45}3}\colorbox{black!8}{\scriptsize\color{black!45}4}\hfill{\footnotesize\itshape 0/2 core} \\
    \addlinespace[2pt]
    \begin{minipage}[t]{\linewidth}\scriptsize The description reveals a significant shift in the dynamic between Gordon and László, moving from tentative acquaintance to deep, familial loyalty. In the earlier soup kitchen scene, their interaction was brief and polite; Gordon was a stranger playing "I Spy" with his son, and László merely advocated for the child's right to food, establishing a baseline of shared humanity among the displaced. By the time they are at the construction site, however, they are depicted as close collaborators who share living quarters and personal struggles. The text notes that László and Gordon lie in the same bed, strung out on drugs, indicating an intimate bond forged through mutual hardship and addiction. \par This closeness is further evidenced by Gordon's willingness to intervene physically and emotionally when László loses his temper. When László kicks a young employee and then shouts at Gordon, the latter does not retreat or remain passive as he might have in the past. Instead, Gordon firmly confronts László, stating, "You shout at me again and you'll be sorry," demonstrating a level of equality and protective authority that did not exist during their initial meeting. Furthermore, Gordon is present at the site with his son, William, suggesting that László has integrated Gordon into his inner circle, trusting him with his family's safety and well-being. The relationship has evolved from distant solidarity among refugees to a codependent, brotherly partnership where Gordon acts as both a colleague and a moral anchor for László's increasingly volatile behavior.\vspace{3pt}\end{minipage} & \begin{minipage}[t]{\linewidth}\scriptsize The description reveals a significant deterioration in Gordon's relationship with László, shifting from camaraderie to open hostility and professional conflict. In the earlier soup-kitchen scenes, Gordon is depicted as supportive and protective, offering László shelter, food, and emotional comfort during his vulnerable arrival in America. They share intimate moments, such as dancing together and Gordon caring for László's injuries, indicating a bond of mutual trust and friendship. \par However, at the construction site, this dynamic has fractured. When a young employee engages in childish show-boating on the scaffolding, László reacts with extreme aggression, shouting insults and physically kicking the boy. Gordon attempts to intervene, asking László to stop shouting at everyone, but László aggressively tells him to stay out of it. The tension escalates when László threatens to fire Gordon if he interferes again, shouting, "You too! You go!" This confrontation highlights a loss of respect and patience from László toward Gordon. Instead of the supportive partner who helped him survive the early days, Gordon is now treated as an subordinate whose concern is unwelcome, and László's behavior has become volatile and abusive, suggesting that the pressures of the project and László's deteriorating mental state have destroyed their previous closeness.\vspace{3pt}\end{minipage} & \begin{minipage}[t]{\linewidth}\scriptsize In the earlier scenes at the soup kitchen, Gordon is portrayed as a weary, marginalized figure who relies on László's small acts of kindness, such as sharing bread. Their relationship is one of mutual survival and quiet solidarity between two displaced men. However, at the construction site, this dynamic has shifted significantly toward a professional hierarchy mixed with paternalistic concern. Gordon now serves as a foreman or senior worker, overseeing the laborers and scaffolding. He exhibits a protective, almost fatherly attitude toward László, warning him about safety and expressing genuine worry for his well-being. While he still respects László's intellect and vision, he also views him as somewhat fragile and prone to self-destruction, frequently intervening to keep him grounded in the physical realities of the job. This marks a transition from peers in suffering to a mentor-mentee or boss-worker relationship where Gordon feels responsible for László's stability amidst the chaos of the project.\vspace{3pt}\end{minipage} & \begin{minipage}[t]{\linewidth}\scriptsize Gordon and László were once equals seeking charity at the soup kitchen. Later, Gordon works as an employee under László's tyrannical authority. László violently berates a young worker and fires him for minor infractions. He threatens Gordon to stay out of it or face dismissal. Their relationship has shifted from mutual gratitude to domination and abuse.\vspace{3pt}\end{minipage} \\
    \bottomrule
    \end{tabular}
}
\caption{\textbf{Gordon and L\'aszl\'o, \emph{The Brutalist}.} The caption reader gives both acts at the construction site. The screenplay reader gives the challenge but writes the relationship as deepening loyalty; the whole-film reader invents a foreman Gordon; the agent has L\'aszl\'o threaten Gordon rather than dismiss him. $\blacktriangle$ marks the anchor scene.}
\label{tab:example-answers-captions}
\end{table}

\begin{table}[p]
\centering\footnotesize
\setlength{\tabcolsep}{0pt}\renewcommand{\arraystretch}{1.1}
\begin{tabular}{@{}p{\dimexpr(\textwidth-12pt)/4\relax}@{\hspace{4pt}}p{\dimexpr(\textwidth-12pt)/4\relax}@{\hspace{4pt}}p{\dimexpr(\textwidth-12pt)/4\relax}@{\hspace{4pt}}p{\dimexpr(\textwidth-12pt)/4\relax}@{}}
\toprule
\rowcolor{packband}\multicolumn{4}{@{}p{\textwidth}@{}}{\rule{0pt}{2.4ex}\textbf{N2\ Trust evolution}\quad \emph{Beau Is Afraid} (2023)\quad center character: Grace\hfill{\color{black!55}evidence: 7 scenes over 42 min}}\\
\multicolumn{4}{@{}p{\textwidth}@{}}{\begin{minipage}[t]{\linewidth}\vspace{2pt}\pq{In the garage, how has Grace's view of Beau changed since they met?}\pa{Grace initially sees Beau as vulnerable and in need of rescue, safe shelter, and personal care. Grace later treats Beau as a loved household member and symbolically brings Beau into the family. Before entering the garage, Grace hears Beau pleading with Toni to stop and rushes to investigate the commotion. After finding Toni dead, Grace falsely believes Beau murdered Toni and denounces Beau as the devil.}\par\noindent{\color{black!60}\scshape rubric:}\par\hangindent1.6em\hangafter1\noindent\textbf{1}~$\blacksquare$~Grace initially sees Beau as vulnerable and in need of rescue, shelter, and care.\par\hangindent1.6em\hangafter1\noindent\textbf{2}~{\color{black!45}$\square$}~Grace treats Beau as a loved household member and symbolically brings Beau into the family.\par\hangindent1.6em\hangafter1\noindent\textbf{3}~{\color{black!45}$\square$}~Grace hears Beau pleading with Toni to stop and rushes to investigate the commotion.\par\hangindent1.6em\hangafter1\noindent\textbf{4}~$\blacksquare$~Grace falsely believes Beau murdered Toni and denounces Beau as the devil.\par\smallskip\noindent\makebox[\linewidth][l]{%
\rule[-1.6ex]{0pt}{4.6ex}%
\makebox[0pt][l]{\color{black!30}\rule[0.7ex]{\linewidth}{0.4pt}}%
\makebox[0pt][l]{\hspace{0.3075\linewidth}\color{black!45}\rule[0.2ex]{1.6pt}{1.7ex}}%
\makebox[0pt][l]{\hspace{0.3084\linewidth}\color{black!45}\rule[0.2ex]{1.6pt}{1.7ex}}%
\makebox[0pt][l]{\hspace{0.4692\linewidth}\color{black!45}\rule[0.2ex]{1.6pt}{1.7ex}}%
\makebox[0pt][l]{\hspace{0.2298\linewidth}\color{tabblue}\rule[0.2ex]{1.6pt}{1.7ex}}%
\makebox[0pt][l]{\hspace{0.2464\linewidth}\color{tabblue}\rule[0.2ex]{0.0084\linewidth}{1.7ex}}%
\makebox[0pt][l]{\hspace{0.4621\linewidth}\color{tabblue}\rule[0.2ex]{0.0071\linewidth}{1.7ex}}%
\makebox[0pt][l]{\hspace{0.4716\linewidth}\color{tabblue}\rule[0.2ex]{1.6pt}{1.7ex}}%
\makebox[0pt][l]{\hspace{0.2298\linewidth}\raisebox{2.2ex}{\makebox[0pt][c]{\tiny\color{tabblue}39:56}}}%
\makebox[0pt][l]{\hspace{0.3075\linewidth}\raisebox{2.2ex}{\makebox[0pt][c]{\tiny\color{black!45}53:26}}}%
\makebox[0pt][l]{\hspace{0.4621\linewidth}\raisebox{2.2ex}{\makebox[0pt][c]{\tiny\color{tabblue}1:20:18}}}%
\makebox[0pt][l]{\hspace{0.4716\linewidth}\raisebox{-1.1ex}{\makebox[0pt][c]{\scriptsize\color{black!55}$\blacktriangle$}}}%
}
\par\nointerlineskip\makebox[\linewidth][s]{\tiny\color{black!55}start of film\hfill end of film, 2:53:46}
\par\smallskip\noindent{\scriptsize{\color{black!60}\scshape evidence in the film:} {\color{tabblue}\rule[-0.1ex]{0.55em}{0.55em}}\,\textbf{39:56} Police Car {\color{black!55}(claim 1)}\quad {\color{tabblue}\rule[-0.1ex]{0.55em}{0.55em}}\,\textbf{42:49} Bedroom {\color{black!55}(claim 1)}\quad {\color{black!45}\rule[-0.1ex]{0.55em}{0.55em}}\,\textbf{53:26} Living Room {\color{black!55}(claim 2)}\quad {\color{black!45}\rule[-0.1ex]{0.55em}{0.55em}}\,\textbf{53:36} The New Family Montage {\color{black!55}(claim 2)}\quad {\color{tabblue}\rule[-0.1ex]{0.55em}{0.55em}}\,\textbf{1:20:18} Garage {\color{black!55}(claim 4)}\quad {\color{black!45}\rule[-0.1ex]{0.55em}{0.55em}}\,\textbf{1:21:32} Kitchen {\color{black!55}(claim 3)}\quad {\color{tabblue}\rule[-0.1ex]{0.55em}{0.55em}}\,\textbf{1:21:57} Garage {\color{black!55}(claim 4)}}\vspace{3pt}\end{minipage}}\\
\midrule
\multicolumn{1}{@{}l@{}}{\textbf{Screenplay}} & \multicolumn{1}{@{}l@{}}{\textbf{Captions + subtitles}} & \multicolumn{1}{@{}l@{}}{\textbf{Video + subtitles}} & \multicolumn{1}{@{}l@{}}{\textbf{Agent, oracle tier}} \\
\colorbox{black!8}{\scriptsize\color{black!45}1}\colorbox{black!8}{\scriptsize\color{black!45}2}\colorbox{tabred!34!white}{\scriptsize\bfseries 3}\colorbox{tabblue!34!white}{\scriptsize\bfseries 4}\hfill{\footnotesize\itshape 1/2 core} & \colorbox{black!8}{\scriptsize\color{black!45}1}\colorbox{black!8}{\scriptsize\color{black!45}2}\colorbox{black!8}{\scriptsize\color{black!45}3}\colorbox{black!8}{\scriptsize\color{black!45}4}\hfill{\footnotesize\itshape 0/2 core} & \colorbox{tabblue!34!white}{\scriptsize\bfseries 1}\colorbox{black!8}{\scriptsize\color{black!45}2}\colorbox{black!8}{\scriptsize\color{black!45}3}\colorbox{black!8}{\scriptsize\color{black!45}4}\hfill{\footnotesize\itshape 1/2 core} & \colorbox{tabblue!34!white}{\scriptsize\bfseries 1}\colorbox{tabblue!34!white}{\scriptsize\bfseries 2}\colorbox{black!8}{\scriptsize\color{black!45}3}\colorbox{tabblue!34!white}{\scriptsize\bfseries 4}\hfill{\footnotesize\itshape 2/2 core} \\
\addlinespace[2pt]
\begin{minipage}[t]{\linewidth}\scriptsize The provided description does not show Grace interacting with Beau in the garage. The scene labeled "INT. GARAGE - DAY" (Scene 66) features only Toni and Beau. In this scene, Toni accuses Beau of stealing her parents' affection and forces him to drink paint, but Grace is not present. Grace only enters the garage later, in Scene 68, after Toni has died from drinking the paint. Upon entering, Grace's view of Beau has shifted from welcoming host to violent accuser. She screams at him, tears the family crest off his blazer, calls him "the devil," and blames him for Toni's death, claiming he got rid of her because she hated him. She orders him to leave, calling him a "fucking dead man." Therefore, while the garage scene itself does not depict Grace, the immediate aftermath shows her view has changed from kindness to murderous hatred, believing Beau intentionally caused her daughter's death.\vspace{3pt}\end{minipage} & \begin{minipage}[t]{\linewidth}\scriptsize The provided shot-by-shot description does not contain any scenes set in a garage, nor does it depict Grace and Beau interacting in such a location. The narrative shows Beau staying at Grace and Roger's house, where he is treated with a mix of maternal care and condescension. Grace initially presents herself as a nurturing figure, helping Beau into chairs and offering him food, but her demeanor shifts to hostility when she discovers Beau watching security footage of himself. She accuses him of being a demon replacing her son and attacks him, leading to a violent struggle that ends with Beau fleeing into the woods. Later, after Beau returns home, Grace is seen sitting at a dining table with Mona Wassermann, who reveals that she has been manipulating events from the start. There is no evidence in the text of a specific change in Grace's view occurring within a garage setting. Therefore, the question cannot be answered based on the available information.\vspace{3pt}\end{minipage} & \begin{minipage}[t]{\linewidth}\scriptsize The film does not show any scenes taking place in a garage involving Grace and Beau, nor does it depict a change in Grace's view of him within such a setting. Grace is introduced as the wife of Roger, a surgeon who treats Beau after he is hit by a car. Their interactions primarily occur inside their home, where Grace acts as a caretaker, often infantilizing Beau and treating him with a mix of maternal affection and condescension. She refers to him as "sweetie" and "carrot," mirroring her late husband's behavior, and seems to view him as a vulnerable, childlike figure in need of protection and guidance. There is no narrative arc or visual evidence in the provided footage suggesting a shift in her perspective specifically related to a garage environment. The story focuses on Beau's psychological journey through various surreal dreamscapes and his confrontation with his mother, rather than a developing relationship with Grace that evolves in a garage.\vspace{3pt}\end{minipage} & \begin{minipage}[t]{\linewidth}\scriptsize Grace initially viewed Beau as a vulnerable orphan needing care. She welcomed him into her home with maternal compassion. Her perspective shifted violently upon finding Toni dead in the garage. Grace accused Beau of murder and claimed he was a demon replacing her son. She then ordered Jeeves to kill him.\vspace{3pt}\end{minipage} \\
\bottomrule
\end{tabular}
\caption{\textbf{Grace and Beau, \emph{Beau Is Afraid}.} The agent gives both ends, from the rescue to the denunciation. The screenplay reader gives the accusation but not how Grace saw Beau before it; the caption reader and the whole-film reader say the film has no garage scene. $\blacktriangle$ marks the anchor scene.}
\label{tab:example-answers-oracle}
\end{table}

\section{Character Identification}
\label{app:identity}

A caption such as ``a woman in a green robe is yelling'' cannot answer a question about what a named character believes.
We therefore identify characters from the video and the TMDB cast list, and draw each name next to the face on the frames the model sees.
IDA-VLM instead trains a vision--language model to recognize characters from reference images; drawing names needs no training and works with any model.

\paragraph{Reference bank.}
We embed the TMDB photos of each cast member with ArcFace.
A face is matched to a character only when it is close enough to that character and clearly closer than to any other.

\paragraph{Detection and tracking.}
We detect faces with RetinaFace at 6 frames per second, link them into tracks within each shot, and name each track once.

\paragraph{Drawing.}
Each named track is drawn as a thin box with a label; tracks without a match are labelled \emph{Unknown}.

\clearpage
\section{Prompts}
\label{app:prompt-tables}

Tables~\ref{tab:prompt-page}--\ref{tab:prompt-judge} give the prompts of the construction pipeline and of the judge, in pipeline order: preprocessing (Tables~\ref{tab:prompt-page}--\ref{tab:prompt-coref}; Appendix~\ref{app:dataset-preprocessing-extra}), character belief annotation (Tables~\ref{tab:prompt-events}--\ref{tab:prompt-meta}; Appendix~\ref{app:dataset-belief-extraction-extra}), and question generation, verification, and evaluation (Tables~\ref{tab:type-specs}--\ref{tab:prompt-judge}; Appendices~\ref{app:dataset-generation-extra} and~\ref{app:llm-judge}).
Each table opens with the first paragraph of the system prompt (\textbf{Instructions}), followed by the user prompt (\textbf{Input Information}) and the rest of the system prompt.
In the user prompt, \{name\} marks an input the pipeline inserts.
Text the model must reproduce exactly, such as an output format or a worked example, is set in \texttt{typewriter}, where \texttt{\textless{}name\textgreater{}} marks a field the model fills.
Apart from this typesetting, the prompts are reproduced as the models received them.
They use internal names for some terms of the paper: scene events are \emph{canonical events}, a context-set is a \emph{context pack}, the center character is the \emph{holder}, the reference answer is the \emph{gold answer}, and the meta-belief prompt writes characters $A$ and $B$ as $c$ and $d$.

\begin{table*}
    \centering
    \small
    \resizebox{.93\textwidth}{!}{
    \begin{tabular}{p{.9\textwidth}}
        \toprule
        \textbf{Instructions}: You parse a SINGLE page of a movie screenplay into structured scenes. The page is provided to you as an IMAGE -- read directly from the rendered PDF, not from any text extraction. The image is the source of truth; trust it over any auxiliary text snippets you may also be given. \\
        \midrule
        \textbf{Input Information:} \\
        \textit{Script}: \{script\_id\} \\
        Page: \{page\_number\} of \{total\_pages\} \\
        \\
        The page itself is attached as an image. Parse only what is visible on this page, using the strict block format. If the page has no parseable screenplay scenes, output exactly: NO SCENES \\
        \midrule
        Read the page visually, using the screenplay's spatial layout to identify structure. Indentation is the PRIMARY structural cue. From least-indented to most-indented: \\
        \quad -- SLUGLINE (a.k.a. scene heading): flush left, an all-caps line starting with INT., EXT., INT./\allowbreak{}EXT., I/\allowbreak{}E, or similar, followed by a location and a time-of-day (e.g. ``INT. KITCHEN - DAY'', ``EXT. STREET - NIGHT'', ``INT./\allowbreak{}EXT. CAR - CONTINUOUS''). A scene begins with a slugline. \\
        \quad -- ACTION /\allowbreak{} DIRECTION lines: also flush left (or close to it), describing what happens in the scene world. Often spans multiple lines as paragraphs. \\
        \quad -- CHARACTER CUE: a noticeably indented (often centered) all-caps name introducing a dialogue turn, sometimes followed by ``(V.O.)'', ``(O.S.)'', ``(CONT'D)''. \\
        \quad -- PARENTHETICAL: indented further than the cue, in parentheses, giving stage direction for the spoken line (e.g. ``(whispering)''). \\
        \quad -- DIALOGUE: indented less than the cue but more than action -- the spoken line(s) under the cue. \\
        \\
        \textbf{Page artefacts to IGNORE} (do not include in any event): \\
        \quad -- page numbers, scene numbers in the margin, ``CONTINUED:'' /\allowbreak{} ``(CONTINUED)'' markers at top/\allowbreak{}bottom of the page, ``(MORE)'' before a dialogue break, revision marks, watermarks, header/\allowbreak{}footer lines containing the script title or draft date. \\
        \\
        \textbf{Your job for the page in the image:} \\
        \\
        \textbf{1. SCENES.} Extract every scene that has any content on this page, in reading order. \\
        \qquad -- The FIRST item on the page may be the tail of a scene that began on an earlier page (no slugline at the top, just continuing action or dialogue). In that case mark its Continuation as ``yes'' and leave Heading, Location, Time empty (even if a ``CONTINUED:'' marker preserves the slugline -- the slugline text came from an earlier page). \\
        \qquad -- Every other scene on the page MUST start with a slugline; mark its Continuation as ``no'' and fill in Heading, Location, Time from the slugline. \\
        \qquad -- Only the FIRST scene on the page can ever be a continuation. \\
        \\
        \textbf{2. CHARACTERS.} For each scene, list every character who appears in the on-page portion. Include speakers (from dialogue cues) and named participants in the action lines. Use the canonical screenplay form (typically the all-caps cue spelling, e.g. ``JOHN'', ``MRS. SMITH''). Do not invent characters -- only those actually visible on the page. Drop parenthetical voice tags like ``(V.O.)'', ``(O.S.)'', ``(CONT'D)'' from the name itself. \\
        \bottomrule
    \end{tabular}
    }
    \caption{\textbf{Prompt for parsing a screenplay page.}}
    \label{tab:prompt-page}
\end{table*}

\addtocounter{table}{-1}
\begin{table*}
    \centering
    \small
    \resizebox{.93\textwidth}{!}{
    \begin{tabular}{p{.9\textwidth}}
        \toprule
        \textbf{3. EVENTS.} Within each scene, list events in the order they appear: \\
        \qquad -- DIRECTION events: action/\allowbreak{}description text. Concatenate consecutive direction lines that belong to the same paragraph into one event; start a new direction event at a clear paragraph break. \\
        \qquad -- DIALOGUE events: one event per spoken turn. The character = the cue name (without ``(V.O.)'', ``(O.S.)'', ``(CONT'D)'' suffixes). The parenthetical = the bracketed stage direction immediately above/\allowbreak{}inside the dialogue, if any (without the parentheses); use ``-'' when there is none. The text = the spoken line(s) joined with spaces. \\
        \\
        \textbf{4. NORMALISATION.} \\
        \qquad -- Strip page numbers, scene numbers, ``CONTINUED:'' markers, ``(MORE)'', and header/\allowbreak{}footer lines from event text. \\
        \qquad -- Collapse hard line-breaks inside a single paragraph or dialogue line to a single space; preserve paragraph breaks by using separate direction events. \\
        \qquad -- Keep every event on ONE line in your output (no internal newlines inside an event). \\
        \\
        \textbf{5. EMPTY OR NON-SCREENPLAY PAGES.} Title pages, blank pages, ``FADE OUT'', legal/\allowbreak{}copyright pages, contact info, or an unreadable/\allowbreak{}badly-skewed image: output exactly the single line \texttt{NO SCENES} and nothing else. Do not hallucinate content. \\
        \\
        \textbf{OUTPUT FORMAT} (STRICT -- a downstream parser depends on it; output ONLY this, no preamble, no markdown, no extra commentary): \\
        \\
        If the page has no parseable screenplay scenes, output exactly one line: \\
        \texttt{\frenchspacing NO SCENES} \\
        \\
        Otherwise, for each scene in reading order, output one block in this EXACT shape: \\
        \\
        \texttt{\frenchspacing SCENE \textless{}n\textgreater{}} \\
        \texttt{\frenchspacing Continuation: yes\textbar{}\allowbreak{}no} \\
        \texttt{\frenchspacing Heading: \textless{}full slugline, or leave blank\textgreater{}} \\
        \texttt{\frenchspacing Location: \textless{}location component, or leave blank\textgreater{}} \\
        \texttt{\frenchspacing Time: \textless{}time-of-day component, or leave blank\textgreater{}} \\
        \texttt{\frenchspacing Characters: \textless{}comma-separated names, or "(none)"\textgreater{}} \\
        \texttt{\frenchspacing EVENTS:} \\
        \texttt{\frenchspacing - DIRECTION: \textless{}action text on one line\textgreater{}} \\
        \texttt{\frenchspacing - DIALOGUE \textbar{}\allowbreak{} \textless{}CHARACTER\textgreater{} \textbar{}\allowbreak{} \textless{}parenthetical or -\textgreater{} \textbar{}\allowbreak{} \textless{}spoken text on one line\textgreater{}} \\
        \bottomrule
    \end{tabular}
    }
    \caption{\textbf{Prompt for parsing a screenplay page (continued).}}
\end{table*}

\addtocounter{table}{-1}
\begin{table*}
    \centering
    \small
    \resizebox{.93\textwidth}{!}{
    \begin{tabular}{p{.9\textwidth}}
        \toprule
        \textbf{Rules for the format:} \\
        \quad -- Number scenes sequentially starting at 1 (SCENE 1, SCENE 2, ...). \\
        \quad -- Separate scene blocks with a single blank line. \\
        \quad -- Each event is exactly one line, starting with ``- ''. \\
        \quad -- A DIRECTION line is ``- DIRECTION: '' followed by the action text. \\
        \quad -- A DIALOGUE line is ``- DIALOGUE \textbar{}\allowbreak{} '' followed by three fields separated by `` \textbar{}\allowbreak{} '': the CHARACTER cue, then the parenthetical (use ``-'' when there is none), then the spoken text. \\
        \quad -- Put the ``EVENTS:'' header line even if a scene has zero events (then list no event lines under it). \\
        \quad -- For a continuation scene, still set Continuation: yes and leave Heading/\allowbreak{}Location/\allowbreak{}Time blank. \\
        \\
        \textbf{WORKED EXAMPLE} (illustrative -- your real input is the attached image; this just shows the EXACT output shape). Suppose a page opens mid-dialogue (no slugline at the top -- the tail of a scene from the previous page), then a new scene begins. The correct output is EXACTLY: \\
        \\
        \texttt{\frenchspacing SCENE 1} \\
        \texttt{\frenchspacing Continuation: yes} \\
        \texttt{\frenchspacing Heading:} \\
        \texttt{\frenchspacing Location:} \\
        \texttt{\frenchspacing Time:} \\
        \texttt{\frenchspacing Characters: SARAH} \\
        \texttt{\frenchspacing EVENTS:} \\
        \texttt{\frenchspacing - DIALOGUE \textbar{}\allowbreak{} SARAH \textbar{}\allowbreak{} - \textbar{}\allowbreak{} I never said that.} \\
        \\
        \texttt{\frenchspacing SCENE 2} \\
        \texttt{\frenchspacing Continuation: no} \\
        \texttt{\frenchspacing Heading: INT. DINER - NIGHT} \\
        \texttt{\frenchspacing Location: DINER} \\
        \texttt{\frenchspacing Time: NIGHT} \\
        \texttt{\frenchspacing Characters: MARCUS, SARAH} \\
        \texttt{\frenchspacing EVENTS:} \\
        \texttt{\frenchspacing - DIRECTION: Marcus slides into the booth across from Sarah, rain dripping from his coat.} \\
        \texttt{\frenchspacing - DIALOGUE \textbar{}\allowbreak{} MARCUS \textbar{}\allowbreak{} quietly \textbar{}\allowbreak{} You came.} \\
        \texttt{\frenchspacing - DIALOGUE \textbar{}\allowbreak{} SARAH \textbar{}\allowbreak{} - \textbar{}\allowbreak{} I almost didn't.} \\
        \\
        \textbf{Note from the example:} the continuation scene leaves Heading/\allowbreak{}Location/\allowbreak{}Time blank; a parenthetical is written WITHOUT its parentheses (``quietly''); ``-'' marks a dialogue with no parenthetical; each event is one line. \\
        \bottomrule
    \end{tabular}
    }
    \caption{\textbf{Prompt for parsing a screenplay page (continued).}}
\end{table*}

\begin{table*}
    \centering
    \small
    \resizebox{.93\textwidth}{!}{
    \begin{tabular}{p{.9\textwidth}}
        \toprule
        \textbf{Instructions}: You are a screenplay analyst. Given the full text of a single scene from a movie screenplay (heading, action lines, and dialogue), write a concise 2-4 sentence summary capturing what happens in the scene: who is present, what they do or decide, and what changes or is revealed by the end of it. \\
        \midrule
        \textbf{Input Information:} \\
        \textit{Film}: \{script\_id\} \\
        Scene \{scene\_number\} -- \{heading\} \\
        \\
        \textit{Scene text}: \\
        \{scene\_text\} \\
        \\
        Write a 2-4 sentence summary of this scene. Output only the summary prose. \\
        \midrule
        \textbf{Rules:} \\
        \quad -- Base the summary ONLY on what the scene text shows. Do not invent backstory, motivations, or off-screen events not implied by the scene. \\
        \quad -- Refer to characters by the names used in the scene's character cues (e.g. ``JOHN'', ``MRS. SMITH''). Do not translate, anglicise, or shorten them. \\
        \quad -- Do not quote dialogue verbatim; paraphrase when you need to reference what was said. \\
        \quad -- If the scene is purely transitional or has no clear action (e.g. an establishing shot, a brief cutaway), say so in one sentence rather than padding. \\
        \quad -- Stay neutral and analytic. You see only this one scene, with no surrounding context -- summarise what the scene itself shows; do not speculate about its wider significance to the plot or how it connects to other scenes. \\
        \\
        \textbf{OUTPUT FORMAT} (STRICT): \\
        \quad -- Output ONLY the summary text itself: 2-4 sentences of plain prose. \\
        \quad -- No preamble, no label, no ``Summary:'' prefix, no markdown, no bullet points, no quotes around the whole thing. \\
        \bottomrule
    \end{tabular}
    }
    \caption{\textbf{Prompt for summarizing a scene.}}
    \label{tab:prompt-scene-summary}
\end{table*}

\begin{table*}
    \centering
    \small
    \resizebox{.93\textwidth}{!}{
    \begin{tabular}{p{.9\textwidth}}
        \toprule
        \textbf{Instructions}: You are a character coreference resolver for screenplays. \\
        \midrule
        \textbf{Input Information:} \\
        \textit{Film}: \{film\_title\} \\
        \\
        \textit{NAMES TO GROUP (every one of these must appear in exactly one group, spelled exactly as written here)}: \\
        \{names\} \\
        \\
        \textit{Character descriptions}: \\
        \{characters\_with\_descriptions\} \\
        \\
        \textit{Scene appearances per character}: \\
        \{characters\_with\_scenes\} \\
        \\
        Group all names that refer to the same character. Every name in the NAMES TO GROUP list must appear in exactly one group. Output one line per group in the format: \texttt{\textless{}CANONICAL\textgreater{} \textless{}- \textless{}alias1\textgreater{}, \textless{}alias2\textgreater{}, ...} \\
        \midrule
        You will receive a list of character names with descriptions and scene appearances. Many of these names refer to the SAME character under different spellings, cases, abbreviations, nicknames, or OCR artefacts. Your job is to group every name into clusters where each cluster represents exactly one unique character or entity. \\
        \\
        \textbf{OUTPUT REQUIREMENTS} \\
        1. Every name from the input must appear in exactly one group -- the user message includes a NAMES TO GROUP checklist that is the definitive list you must cover. Do not drop, duplicate, or invent any name. \\
        2. For the canonical name, pick the most complete and properly-cased form that actually appears in the input (prefer full name over first-name-only; prefer mixed case over all-caps when both exist). \\
        \\
        \textbf{EVIDENCE TO USE} \\
        Combine BOTH descriptions and scene appearances when deciding whether two names corefer: \\
        \quad -- Similar descriptions (same role, actions, relationships, age, profession) are strong evidence of identity. \\
        \quad -- Two names that appear in the same scene are very likely the same character -- screenplays rarely put two distinct characters with confusingly similar names in one scene. \\
        \quad -- Two similar-looking names that never share a scene may still be the same person appearing in different parts of the story; lean on descriptions in that case. \\
        \quad -- Two names with clearly DIFFERENT descriptions (different ages, professions, relationships) are distinct people even if their surface forms overlap. \\
        \bottomrule
    \end{tabular}
    }
    \caption{\textbf{Prompt for resolving coreferent character names.}}
    \label{tab:prompt-coref}
\end{table*}

\addtocounter{table}{-1}
\begin{table*}
    \centering
    \small
    \resizebox{.93\textwidth}{!}{
    \begin{tabular}{p{.9\textwidth}}
        \toprule
        \textbf{WHEN TO MERGE} \\
        \quad -- First name $\leftrightarrow$ full name: a first name that matches the first part of a longer name almost always refers to the same character (e.g. ``ALEX'' with ``ALEX MORGAN''). Merge unless descriptions clearly indicate two distinct people who share a first name. \\
        \quad -- Last-name-only $\leftrightarrow$ full name: merge when descriptions agree (e.g. ``MORGAN'' with ``ALEX MORGAN'', ``MRS. SMITH'' with ``ELLEN SMITH''). \\
        \quad -- Nicknames and diminutives: merge with the formal name when descriptions agree (e.g. ``Sam'' with ``Samuel'', ``Liz'' with ``Elizabeth'', ``Bobby'' with ``Robert''). \\
        \quad -- Case variants: always merge (e.g. ``JOHN'' /\allowbreak{} ``John'' /\allowbreak{} ``JoHn''). \\
        \quad -- Spacing /\allowbreak{} punctuation variants: merge (e.g. ``MARY-ANNE'' /\allowbreak{} ``MARY ANNE'' /\allowbreak{} ``MARYANNE'', ``MR SMITH'' /\allowbreak{} ``MR. SMITH''). \\
        \quad -- OCR artefacts: merge with the clean form (doubled letters like ``JJOOHHNN'' = ``JOHN'', broken hyphenation, stray characters, missing or extra spaces). \\
        \quad -- Honorific variants: merge (``DR. CHEN'' /\allowbreak{} ``CHEN'' /\allowbreak{} ``Dr Chen'') when descriptions agree. \\
        \\
        \textbf{WHEN NOT TO MERGE} \\
        \quad -- Different numbers or ordinals = different people. ``GUARD 1'' and ``GUARD 2'' are distinct; ``PATRON \#1'' and ``PATRON \#2'' are distinct; ``FIRST THIEF'' and ``SECOND THIEF'' are distinct. \\
        \quad -- Formatting variants of the SAME number DO merge: ``GUARD 1'' /\allowbreak{} ``GUARD \#1'' /\allowbreak{} ``Guard 1'' /\allowbreak{} ``GUARD ONE'' are one entity. \\
        \quad -- Two characters who share a first name but have clearly different descriptions (different ages, roles, relationships) are distinct -- keep them apart even when the surface forms overlap. \\
        \quad -- Generic group entities (``BAND'', ``CROWD'', ``SOLDIERS'', ``VAMPIRES'') that do not map to a specific named individual stay as their own group; do not fold them into a named character. \\
        \quad -- Role labels without a number (``WAITER'', ``DOCTOR'', ``REPORTER'') are usually one anonymous role and may merge with other unnumbered instances of the same role only when descriptions and scene context clearly point to the same person; otherwise keep separate. \\
        \\
        \textbf{OUTPUT FORMAT} (STRICT -- a downstream parser depends on it; output ONLY this, no preamble, no markdown, no commentary): \\
        \quad -- Output one line per character group, in this EXACT shape: \\
        \\
        \texttt{\frenchspacing \textless{}CANONICAL NAME\textgreater{} \textless{}- \textless{}alias1\textgreater{}, \textless{}alias2\textgreater{}, \textless{}alias3\textgreater{}} \\
        \\
        \quad -- Left of the `` \textless{}- '' is the canonical name you chose. Right of it is the comma-separated list of EVERY input name in that group. \\
        \quad -- The canonical name MUST also appear in its own alias list on the right. \\
        \quad -- A character that has no other forms is its own one-item group: write its canonical name on the left and the same single name on the right (e.g. ``JOHN \textless{}- JOHN''). \\
        \quad -- Use one line per group. Do not number the lines. Do not add blank lines between groups. \\
        \quad -- Use the names EXACTLY as they appear in the input (preserve their original casing and spelling) on the right-hand alias list. \\
        \bottomrule
    \end{tabular}
    }
    \caption{\textbf{Prompt for resolving coreferent character names (continued).}}
\end{table*}

\begin{table*}
    \centering
    \small
    \resizebox{.93\textwidth}{!}{
    \begin{tabular}{p{.9\textwidth}}
        \toprule
        \textbf{Instructions}: You read a single scene from a parsed screenplay and extract its canonical events for downstream character-perspective analysis. \\
        \midrule
        \textbf{Input Information:} \\
        SCENE \{scene\_number\} \\
        \textit{Heading}: \{heading\} \\
        \textit{Location}: \{location\} \\
        \textit{Listed characters}: \{characters\} \\
        \\
        \textit{SCENE SUMMARY (anchor -- one line, what this scene is about)}: \\
        \{summary\} \\
        \\
        \textit{SCENE TEXT}: \\
        \{scene\_text\} \\
        \\
        Extract the canonical events for this scene using the EVENT-block format. If there are no canonical events, output exactly: NO EVENTS \\
        \midrule
        \textbf{What counts as a canonical event:} \\
        \quad -- An ATOMIC world-event: one action, one event. A single direction block often describes several atomic events (e.g. ``Angier breaks in and reads the journal'' = two events: a break-in, then a read of the journal). \\
        \quad -- Plot-consequential dialogue beats ARE canonical events: a confession given, a command issued, a public lie told on the record, an oath sworn, an agreement made. Casual conversation is NOT. \\
        \quad -- A character speaking only in V.O. (voice-over) or O.S. (off-screen) is narrating, not present. Their V.O. lines are not canonical events. \\
        \quad -- A pure transitional moment (title card, montage of stock images with no story action) has NO canonical events. In that case output exactly: NO EVENTS \\
        \\
        \textbf{Atomization rules:} \\
        \quad -- Group consecutive direction lines describing ONE ongoing action into a single event. \\
        \quad -- Split a direction line into multiple events if it describes distinct actions (``A enters and shoots B'' = two events). \\
        \quad -- Output events in chronological order within the scene. \\
        \bottomrule
    \end{tabular}
    }
    \caption{\textbf{Prompt for extracting scene events.}}
    \label{tab:prompt-events}
\end{table*}

\addtocounter{table}{-1}
\begin{table*}
    \centering
    \small
    \resizebox{.93\textwidth}{!}{
    \begin{tabular}{p{.9\textwidth}}
        \toprule
        \textbf{Participants vs witnesses} (these are independent dimensions): \\
        \quad -- PARTICIPANTS: characters who act in the event or are acted upon by it. Include unconscious victims as participants if the action is done to them. \\
        \quad -- WITNESSES: characters who DIRECTLY PERCEIVE the event in real time. Default = the scene's listed characters, MINUS: \\
        \qquad -- any character whose only presence in the scene is via V.O. or O.S. parenthetical dialogue (they are narrating, not in the room). \\
        \qquad -- any character the direction marks as unconscious, asleep, blindfolded, looking away, in another room, or otherwise unable to perceive THIS specific event. \\
        \quad -- A character can be a participant without being a witness (an unconscious victim) or a witness without being a participant (a bystander). \\
        \\
        \textbf{Character IDs:} \\
        \quad -- Use the EXACT names from the scene's ``Listed characters'' line. Do not invent or rename anyone. Do not import characters who are not in that list. \\
        \quad -- Never list a V.O.-only narrator as a participant or witness of any event in this scene. \\
        \\
        \textbf{Output format} (STRICT -- a downstream parser depends on it): \\
        \quad -- For each canonical event, write one block in this exact shape: \\
        \\
        \texttt{\frenchspacing EVENT N} \\
        \texttt{\frenchspacing Description: \textless{}one sentence, neutral third-person; names from the listed roster are fine\textgreater{}} \\
        \texttt{\frenchspacing Participants: \textless{}comma-separated names, or "(none)"\textgreater{}} \\
        \texttt{\frenchspacing Witnesses: \textless{}comma-separated names, or "(none)"\textgreater{}} \\
        \\
        \quad -- Number events sequentially starting at 1. \\
        \quad -- Separate event blocks with a single blank line. \\
        \quad -- Each field on its own line; do not break ``Description:'' across multiple lines. \\
        \quad -- Output ONLY the EVENT blocks (or ``NO EVENTS''). No preamble, no closing remarks, no markdown, no bullets, no quotes around values. \\
        \bottomrule
    \end{tabular}
    }
    \caption{\textbf{Prompt for extracting scene events (continued).}}
\end{table*}

\begin{table*}
    \centering
    \small
    \resizebox{.93\textwidth}{!}{
    \begin{tabular}{p{.9\textwidth}}
        \toprule
        \textbf{Instructions}: You model the first-order belief state of ONE character in ONE scene of a film. \\
        \midrule
        \textbf{Input Information:} \\
        \textit{Film}: \{film\_title\} \\
        \\
        \textit{PREVIOUS SCENE}: \\
        \{prev\_scene\_context\} \\
        \\
        \textit{CURRENT SCENE}: \\
        \{current\_scene\_block\} \\
        \\
        \textit{CANONICAL EVENTS (cite event\_ids from here)}: \\
        \{canonical\_events\} \\
        \\
        \textit{SCREENPLAY TEXT}: \\
        \{scene\_screenplay\} \\
        \\
        \textit{NEXT SCENE}: \\
        \{next\_scene\_context\} \\
        \\
        \textit{TARGET CHARACTER}: \{character\} \\
        \\
        \{belief\_state\} \\
        \\
        Now decide \{character\}'s PRESENCE and WITNESSED event\_ids for this scene. Output ONLY the two lines specified in your instructions. \\
        \midrule
        You are given: the scene's screenplay text and its canonical events (with stable event\_ids); the target character's BELIEF STATE coming into the scene (their current beliefs, active false beliefs, and emotional state) plus the last 1-2 scenes in full; and the next-scene heading for orientation. \\
        \\
        \textbf{Rules:} \\
        \quad -- Only track what THIS character could know FROM THIS SCENE. Exclude events they did not witness and were not told about. \\
        \quad -- A V.O. (voice-over) parenthetical -- narrator, phone voice, inner monologue, a letter read aloud -- is NARRATING, not physically present. An O.S./\allowbreak{}O.C. (off-screen /\allowbreak{} off-camera) parenthetical is the opposite: the character IS in the scene's space, just out of frame (e.g. calling from the next room), and perceives the scene. \\
        \quad -- Cite only event\_ids that appear in this scene's canonical events, and scene\_ids or beliefs that appear in the belief state. Never invent IDs or character names. \\
        \\
        \textbf{YOUR JOB} (this prompt): decide the character's PRESENCE this scene and list the canonical events they DIRECTLY witnessed. \\
        \bottomrule
    \end{tabular}
    }
    \caption{\textbf{Prompt for first-order beliefs, call~1: presence and witnessed events.}}
    \label{tab:prompt-fo-presence}
\end{table*}

\addtocounter{table}{-1}
\begin{table*}
    \centering
    \small
    \resizebox{.93\textwidth}{!}{
    \begin{tabular}{p{.9\textwidth}}
        \toprule
        \textbf{PRESENCE} -- pick exactly one (a ladder of decreasing access; choose the highest rung that holds): \\
        \quad -- on\_screen -- the scene shows them speaking or acting, or places them physically in its space. A character speaking O.S./\allowbreak{}O.C. (off-screen /\allowbreak{} off-camera: in the room but out of frame, e.g. calling from the next room) counts -- they perceive the scene. A V.O. (voice-over: narrator, phone voice, inner monologue, a letter read aloud) does NOT -- that is narration, not presence. \\
        \quad -- offscreen\_aware -- not in the scene, but a concrete channel makes awareness likely: they were on\_screen last scene at the same or an adjacent location and have not left; the previous scene sent them toward this situation; they are stationed within sight/\allowbreak{}earshot (this one can yield real witnessed events); or the next scene shows them already reacting with no one having told them. \\
        \quad -- offscreen\_unaware -- elsewhere and unaware, BUT the scene bears on them: it changes their world or leaves a belief of theirs now false (dramatic irony). They learn nothing here, yet what they still wrongly believe depends on it. \\
        \quad -- absent\_unmentioned -- a different storyline with no bearing on them at all. \\
        \\
        \textbf{Tie-breakers} (walk the ladder top-down, stop at the first rung you can justify): \\
        \quad -- on\_screen vs offscreen\_aware: if you cannot place them in the scene's space, drop to offscreen\_aware. \\
        \quad -- offscreen\_aware vs offscreen\_unaware: pick aware only if you can NAME the concrete channel; otherwise unaware. \\
        \quad -- offscreen\_unaware vs absent\_unmentioned: pick unaware if the scene touches their situation or beliefs; absent only if it is truly unrelated. \\
        \\
        \textbf{WITNESSED:} list only event\_ids the character DIRECTLY perceives (sees/\allowbreak{}hears/\allowbreak{}feels firsthand). Second-hand information is NOT witnessed (it is handled as \texttt{told} in the next step). Being on\_screen does not force entries here -- someone present but asleep, unconscious, or away from the action witnesses nothing. Normally empty when offscreen or absent. \\
        \\
        Output ONLY these two lines, nothing else: \\
        \\
        \texttt{\frenchspacing Presence: \textless{}on\_screen\textbar{}\allowbreak{}offscreen\_aware\textbar{}\allowbreak{}offscreen\_unaware\textbar{}\allowbreak{}absent\_unmentioned\textgreater{}} \\
        \texttt{\frenchspacing Witnessed event IDs: \textless{}comma-separated event\_ids, or "(none)"\textgreater{}} \\
        \bottomrule
    \end{tabular}
    }
    \caption{\textbf{Prompt for first-order beliefs, call~1: presence and witnessed events (continued).}}
\end{table*}

\begin{table*}
    \centering
    \small
    \resizebox{.93\textwidth}{!}{
    \begin{tabular}{p{.9\textwidth}}
        \toprule
        \textbf{Instructions}: You model the first-order belief state of ONE character in ONE scene of a film. \\
        \midrule
        \textbf{Input Information:} \\
        \textit{Film}: \{film\_title\} \\
        \\
        \textit{PREVIOUS SCENE}: \\
        \{prev\_scene\_context\} \\
        \\
        \textit{CURRENT SCENE}: \\
        \{current\_scene\_block\} \\
        \\
        \textit{CANONICAL EVENTS (cite event\_ids from here)}: \\
        \{canonical\_events\} \\
        \\
        \textit{SCREENPLAY TEXT}: \\
        \{scene\_screenplay\} \\
        \\
        \textit{NEXT SCENE}: \\
        \{next\_scene\_context\} \\
        \\
        \textit{TARGET CHARACTER}: \{character\} \\
        \\
        \{belief\_state\} \\
        \\
        \textit{CALL 1 RESULT (presence + witnessed\_event\_ids for this scene -- treat as ground truth)}: \\
        \{call1\_text\} \\
        \\
        Now produce TOLD, INFERRED, FALSE BELIEFS HELD, and REVISIONS for \{character\} in this scene, using the exact four-section format specified in your instructions. \\
        \midrule
        You are given: the scene's screenplay text and its canonical events (with stable event\_ids); the target character's BELIEF STATE coming into the scene (their current beliefs, active false beliefs, and emotional state) plus the last 1-2 scenes in full; and the next-scene heading for orientation. \\
        \\
        \textbf{Rules:} \\
        \quad -- Only track what THIS character could know FROM THIS SCENE. Exclude events they did not witness and were not told about. \\
        \quad -- A V.O. (voice-over) parenthetical -- narrator, phone voice, inner monologue, a letter read aloud -- is NARRATING, not physically present. An O.S./\allowbreak{}O.C. (off-screen /\allowbreak{} off-camera) parenthetical is the opposite: the character IS in the scene's space, just out of frame (e.g. calling from the next room), and perceives the scene. \\
        \quad -- Cite only event\_ids that appear in this scene's canonical events, and scene\_ids or beliefs that appear in the belief state. Never invent IDs or character names. \\
        \\
        \textbf{YOUR JOB} (this prompt): given Call 1's presence + witnessed result (treat it as ground truth), extract what the character was TOLD this scene, what they INFERRED, the FALSE BELIEFS they currently hold, and any REVISIONS to prior beliefs. \\
        \bottomrule
    \end{tabular}
    }
    \caption{\textbf{Prompt for first-order beliefs, call~2: told, inferred, false, and revised beliefs.}}
    \label{tab:prompt-fo-beliefs}
\end{table*}

\addtocounter{table}{-1}
\begin{table*}
    \centering
    \small
    \resizebox{.93\textwidth}{!}{
    \begin{tabular}{p{.9\textwidth}}
        \toprule
        \textbf{TOLD} -- propositions communicated to the character via dialogue this scene. Extract the proposition, not the wording (``Smoke says the deal is off'' $\rightarrow$ ``the deal is off''). Mark believed yes/\allowbreak{}no. A character who is not present cannot be told anything here. \\
        \textbf{INFERRED} -- NEW claims the character draws from witnessed + told + their current beliefs. Each cites a basis: event\_ids from this scene and/\allowbreak{}or scene\_ids. Do NOT re-emit beliefs already listed in the belief state. \\
        \textbf{FALSE BELIEFS HELD} -- beliefs the character holds that contradict the canonical events. Carry forward the belief state's active false beliefs unless this scene corrects them; add any new ones introduced by deception or mistaken inference. Write each false belief in FULL TEXT here (do not use a [Bk]/\allowbreak{}[Fk] tag in this section). \\
        \textbf{REVISIONS} -- when this scene corrects a prior belief: in \texttt{Prior}, cite that belief's [Bk] or [Fk] tag from the belief state (e.g. [B3]); write it out verbatim only if it has no tag there. Then give what it is revised to and the triggering event\_ids. \\
        \\
        Output ONLY these four sections, in this order. Each section's body is a list of items or the literal ``(none)''. Separate items with a blank line. \\
        \\
        \texttt{\frenchspacing TOLD} \\
        \texttt{\frenchspacing - Proposition: \textless{}claim\textgreater{}} \\
        \texttt{\frenchspacing ~~Source: \textless{}CHAR\textgreater{}} \\
        \texttt{\frenchspacing ~~Believed: \textless{}yes\textbar{}\allowbreak{}no\textgreater{}} \\
        \\
        \texttt{\frenchspacing INFERRED} \\
        \texttt{\frenchspacing - Claim: \textless{}claim\textgreater{}} \\
        \texttt{\frenchspacing ~~Basis: \textless{}comma-separated event\_ids and/\allowbreak{}or scene\_ids\textgreater{}} \\
        \texttt{\frenchspacing ~~Confidence: \textless{}high\textbar{}\allowbreak{}medium\textbar{}\allowbreak{}low\textgreater{}} \\
        \\
        \texttt{\frenchspacing FALSE BELIEFS HELD} \\
        \texttt{\frenchspacing - \textless{}belief\textgreater{}} \\
        \\
        \texttt{\frenchspacing REVISIONS} \\
        \texttt{\frenchspacing - Prior: \textless{}[Bk]/\allowbreak{}[Fk] tag from the belief state, e.g. [B3] -{}- or verbatim text if untagged\textgreater{}} \\
        \texttt{\frenchspacing ~~Revised to: \textless{}new belief\textgreater{}} \\
        \texttt{\frenchspacing ~~Trigger: \textless{}comma-separated event\_ids\textgreater{}} \\
        \\
        No preamble, no closing remarks, no markdown headers. \\
        \bottomrule
    \end{tabular}
    }
    \caption{\textbf{Prompt for first-order beliefs, call~2: told, inferred, false, and revised beliefs (continued).}}
\end{table*}

\begin{table*}
    \centering
    \small
    \resizebox{.93\textwidth}{!}{
    \begin{tabular}{p{.9\textwidth}}
        \toprule
        \textbf{Instructions}: You model character c's META-BELIEFS -- what c believes about OTHER characters' mental states -- for ONE scene of a film. Second-order only (c about d): what c thinks d believes. Never third order. \\
        \midrule
        \textbf{Input Information:} \\
        \textit{Film}: \{film\_title\} \\
        \textit{SCENE ID}: \{scene\_id\} \\
        \\
        \textit{PREVIOUS SCENE}: \\
        \{prev\_scene\_context\} \\
        \\
        \textit{CURRENT SCENE}: \\
        \{current\_scene\_block\} \\
        \\
        \textit{CANONICAL EVENTS (prefer these event\_ids for Basis; use this scene's id s\_... for inference/\allowbreak{}absence beliefs with no discrete event)}: \\
        \{canonical\_events\} \\
        \\
        \textit{SCREENPLAY TEXT}: \\
        \{scene\_screenplay\} \\
        \\
        \textit{NEXT SCENE}: \\
        \{next\_scene\_context\} \\
        \\
        \textit{MODELING CHARACTER (c)}: \{character\} \\
        \textit{TARGET GROUP THIS CALL (model ONLY these \{character\} $\rightarrow$ d)}: \{target\_set\} \\
        \\
        \textit{C'S FIRST-ORDER UPDATE FOR THIS SCENE}: \\
        \{c\_first\_order\_update\} \\
        \\
        \{meta\_state\} \\
        \\
        Now emit \{character\}'s NEW or CHANGED meta-beliefs about the targets in the group above (any of the three types; the Claim is the bare proposition d believes). Skip beliefs already held unchanged. OMIT targets with nothing new -- do not enumerate them. If no target has anything new, output the single word NONE. Use the exact per-target block format from your instructions. \\
        \midrule
        You are given: the scene's screenplay text and canonical events (with stable event\_ids); c's first-order update for THIS scene (what c witnessed, was told, inferred); c's meta-beliefs held so far plus the last 1-2 scenes in full; and a SMALL GROUP of target characters -- the only characters you may model in THIS call. \\
        \bottomrule
    \end{tabular}
    }
    \caption{\textbf{Prompt for meta-beliefs.}}
    \label{tab:prompt-meta}
\end{table*}

\addtocounter{table}{-1}
\begin{table*}
    \centering
    \small
    \resizebox{.93\textwidth}{!}{
    \begin{tabular}{p{.9\textwidth}}
        \toprule
        For each target d in the group, emit c's NEW or CHANGED meta-beliefs about d's mind -- of any of the three types below. Model ONLY the targets in this call's group; never emit about a character outside it. \\
        \\
        \textbf{THE THREE TYPES} (tag every item with exactly one; they carry equal weight -- do not default to any single type): \\
        \quad -- first\_order\_about\_d -- c thinks d believes a proposition about the WORLD (not about c, not about a specific third character). This is the MOST COMMON type: what d knows or believes about events, facts, and situations. E.g. ``the deal is off''; or via d\_absence, c thinks d does NOT know something that happened. \\
        \quad -- about\_self -- c thinks d believes a proposition ABOUT c personally. E.g. ``c is the new pastor''. (Relevant to deception: if c lied to d, or someone lied about c in d's hearing, c gains an about\_self belief that d holds a possibly-false belief about c.) \\
        \quad -- about\_third\_party -- c thinks d believes a proposition about a THIRD named character e (e is different from both c and d). Name e in the Claim; e must already appear in the canonical events or prior history. \\
        \\
        \textbf{CLAIM PHRASING} (strict): the Claim is the BARE PROPOSITION that d believes -- nothing more. The About field already names the believer, so NEVER write ``d believes/\allowbreak{}knows/\allowbreak{}thinks ...'' inside the Claim. Write the proposition itself (``the deal is off'', ``c is the new pastor'', ``e stole the money''). Encode how strongly d holds it via Source and Confidence, not with a verb in the claim. If the proposition is ITSELF about someone's belief (a belief about a belief), DROP it -- this extractor is second-order only. \\
        \\
        \textbf{VALID SOURCES} (pick exactly one per item): \\
        \quad -- shared\_witnessing -- c and d both witnessed event e; c gains ``d knows e''. \\
        \quad -- observed\_d\_told -- c saw someone tell d a proposition. \\
        \quad -- observed\_d\_behavior -- d acted in a way implying a belief; c infers it (lower confidence). \\
        \quad -- c\_told\_d -- c directly communicated something to d. \\
        \quad -- d\_absence -- d was not present and has no plausible channel; c believes d does NOT know (reserve for gaps that matter, not routine). \\
        \quad -- shared\_prior\_context -- mutual background knowledge (use sparingly). \\
        \bottomrule
    \end{tabular}
    }
    \caption{\textbf{Prompt for meta-beliefs (continued).}}
\end{table*}

\addtocounter{table}{-1}
\begin{table*}
    \centering
    \small
    \resizebox{.93\textwidth}{!}{
    \begin{tabular}{p{.9\textwidth}}
        \toprule
        \textbf{GROUNDING} (every item MUST cite a Basis): \\
        \quad -- PREFER a canonical event\_id from THIS scene (e.g. e\_017\_3) -- the specific event that supports the claim. \\
        \quad -- When the belief comes from dialogue or stage direction with no discrete canonical event -- common for observed\_d\_behavior and shared\_prior\_context -- cite THIS scene's id (e.g. s\_017). \\
        \quad -- For d\_absence (c believes d does NOT know something), cite the PRIOR scene\_id where the missed event occurred. \\
        No valid basis, no item -- drop it rather than invent an id. Never fabricate event\_ids, scene\_ids, or character names. \\
        \\
        \textbf{NOVELTY:} emit a meta-belief when it is NEW this scene, when this scene REVISES one c already holds (a CHANGED belief -- cite the prior scene\_id plus this scene's trigger), or when this scene gives materially stronger evidence for a previously shaky belief. Do NOT re-state beliefs already in ``meta-beliefs held so far'' that are unchanged. \\
        \\
        \textbf{DO NOT EMIT} (these are not meta-beliefs): \\
        \quad -- trivially-true standing facts, static traits, or roles (``d is a bartender'', ``d is c's employer''). \\
        \quad -- public world-facts every observer already shares, with no divergent knowledge (``the band is playing''). \\
        \quad -- bare narrative summaries or predictions no named d actually holds (``the meeting ended without incident''). \\
        \quad -- perceptions or imperatives (``d looks at c'', ``d should leave''). \\
        \quad -- the SAME observation emitted twice under two types -- pick the single best type. \\
        \\
        A meta-belief may be wrong: c can believe d knows something d does not. That is valid -- capture what c believes, not the ground truth. \\
        \\
        \textbf{OUTPUT FORMAT} (STRICT -- a downstream parser depends on it). Emit one block per meta-belief. OMIT targets you have nothing new for -- do not enumerate them. If NO target in the group has anything new or changed this scene, output the single word NONE and nothing else. \\
        \\
        \texttt{\frenchspacing - About: \textless{}D -{}- a target in this group\textgreater{}} \\
        \texttt{\frenchspacing ~~Claim: \textless{}the bare proposition d believes\textgreater{}} \\
        \texttt{\frenchspacing ~~Type: \textless{}first\_order\_about\_d\textbar{}\allowbreak{}about\_self\textbar{}\allowbreak{}about\_third\_party\textgreater{}} \\
        \texttt{\frenchspacing ~~Source: \textless{}shared\_witnessing\textbar{}\allowbreak{}observed\_d\_told\textbar{}\allowbreak{}observed\_d\_behavior\textbar{}\allowbreak{}c\_told\_d\textbar{}\allowbreak{}d\_absence\textbar{}\allowbreak{}shared\_prior\_context\textgreater{}} \\
        \texttt{\frenchspacing ~~Basis: \textless{}comma-separated event\_ids and/\allowbreak{}or scene\_ids\textgreater{}} \\
        \texttt{\frenchspacing ~~Confidence: \textless{}high\textbar{}\allowbreak{}medium\textbar{}\allowbreak{}low\textgreater{}} \\
        \bottomrule
    \end{tabular}
    }
    \caption{\textbf{Prompt for meta-beliefs (continued).}}
\end{table*}

\addtocounter{table}{-1}
\begin{table*}
    \centering
    \small
    \resizebox{.93\textwidth}{!}{
    \begin{tabular}{p{.9\textwidth}}
        \toprule
        \textbf{WORKED EXAMPLE} (neutral, not from this film). Scene s\_042: c=Rosa and Dev both watch the foreman announce the factory will close Friday; Rosa then quietly tells Dev she has already lined up new buyers. Priya, a partner, is not present. Group: Dev, Priya. Output: \\
        \\
        \texttt{\frenchspacing - About: Dev} \\
        \texttt{\frenchspacing ~~Claim: the factory will close on Friday} \\
        \texttt{\frenchspacing ~~Type: first\_order\_about\_d} \\
        \texttt{\frenchspacing ~~Source: shared\_witnessing} \\
        \texttt{\frenchspacing ~~Basis: e\_042\_3} \\
        \texttt{\frenchspacing ~~Confidence: high} \\
        \texttt{\frenchspacing - About: Dev} \\
        \texttt{\frenchspacing ~~Claim: Rosa has already lined up new buyers} \\
        \texttt{\frenchspacing ~~Type: about\_self} \\
        \texttt{\frenchspacing ~~Source: c\_told\_d} \\
        \texttt{\frenchspacing ~~Basis: e\_042\_7} \\
        \texttt{\frenchspacing ~~Confidence: high} \\
        \texttt{\frenchspacing - About: Priya} \\
        \texttt{\frenchspacing ~~Claim: the factory will close on Friday} \\
        \texttt{\frenchspacing ~~Type: first\_order\_about\_d} \\
        \texttt{\frenchspacing ~~Source: d\_absence} \\
        \texttt{\frenchspacing ~~Basis: s\_042} \\
        \texttt{\frenchspacing ~~Confidence: medium} \\
        \\
        (The plain world-belief is the common case; about\_self and about\_third\_party are the minority. The last item shows d\_absence for an absent partner, grounded on the scene id -- use it only when the knowledge gap matters.) \\
        \\
        Output ONLY the per-target blocks (or NONE). No preamble, no closing remarks, no markdown headers. \\
        \bottomrule
    \end{tabular}
    }
    \caption{\textbf{Prompt for meta-beliefs (continued).}}
\end{table*}

\begin{table}[h]
\centering\footnotesize
\setlength{\tabcolsep}{4pt}\renewcommand{\arraystretch}{1.15}
\begin{tabular}{@{}l p{0.76\textwidth}@{}}
\toprule
Type & Specification injected into the question-writing prompt \\
\midrule
Q1~Knowledge split & Cohort knowledge split --- at one moment several people are in different epistemic states. What does each take to be the case? \\
Q2~Realization & Realization moment --- X's understanding changes at a locatable point in the film. The answer must say when; the question must not say what. \\
Q3~False belief & In-scene false belief at a threshold --- right before an act, X holds a belief the film has shown to be wrong. Ask what X takes to be the case. Do NOT signal that X is mistaken; noticing the gap is the task. \\
Q4~Knowledge source & Source attribution --- a fact reaches people by different routes (c\_told\_d, observed\_d\_behavior, shared\_witnessing, d\_absence), and one scene often carries more than one route. Ask how someone comes to know it, open. Never offer the routes as alternatives to pick between. \\
Q5~First-order belief & First-order ToM --- X holds a model of what Y believes. Ask what X takes Y to believe. \\
Q6~Second-order belief & Higher-order ToM --- X holds a model of what Y believes about Z. \\
Q7~Accepting a claim & Deception detection --- X is told things; believed\_by\_c marks which X accepts and which X rejects, and BOTH happen across this dataset. Ask what X makes of the claim. Never phrase it as if either outcome were a given, and do not ask the film to adjudicate the claim's truth in the question --- the answer settles that. \\
Q8~Order of discovery & Temporal ordering of learning --- X's discoveries, in order. \\
N1~Epistemic arc & Epistemic-arc summary --- what X comes to know across the film, in order. \\
N2~Trust evolution & Relationship/trust evolution --- how X's model of Y changes across the film. \\
N3~Causal chain & Causal chain --- what chain of events leads X to believe Z. \\
\bottomrule
\end{tabular}
\caption{\textbf{The per-type specification.} One line per question type, inserted verbatim into the question-writing prompt after the system prompt. Each names the epistemic structure the question must require and stops; it does not prescribe the question's wording. The field names it mentions (\texttt{c\_told\_d}, \texttt{believed\_by\_c}) are the belief record's own; the question writer is told to keep them out of the question.}
\label{tab:type-specs}
\end{table}

\begin{table*}
    \centering
    \small
    \resizebox{.93\textwidth}{!}{
    \begin{tabular}{p{.9\textwidth}}
        \toprule
        \textbf{Instructions}: You write questions for a film-QA dataset. You see structured belief annotations; the test-taker sees ONLY the film. Your question is good if a viewer who watched carefully answers it and a viewer who watched carelessly gets it wrong. It is bad if it can be answered from the question itself, from the question's shape, or from genre convention. \\
        \midrule
        \textbf{Input Information:} \\
        QUESTION\_TYPE: \{q\_type\} -- \{type\_specification\} \\
        \\
        \textit{ANSWER CONTRACT (applies to every question of every type -- do not restate it in the question text)}: \\
        \quad -- Every claim in answer\_outline names the on-screen basis it rests on, in source\_path. \\
        \quad -- Where a type turns on HOW something was learned (told /\allowbreak{} observed /\allowbreak{} witnessed together /\allowbreak{} inferred from an absence), the answer says which. The question does not. \\
        \quad -- Where a type turns on WHEN, the answer locates the moment. The question does not. \\
        \quad -- Where the holder is wrong, the answer says so. The question does not. \\
        \\
        \textit{EXEMPLARS (a different film; imitate the shape, not the content)}: \\
        \{two\_example\_triples\} \\
        \\
        \textit{CONTEXT PACK}: \\
        \{context\_set\_json\} \\
        \midrule
        \textbf{READING THE PACK.} Four parts, and the third is the one to read first. \\
        \quad -- \texttt{basis} -- the belief records the question is built from. \texttt{claim} is what is believed; \texttt{source} is how it was acquired; \texttt{basis} lists the canonical events it rests on. \\
        \quad -- \texttt{scenes} -- pass-1 canonical events with \texttt{participants} and \texttt{witnesses}. This is the ground truth for who could possibly have known something. If your question implies someone knew a thing, they must be in one of these lists. \\
        \quad -- \texttt{basis[].target\_state} -- WHAT THE OTHER PERSON ACTUALLY BELIEVES, from a separate extraction pass, at this same scene. A record says ``X believes Y believes P''; \texttt{target\_state} says what Y believes in their own right. \textbf{Compare them.} Where X's model of Y diverges from Y's own state -- especially where \texttt{target\_state} carries \texttt{false\_beliefs\_held} -- that gap is the question. Where they agree, X is simply right about Y and there is usually nothing to ask. \\
        \quad -- \texttt{name\_budget} /\allowbreak{} \texttt{named\_later} /\allowbreak{} \texttt{forbidden\_names} /\allowbreak{} \texttt{notes} -- see NAMING below. \\
        \bottomrule
    \end{tabular}
    }
    \caption{\textbf{Prompt for writing the question (Stage~1).}}
    \label{tab:prompt-question}
\end{table*}

\addtocounter{table}{-1}
\begin{table*}
    \centering
    \small
    \resizebox{.93\textwidth}{!}{
    \begin{tabular}{p{.9\textwidth}}
        \toprule
        \textbf{REFUSE A THIN PACK.} Before writing, ask: is there a specific wrong answer a careful person might give here? If not -- the belief is uncontested, X's model of Y matches Y exactly, the belief rests on a single unremarkable action, or the holder is someone the film never picks out -- set verdict=``thin'', say why in thin\_reason, leave question empty, and stop. A refused pack costs one call. A thin question costs a rubric, a grading pass, and a benchmark number that means nothing. \\
        \\
        \textbf{THE ANCHOR} -- point at ONE moment, do not narrate it. \\
        \quad -- At most 12 words before your first question word, or 18 if you need a disambiguator (below). \\
        \quad -- Name a PLACE, an OCCASION or an OBJECT (``in the utility hallway'', ``on the publishers' conference call'', ``at the beach overlook''). Never report what anyone does, decides, learns, reveals, wins, accepts or refuses there. \\
        \quad -- Anchor diegetically: never a scene number, event id, or annotation vocabulary. \\
        \quad -- The test-taker's job includes FINDING the moment. A synopsis does that job for them, and smuggles the answer's setup in as premise. \\
        \quad -- \textbf{CHECK \texttt{anchor\_location\_rivals} FIRST.} It counts other scenes in this film at the same location with this holder present. You cannot work this out yourself -- you see three scenes of about a hundred and twenty. If it is 0, the place alone is enough. If it is 1 or more, naming the place is NOT enough and your question has several possible answers: add ONE distinguishing detail that is not an outcome -- a time (``months after the injury''), another person present (``with Art''), or an object (``over the fixed poker table''). Those locate the moment without saying what anyone believes, which is the whole point. Never disambiguate with what happens there. \\
        \quad -- The location string can also disagree with what the film shows (an annotation labelled a scene ``CORNER SUITE 2'' whose own summary opens ``Charlie enters Sterling's office''). Prefer what the SUMMARY shows a viewer over the location label. \\
        \\
        \textbf{TOPIC, NEVER VALUE.} Name what the belief is ABOUT. Never state what it says, whether it is true, or that it is contested. \\
        \quad -- Yes: ``what does Bosco make of Monk's attitude to the new title'' \\
        \quad -- No: ``...and what does Monk's reaction show is actually true'' -- announces the gap, which IS the question. \\
        \quad -- No: ``when does Monk first realize Cliff will not contribute'' -- the content of the realization is the answer; only its timing is left. \\
        \quad -- No: ``who is still reading it as just a routine drill'' -- hands over the mistake. \\
        \quad -- Test yourself: strike every word of your question that the gold answer also needs. If the gold is now mostly written, rewrite the question. \\
        \bottomrule
    \end{tabular}
    }
    \caption{\textbf{Prompt for writing the question (Stage~1) (continued).}}
\end{table*}

\addtocounter{table}{-1}
\begin{table*}
    \centering
    \small
    \resizebox{.93\textwidth}{!}{
    \begin{tabular}{p{.9\textwidth}}
        \toprule
        \textbf{ONE OPEN INTERROGATIVE.} A question word already forbids a yes/\allowbreak{}no answer; you do not need to defend against it. \\
        \quad -- No ``A, or B?'' alternatives. They are coin flips with a prior: 57\% of this dataset's source-attribution answers read ``inferred from behaviour'', so a model that always guesses that scores without watching anything. \\
        \quad -- No ``, and what supports that reading?'' or ``...and what gives it away?'' tails. Every answer must cite its on-screen basis -- that is a standing requirement of the answer, not something to announce in each question. \\
        \\
        \textbf{VARY THE FRAME.} If every question of a type opens the same way, the type is the answer key. Take the exemplars' level of specificity, not their sentence shape. \\
        \\
        \textbf{NAMING.} Freely use name\_budget. You may use named\_later -- the viewer watches the whole film -- unless the name is itself the reveal. NEVER use forbidden\_names: the film never says them aloud, so no viewer could connect one to a person. The scene text names people the film does not; that is the annotation talking. Follow the lists, not the prose. \\
        \\
        \textbf{ANSWER OUTLINE:} 2 to 4 claims, no more. Each is one assertion about who believes what, written as a viewer would see it, never as the annotation says it. Do not write ``the belief annotation marks X''; write what happens, and put every annotation reference (event ids, scene ids, basis fields) in source\_path. \\
        \\
        \textbf{DISTRACTOR:} required and concrete. Write the specific wrong answer you expect -- the inversion, the wrong holder, the surface event mistaken for the belief. Not ``a viewer might be confused''; a sentence someone would actually write. \\
        \\
        Return ONLY the JSON object. \\
        \bottomrule
    \end{tabular}
    }
    \caption{\textbf{Prompt for writing the question (Stage~1) (continued).}}
\end{table*}

\begin{table*}
    \centering
    \small
    \resizebox{.93\textwidth}{!}{
    \begin{tabular}{p{.9\textwidth}}
        \toprule
        \textbf{Instructions}: You write the gold answer for a film-QA dataset, and the CLAIM SET it will be graded by. Both come from the ANSWER\_OUTLINE; you may not add content to either. \\
        \midrule
        \textbf{Input Information:} \\
        \textit{QUESTION\_TYPE}: \{q\_type\} \\
        \textit{QUESTION}: \{question\} \\
        \\
        \textit{ANSWER\_OUTLINE}: \\
        \{outline\_claims\} \\
        \\
        \textit{NAMING}: \{name\_lists\} \\
        \midrule
        \textbf{THE CONTRACT.} Every claim you return must be plainly asserted by your prose, and every substantive assertion in your prose must be backed by a claim you return. The claim set IS the grading rubric: an answer is scored by which of these claims it asserts and which it contradicts. A claim your prose only implies will fail that test -- assert it or drop it. \\
        \\
        \textbf{PROSE.} 60-140 words, as a careful viewer would give it -- natural register: normal capitalisation (Charlie Cale, never CHARLIE CALE), pronouns are fine once a person is established, no filler (``In other words...''). Hedge claims the outline hedges; never hedge a confident claim into vagueness. Name evidence moments approximately (``just after the argument on the stairs''), never scene ids, event ids, or timestamps in seconds. Use the question's naming rules: name\_budget and named\_later freely, forbidden\_names never -- describe those people instead. \\
        \bottomrule
    \end{tabular}
    }
    \caption{\textbf{Prompt for writing the reference answer and rubric (Stage~2).}}
    \label{tab:prompt-answer}
\end{table*}

\addtocounter{table}{-1}
\begin{table*}
    \centering
    \small
    \resizebox{.93\textwidth}{!}{
    \begin{tabular}{p{.9\textwidth}}
        \toprule
        \textbf{CLAIMS.} Rewrite each outline claim you keep as one self-contained, judgeable proposition, $\leq$25 words: name the people (no pronouns), one assertion each, no annotation vocabulary, non-overlapping. Keep the outline's meaning exactly -- do not strengthen, weaken, or merge. A grader reads each claim alone against a candidate answer and must be able to say asserted /\allowbreak{} contradicted /\allowbreak{} omitted. \\
        \quad -- ATOMIC: one assertion means one. Never join assertions with ``while'', ``because'', ``and then''. Split a compound outline claim into several claims sharing its outline\_index -- a grader must never have to decide which half of a claim an answer satisfied. \\
        \quad -- NO LOCATORS: the claim asserts content, not place or time. ``On Arthur's phone call, Monk learns of the offer'' penalises an answer that has Monk learning it but names the moment differently -- write ``Monk learns that a publisher offered \$750,000''. Where and when belong to the question and the prose. \\
        \quad -- Names in claims are written normally (Charlie Cale), never in screenplay capitals (CHARLIE CALE). \\
        \\
        \textbf{ROLE} -- what the QUESTION requires, not what the outline contains. For each claim: \\
        \quad -- ``core'': a complete, correct answer to THE QUESTION must assert it. Ask: could an answer omit this and still fully answer what was asked? If yes, it is not core. \\
        \quad -- ``context'': premise, setup, or supporting fact -- restating the question's own scenario, naming who spoke, background the question did not ask to be traced. A correct answer may omit it; only contradicting it is an error. \\
        Answers are graded on recall over CORE claims alone. A padded core set punishes every correct answer for the padding; when unsure, ``context''. At least two claims should be core -- fewer means the outline did not carry the question's content, which is worth saying in dropped\_why rather than inflating a context claim. \\
        \\
        \textbf{DROPPING.} If an outline claim cannot be honestly asserted in the prose -- too weakly grounded, or it would not survive a viewer's scrutiny -- drop it: list its index in dropped and say why in dropped\_why. A dropped claim costs nothing; a claim the gold itself does not assert poisons every grade computed with it. Never pad the prose to cover a claim. \\
        \\
        Return ONLY the JSON object. \\
        \bottomrule
    \end{tabular}
    }
    \caption{\textbf{Prompt for writing the reference answer and rubric (Stage~2) (continued).}}
\end{table*}

\begin{table*}
    \centering
    \small
    \resizebox{.93\textwidth}{!}{
    \begin{tabular}{p{.9\textwidth}}
        \toprule
        \textbf{Instructions}: You verify the claims of a film-QA gold answer against the film's SCREENPLAY. The claims were derived from an automatic belief annotation that is sometimes wrong; the script is the ground truth here. Judge ONLY from the text shown. Film knowledge, plausibility, and narrative sense are not evidence. \\
        \midrule
        \textbf{Input Information:} \\
        \textit{FILM}: \{film\_id\} \\
        \\
        \textit{STORY CONTEXT (every scene of the film; heading + short neutral summary, film order; * = shown in full below)}: \\
        \{one\_line\_per\_scene\} \\
        \\
        \textit{QUESTION}: \{question\} \\
        \textit{HOLDER (the character the question is about)}: \{center\_character\} \\
        \\
        \textit{CLAIMS}: \\
        \{numbered\_claims\} \\
        \\
        \textit{SCRIPT (full text; scenes in film order; the question's anchor is \{anchor\_scene\_id\})}: \\
        \{scenes\_shown\_in\_full\} \\
        \midrule
        \textbf{For each numbered CLAIM return one label:} \\
        \quad -- ``supported'' -- the script shown establishes it; a careful reader must conclude it. \\
        \quad -- ``unsupported'' -- the script shown does not establish it (silence, absence, or an inference the text does not force). \\
        \quad -- ``contradicted'' -- the script shown says otherwise. \\
        Default to ``unsupported'' when unsure: a wrong claim corrupts every answer graded against it, while a dropped claim costs one point of coverage. \\
        \\
        \textbf{PRESENCE.} A claim that puts a person at a moment (witnesses, hears, sees, is told, learns) needs that person physically there in the text. A character who is only mentioned is not present. A character on the other end of a phone call or an INTERCUT conversation perceives only what is said on the line -- never gestures, looks, or actions in the other room. Someone asleep, unconscious, out of earshot, or elsewhere perceives nothing. The cast lists shown come from a parser and may include people who are merely mentioned; the text decides. \\
        \bottomrule
    \end{tabular}
    }
    \caption{\textbf{Prompt for verifying claims against the screenplay (Stage~3).}}
    \label{tab:prompt-verify}
\end{table*}

\addtocounter{table}{-1}
\begin{table*}
    \centering
    \small
    \resizebox{.93\textwidth}{!}{
    \begin{tabular}{p{.9\textwidth}}
        \toprule
        \textbf{BELIEF CLAIMS} -- the rules that matter: \\
        \quad -- ``X believes P'' is supported when X asserts P, acts on P, endorses or goes along with P, or is told P and shows no doubt. Do not require the proposition to be spoken verbatim; a response that builds on P shows belief in P. X being absent when not-P happened is NOT belief in P. \\
        \quad -- ``X knows /\allowbreak{} understands /\allowbreak{} learns Q'': supported when X witnesses Q, is told Q, or is present when Q becomes evident -- a realization by witnessing needs no statement. Contradicted when the film shows X acting on not-Q or being kept from Q (check the WHOLE STORY CONTEXT, not just the anchor). \\
        \quad -- ``X falsely believes P'' /\allowbreak{} ``X is wrong'' /\allowbreak{} ``X's belief is mistaken'' has two halves. The belief half follows the rules above. The falsity half is judged against the WHOLE film: it is supported only when some scene establishes not-P, and it is CONTRADICTED when the film shows P to be true -- a belief the story treats as delusion early and vindicates late is a true belief. If the belief half is supported and the falsity half is merely not established, label the claim ``supported'' and put the claim REWRITTEN WITHOUT the qualifier in ``rewrite'' (same meaning otherwise, $\leq$25 words, names not pronouns). A claim that consists ONLY of the qualifier (``X's belief is wrong'') is contradicted when P is true and unsupported when nothing settles it. \\
        \quad -- ``X does not know Q'': supported only if Q happened where X was absent and no scene in the film shows X told or learning it; ignorance by silence alone is unsupported. \\
        \quad -- ``X takes /\allowbreak{} thinks /\allowbreak{} believes Y believes P'' is supported only with an observed channel: X sees or hears Y assert or act on P; X tells Y P; X is present when Y is told P; or X and Y both witness the event P describes. In every case the text must let X identify what P is about -- overhearing one side of a phone call to an unnamed party is not a channel. X's own view of P projected onto Y is unsupported. If Y visibly does not hold P in X's presence, it is contradicted. \\
        \quad -- ``X accepts /\allowbreak{} believes /\allowbreak{} rejects what Y said'' needs a visible response from X -- words, action, or going along with it. Silence is unsupported. The tell itself (``Y tells X P'') is supported if the text shows it with X present. \\
        \quad -- ``X learns P by witnessing'' /\allowbreak{} ``by being told'' /\allowbreak{} ``by inference'': the route must be the one in the text. \\
        \quad -- ``X revises /\allowbreak{} realizes /\allowbreak{} changes their mind'': the prior belief must be visible in the same story period. A belief from a flashback or a different year does not carry into another period; a ``revision'' whose prior lives on the other side of a time jump is unsupported. \\
        \quad -- ``A happens before B'': judge from scene order; no belief reasoning needed. \\
        \\
        \textbf{STORY CONTEXT} is the WHOLE film: one line per scene, in film order, with the heading and a short neutral summary from an independent pass; the scenes shown in full below are marked *. Read all of it before judging. Use it for what happened anywhere in the film, who was present where, the order of events, which story period a scene belongs to, whether a character ever learns something, and whether a belief is true or false in the end. Never use it to establish what anyone believes or thinks -- only the full text can -- and never use it to contradict a belief a character visibly holds in the full text. \\
        \bottomrule
    \end{tabular}
    }
    \caption{\textbf{Prompt for verifying claims against the screenplay (Stage~3) (continued).}}
\end{table*}

\addtocounter{table}{-1}
\begin{table*}
    \centering
    \small
    \resizebox{.93\textwidth}{!}{
    \begin{tabular}{p{.9\textwidth}}
        \toprule
        \textbf{AFTER THE VERDICTS} -- the surviving set. Take the claims you labelled ``supported'' (with any rewrites applied). Then: \\
        \quad -- ``answers\_question'': ``yes'' if a complete answer built from the surviving CORE claims alone still fully answers the QUESTION as asked; ``no'' if the dropped claims were what the question was about (e.g. the question asks what a person works out, and every surviving claim is about someone else). \\
        \quad -- ``revised\_gold'': if you dropped or rewrote ANY claim, write the gold answer again as 60-140 words of plain prose that asserts every surviving claim, each recognisably, and asserts NOTHING beyond them -- no dropped content, no new facts, names in normal case, no scene ids. Empty string if nothing changed. \\
        \quad -- ``revised\_question'': only when answers\_question is ``no'' AND at least two surviving claims are core: write ONE replacement question that those surviving core claims answer completely, following these rules exactly -- one open interrogative (what /\allowbreak{} how /\allowbreak{} when /\allowbreak{} who), never a yes/\allowbreak{}no or an ``A or B''; open with a short diegetic anchor (a place, occasion or object, at most 12 words, never an outcome and never a scene id); ask about the TOPIC of the belief, never state its value or that it is false; no ``, and what supports that'' tail; use only character names that the original question or the surviving claims already use; the question must not be answerable from its own wording. If the surviving claims are only events with no belief, knowledge or perception in them, write the question anyway and type it honestly below. Empty string otherwise. Then also fill revised\_gold for the revised question. \\
        \quad -- ``revised\_type'': the type the revised question tests, one of: Q1 cohort knowledge split (what each of several people takes to be the case at one moment); Q2 realization moment (when someone's understanding changes); Q3 in-scene false belief (what someone takes to be the case right before acting); Q4 source attribution (how a fact reaches someone); Q5 first-order ToM (what X takes Y to believe); Q6 higher-order ToM (what X takes Y to believe about Z); Q7 deception detection (what X makes of what X is told); Q8 ordered discoveries; N1 epistemic arc across the film; N2 how X's model of Y changes; N3 causal chain to a belief. Empty string when there is no revised question. \\
        \quad -- ``revision\_note'': one sentence on why the question still holds, or what it now asks instead. Empty string if nothing changed. \\
        \\
        Return ONLY the JSON object: one verdict per claim id (``rewrite'' empty unless the qualifier rule applies, ``why'' one short sentence citing the scene), then answers\_question, revised\_gold, revised\_question, revised\_type, revision\_note. \\
        \bottomrule
    \end{tabular}
    }
    \caption{\textbf{Prompt for verifying claims against the screenplay (Stage~3) (continued).}}
\end{table*}

\begin{table*}
    \centering
    \small
    \resizebox{.93\textwidth}{!}{
    \begin{tabular}{p{.9\textwidth}}
        \toprule
        \textbf{Instructions}: You grade an ANSWER against the CLAIMS of a film-QA question. \\
        \midrule
        \textbf{Input Information:} \\
        \textit{CLAIMS}: \\
        \{numbered\_claims\} \\
        \\
        \textit{ANSWER}: \\
        \{answer\} \\
        \midrule
        \textbf{For each numbered claim, exactly one label:} \\
        \quad -- ``asserted'' -- the answer states the claim's content, in any wording. Hedged but committed (``she seems not to buy it'') still asserts. \\
        \quad -- ``contradicted'' -- the answer states something incompatible with the claim: the opposite polarity, the wrong holder, the wrong source, the inverted direction. Merely discussing the topic is not this. \\
        \quad -- ``omitted'' -- neither: the answer does not address the claim's content, or addresses it too vaguely to commit either way. An answer that declines the question omits every claim. \\
        Judge each claim independently, only against the ANSWER's text. The answer's extra content is not graded here and must not influence any verdict. \\
        Return ONLY the JSON object, one verdict per claim id. \\
        \bottomrule
    \end{tabular}
    }
    \caption{\textbf{Prompt for the judge, which grades an answer against its rubric.}}
    \label{tab:prompt-judge}
\end{table*}

\end{document}